\documentclass{article}

\usepackage{microtype}
\usepackage{graphicx}
\usepackage{subfigure}
\usepackage{booktabs} % for professional tables

\usepackage{hyperref}

\usepackage[accepted]{icml2023}

\usepackage{amsmath}
\usepackage{amssymb}
\usepackage{mathtools}
\usepackage{amsthm}
\usepackage{multirow}

\usepackage[capitalize,noabbrev]{cleveref}

\theoremstyle{plain}
\newtheorem{theorem}{Theorem}[section]

\newtheorem{lemma}[theorem]{Lemma}

\theoremstyle{definition}

\theoremstyle{remark}

\usepackage[textsize=tiny]{todonotes}

\icmltitlerunning{Maximum Causal Entropy Inverse Constrained Reinforcement Learning}

\begin{document}

\twocolumn[
\icmltitle{Maximum Causal Entropy Inverse Constrained Reinforcement Learning}



\icmlsetsymbol{equal}{*}

\begin{icmlauthorlist}
\icmlauthor{Mattijs Baert}{equal,idlab,imec}
\icmlauthor{Pietro Mazzaglia}{idlab,imec}
\icmlauthor{Sam Leroux}{idlab,imec}
\icmlauthor{Pieter Simoens}{idlab,imec}

\end{icmlauthorlist}

\icmlaffiliation{idlab}{IDLab, Department of Information Technology at Ghent University}
\icmlaffiliation{imec}{imec}

\icmlcorrespondingauthor{Mattijs Baert}{mattijs.baert@ugent.be}
%\icmlcorrespondingauthor{Firstname2 Lastname2}{first2.last2@www.uk}

\icmlkeywords{Inverse Constrained Reinforcement Learning, Principle of Maximum Causal Entropy, Constraint Inference, Safe Reinforcement Learning, Constrained Reinforcement Learning}

\vskip 0.3in
]



\printAffiliationsAndNotice{} % otherwise use the standard text.

\begin{abstract}
When deploying artificial agents in real-world environments where they interact with humans, it is crucial that their behavior is aligned with the values, social norms or other requirements of that environment.
However, many environments have implicit constraints that are difficult to specify and transfer to a learning agent. 
To address this challenge, we propose a novel method that utilizes the principle of maximum causal entropy to learn constraints and an optimal policy that adheres to these constraints, using demonstrations of agents that abide by the constraints.
We prove convergence in a tabular setting and provide an approximation which scales to complex environments.
We evaluate the effectiveness of the learned policy by assessing the reward received and the number of constraint violations, and we evaluate the learned cost function based on its transferability to other agents.
Our method has been shown to outperform state-of-the-art approaches across a variety of tasks and environments, and it is able to handle problems with stochastic dynamics and a continuous state-action space.
\end{abstract}

\section{Introduction}
\label{introduction}
The behavior of individuals in today's society is shaped by social norms, which provide predictability, safety and efficiency in human interaction. Similarly, successful integration of artificial agents in the real-world requires \textit{value alignment} of their behavioral policies
\cite{christian2020alignment,balakrishnan2019incorporating,russell2019human}.
As an example, a self-driving car should not only efficiently transport its passengers to their destination, but it should also adhere to traffic regulations, such as traffic signs, traffic lights, and road conditions, to ensure the safety of all road users.
Furthermore it is essential to program social norms into the agent prior to deployment to ensure that rule violations are avoided at all times.
%One difficulty in training value aligned agents is that the goal of the task itself and the social norms are often conflicting objectives.
%For example, a self-driving car should transport us efficiently, but without draining its battery or crossing a red light to arrive faster.
% For example, we want to ensure a self-driving car will follow all traffic regulations and maybe also minimize energy use.
%[Give some examples, autonomous car, elder care robot,...]
% with the applicable social norms
%In the framework of reinforcement learning (RL), social norms can be modeled as constraints on the learning agent's behavior.
Learning value-aligned policies can be viewed as optimizing an objective subject to a set of constraints.
In the framework of Reinforcement Learning (RL) this entails finding a policy that maximizes a combination of objectives, i.e. the goal of the task and the constraints, where each objective is weighted according to its importance.
%using the framework of Reinforcement Learning (RL) entails finding the optimal policy that maximizes a combination of objectives, where each objective is weighted according to its importance.
Not only does each combination of weights lead to a different Pareto optimal solution \cite{van2014multi}, also tuning these coefficients is a difficult and time-consuming process, as the task goal and the social norms often conflict with one another~\cite{mania2018simple}.
An alternative approach for solving the constrained RL problem is by utilizing a primal-dual method to learn the optimal coefficients \cite{bhatnagar2012online,tessler2018reward}.
Although it is straightforward to define constraints for simple environments, this is not the case for real-world settings, which often comprise multiple and implicit constraints.
Returning to the traffic example, human drivers also adopt implicit rules, for example driving slower when traffic is heavy or leaving more space when driving behind a truck.
We propose to solve this problem by learning a cost function that embodies the constraints of the environment from demonstrations of constraint-abiding agents.% (i.e., humans).
\\
Recent advancements in Inverse Reinforcement Learning (IRL) have enabled learning a reward function from expert demonstrations in challenging environments \cite{ho2016generative,finn2016guided,fu2017learning}. However, there are relatively few studies that focus on learning a cost function, also known as Inverse Constrained Reinforcement Learning (ICRL).
Although IRL and ICRL seem very related, there is a main difference in how they handle states which do not occur during the expert demonstrations.
The states which are not visited by the expert can be subdivided into a group of constrained states and a group of states which are unconstrained but correspond with low rewards.
IRL does not distinguish these two groups of unvisited states which could cause constraint violations when the agent ends up in states which were never visited by the expert.
ICRL, on the other hand, explicitly distinguishes these two groups by assigning high costs to constrained states. 
%the learned cost function and makes sure constrained states will never be visited.
%The states which are not visited by the expert can be subdivided into two groups.
%The first group consists of states which are unconstrained but correspond with low rewards.
%The second group corresponds with the constrained states.
%IRL, unlike ICRL, does not distinguish these two groups of unvisited states which could cause constraint violations when the agent ends up in states which were never visited by the expert.
%ICRL, on the other hand, explicitly distinguishes these two groups by the learned cost function and makes sure constrained states will never be visited.
\\
Various ICRL approaches use the principle of maximum entropy to learn a set of constraints that conform to expert data while remaining as unbiased as possible \cite{scobee2019maximum,stocking2021discretizingdynamics,glazier2022learningbehavioral}.
This work has been extended to domains with unknown transition dynamics and continuous state-action spaces \cite{malik2021inverse,liu2022benchmarking}.
The principle of maximum (non-causal) entropy holds true only for environments with deterministic transition dynamics.
To learn constraints in stochastic environments, \citet{mcpherson2021maximum} proposed a method based on the principle of maximum causal entropy. However, this algorithm's running time increases cubically with the size of the state space, making it unable to scale to continuous state-action spaces.
This limitation also applies to methods based on Bayesian theory \cite{papadimitriou2021bayesian}, which define the set of constraints as a collection of discrete states.
\\
Our main contribution is an ICRL method that, to the best of our knowledge, is the first to scale to environments with  a continuous state-action space and stochastic dynamics.
We formulate a new ICRL objective drawing from feature expectation matching \cite{abbeel2004apprenticeship} and in contrast to most previous work, from the principle of maximum causal entropy \cite{ziebart2010modeling} instead of the principle of maximum (non-causal) entropy \cite{ziebart2008maximum}.
From this objective, we derive two algorithms to learn both a cost function, representing constraints, and a policy that adheres to the learned constraints.
First, we propose an algorithm based on policy iteration and present proof of its convergence.
Next, we introduce an approximation of the first method that can be implemented using deep neural networks and scales to problems with a continuous state-action space.
We evaluate the performance of the learned policy by measuring the received reward and the number of constraint violations in both virtual and realistic environments.
Next, we evaluate the learned cost function on its transferability to different types of agents with other reward functions.
Finally, we examine the importance of different parts of the proposed method through an ablation study.
Based on our empirical evaluation, we have found that our method surpasses the performance of state-of-the-art techniques.
\\
The structure of the paper is as follows: In Section \ref{sec:background}, we present the necessary background information on constrained Markov decision processes and inverse reinforcement learning. 
Our proposed method is outlined in Section \ref{sec:method}.
We evaluate and compare the results of our method to existing state-of-the-art approaches in Section \ref{sec:experiments}. 
In Section \ref{sec:related}, we discuss related work.
Finally, we conclude and provide future directions in Section \ref{sec:conclusion}.

\section{Background}
\label{sec:background}
\subsection{Constrained Markov Decision Process}
A Markov decision process (MDP) is defined by a state space $\mathcal{S}$, an action space $\mathcal{A}$, a discount factor $\gamma \in [0,1]$, a transition distribution $p(s^{\prime} \mid s, a)$ which specifies the probability of transitioning to state $s^\prime$ when performing action $a$ while in state $s$, an initial state distribution $\mathcal{I}(s)$ and a bounded reward function $R: \mathcal{S} \times \mathcal{A} \mapsto [r_{\min}, r_{\max}]$ specifying the scalar reward the agent receives for applying action $a$ while in state $s$. 
We consider an agent interacting with the environment at discrete timesteps $t$ generating a sequence of transitions called a trajectory $\tau = ((s_0, a_0, s_1), ..., (s_{T-1}, a_{T-1}, s_{T}))$ of length $T$.
We define the reward of a trajectory as the sum of discounted rewards: $R(\tau) = \sum_{t=0}^{T-1}\gamma^t R(s_t,a_t)$.
At every timestep, the agent selects its action based on a policy $\pi : \mathcal{S} \mapsto \mathcal{P}(\mathcal{A}$) which is a mapping from a state to a probability distribution over actions.
The goal of forward reinforcement learning is to find the policy which maximizes the expected sum of discounted rewards: $\max_{\pi}\mathbb{E}_{\tau \sim \pi} R(\tau)$.

A constrained Markov decision process (CMDP) $\mathcal{M}^C$ \cite{altman1999constrained} is defined as an MDP augmented with a non-negative bounded cost function $C: \mathcal{S} \times \mathcal{A} \mapsto [c_{\min}, c_{\max}]$ and a budget $\alpha \geq 0$.
$C(s,a)$ denotes the cost of taking action $a$ in state $s$ and the cost of a trajectory is defined as the sum of discounted costs: $C(\tau) = \sum_{t=0}^{T-1}\gamma^t C(s_t,a_t)$.
The goal of constrained reinforcement learning is defined as to find the policy which maximizes the expected sum of discounted rewards while the expected sum of discounted costs is smaller than the budget:
\begin{equation}
\max_{\pi} \mathbb{E}_{\tau \sim \pi} R(\tau) \;\; \textrm{s.t.} \;\; \mathbb{E}_{\tau \sim \pi} C(\tau) < \alpha.
\end{equation}
When $\alpha$ is 0, constraints can be interpreted as hard constraints meaning the resulting policy should never visit states which incur any cost.
When $\alpha > 0$, the resulting policy is allowed to obtain cost $>0$ in some cases, i.e. soft constraints.

\subsection{Inverse Reinforcement Learning}
Inverse Reinforcement Learning (IRL) is a learning problem which, given an expert data distribution $\mathcal{D}$ represented by a finite set of trajectories, tries to identify the reward function $R(s,a)$ the expert agent is optimizing.
Assume the unknown reward function is defined as $R^{\prime}(s,a) = \omega \cdot \phi(s,a)$ with $\omega \in \mathbb{R}^k$ a $k$-dimensional vector of real numbers and  $\phi: \mathcal{S} \times \mathcal{A} \mapsto \mathbb{R}^{k}_{+}$ denoting a fixed mapping from the state-action space to a $k$-dimensional vector of non-negative features.
We define the feature representation of a trajectory as the sum of discounted feature vectors: $\phi(\tau) = \sum^{T-1}_{t=0} \gamma^t \phi(s_t,a_t)$.
With forward RL we can obtain a policy which maximizes $R^{\prime}$ (i.e. the nominal policy).
The problem of IRL can then be rephrased as finding values of $\omega$ such that the expected features when following the nominal policy match the expected features under the expert data distribution, i.e. feature expectation matching \cite{abbeel2004apprenticeship}.
However, this is an ill-posed problem as many assignments of $\omega$ will result in matching expected features.
To resolve this ambiguity, \citet{ziebart2010modeling} proposed to choose from all values $\omega$ which result in matching feature expectation those that maximize the causal entropy, i.e. Maximum Causal Entropy Inverse Reinforcement Learning (MCE-IRL).
Under Markovian dynamics the causal entropy of a trajectory is defined as the discounted sum of the conditional entropy of the current action given the current state: $H(\tau) = \sum_{t=0}^{T-1} \gamma^t H(a_t \mid s_{t})$.
Although the original feature matching objective only considers reward functions that are linear in  $\phi$, state-of-the-art methods are scalable to complex problems by parameterizing $R^{\prime}$ with a deep neural network \cite{wulfmeier2015maximum,ho2016generative}.
For an elaborate introduction to MCE-IRL we refer the reader to the work of \citet{gleave2022primer}.

\section{Maximum Causal Entropy Inverse Constrained Reinforcement Learning}
\label{sec:method}
Inverse Constrained Reinforcement Learning (ICRL) methods try to learn a cost function which represents the constraints applicable to a particular environment from demonstrations of agents abiding these constraints.
Inspired by the principle of maximum causal entropy \cite{ziebart2010modeling} and feature expectation matching \cite{abbeel2004apprenticeship}, we propose a novel objective for ICRL.
%First we derive a soft value iteration algorithm and prove convergence in a tabular setting.
%Next, we derive a soft policy gradient algorithm which scales to continuous state-action spaces.
%Finally, we extend this to non-linear constraints by parameterizing the feature mapping by a neural network.
We adopt the IRL terminology and will refer to constraint-abiding agents as expert agents.

\subsection{Objective}
Given an expert data distribution $\mathcal{D}$ and a reward function $R$, we define the primal problem as finding the stochastic policy $\pi(a \mid s)$ (i.e. the nominal policy) which maximizes the given reward signal and the causal entropy while matching feature expectations between the nominal policy and the expert data distribution:
\begin{align}
    \label{eq:primal}
    \begin{split}
    & \max_{\pi \in \Pi} \quad \mathbb{E}_{\tau \sim \pi} \left[
    R(\tau) + \beta H(\tau) \right]  \quad \textrm{s.t.} \\
    & \mathbb{E}_{\tau \sim \mathcal{D}}  \left[\phi(\tau)\right] - \mathbb{E}_{\tau \sim \pi} \left[\phi(\tau)\right] \leq \alpha.
    \end{split}
\end{align}
%\begin{align}
%    \label{eq:primal}
%    \begin{split}
%    & \max_{\pi \in \Pi} \quad \mathbb{E}_{\tau \sim \pi} \left[\sum_{t=0}^{T-1} \gamma^t \left( 
%    R(s_t,a_t) + \beta H(a_t \mid s_t) \right) \right]  \quad \textrm{s.t.} \\
%    & \mathbb{E}_{\tau \sim \mathcal{D}}  \left[\sum_{t=0}^{T-1} \gamma^{t} \phi(s_{t},a_{t})\right] - \mathbb{E}_{\tau \sim \pi} \left[\sum_{t=0}^{T-1} \gamma^{t} \phi(s_t,a_t)\right] \leq \alpha.
%    \end{split}
%\end{align}
Where $\Pi$ denotes the set of normalized policies with non-negative probabilities for all states and actions:
\begin{multline}
    \label{eq:theta-search-space}
    \pi \in \Pi \;\;\; \Leftrightarrow \;\;\; \pi(a \mid s) \geq 0 \;\; \\ \textrm{and} \;\; \int \pi(a \mid s) \, \mathrm{d}a=1 \;\;\;\; (\forall s \in \mathcal{S}).
\end{multline}
We define the entropy coefficient $\beta$ to balance the reward and the entropy objective.
%$S_{0:T-1}$ and $A_{0:T-1}$ denote series of random state and action variables respectively sampled from $\pi_{\theta}$ or $\mathcal{D}$.
The feature matching constraint is defined as an inequality constraint with a budget $\alpha$.
%When $\alpha > 0$, small feature mismatches are allowed.
To solve the constrained optimization problem in eq.~\eqref{eq:primal}, we define the Lagrangian and calculate it's saddle points by solving the dual problem.
The Lagrangian of this primal problem is found by replacing the feature matching constraint by a weighted penalty term with weights $\lambda \in \mathbb{R}^k : \lambda_i \geq 0 \; \textrm{for} \; 0,\:...,\:k-1$ (i.e. dual variable vector).
The Karush-Kuhn-Tucker (KKT) conditions state that for any pair of optimal points $(\pi, \lambda)$, all values of $\lambda$ should be non-negative (Sec. 5.5.3 \citet{boyd2004convex}).
%because of the KKT conditions all values of $\lambda$ should be non-negative.
The Lagrangian of the primal problem (eq.~\ref{eq:primal}) can be written as:
\begin{multline}
    \label{eq:lagrangian}
    \mathcal{L}(\pi, \lambda) = \mathbb{E}_{\tau \sim \pi} \left[R(\tau) + \beta H(\tau) \right] \\
    + \lambda \cdot \left(\mathbb{E}_{\tau \sim \mathcal{D}}  \left[\phi(\tau)\right] -\mathbb{E}_{\tau \sim \pi} \left[\phi(\tau)\right] - \alpha\right).
\end{multline}%\endgroup
% Note that $\lambda$ has the same dimensions $k$ as $\phi(s, a)$.
% Then the dual function can be defined as
% \begin{equation}
%     g(\lambda) = \max_{\theta \in \Theta}\mathcal{L}(\theta, \lambda),
% \end{equation}
% and the dual problem
% \begin{equation}
    %\min_{\lambda \in \mathbb{R}^k_+} g(\lambda).
%     \min_{\lambda} g(\lambda).
% \end{equation}
Then the dual problem can be defined as
\begin{equation}
    \label{eq:dual}
    \min_{\lambda} \max_{\pi \in \Pi}\mathcal{L}(\pi, \lambda).
\end{equation}
We solve the dual problem by first maximizing the Lagrangian with respect to the policy $\pi$ and then taking a single gradient step minimizing the Lagrangian with respect to the dual variable vector $\lambda$ (i.e. dual ascent).
\begin{theorem}
    \label{lemma:dual-convergence}
    The dual problem can be solved by alternately optimizing for $\pi$ and $\lambda$.
    When the policy update is performed on a faster time-scale than the updates on the dual variable vector, $\pi$ and $\lambda$ will converge to a local optimum. \\
    Proof. Chapter 6 of \cite{borkar2009stochastic}.
\end{theorem}
We first discuss optimizing eq.~\eqref{eq:lagrangian} w.r.t. the policy $\pi$ (Sec. \ref{subsec:soft-polic-iteration}) and then optimizing the dual variables $\lambda$ (Sec. \ref{subsec:optimizing-the-dual-variables}).
Next, in Sec. \ref{subsec:soft-policy-gradient} we present an approximation for the method presented in Sec. \ref{subsec:soft-polic-iteration}.
Finally, we extend our method to non-linear cost functions (Sec. \ref{subsec:non-linear-cost-functions}).
Algorithm \ref{alg:algorithm} provides an overview of the different steps of the proposed method.
% When the policy update is performed on a faster time-scale than the updates on the dual variable vector, $\pi$ and $\lambda$ will converge to a local minima, for a full proof we refer the reader to Chapter 6 of \cite{borkar2009stochastic}.
\begin{algorithm}[tb]
\caption{MCE-ICRL algorithm}
\label{alg:algorithm}
\textbf{Input}: reward function $R(s,a)$, expert demonstrations $\mathcal{D}$ \\
\textbf{Output}: cost function $C$, nominal policy $\pi$ \\
\textbf{Parameter}: number of iterations $\eta$
\begin{algorithmic}[1]
\STATE Initialize $\theta$, $\lambda$ and $\zeta$ 
\STATE Learn nominal policy $\pi_{\theta}^0$ with zero cost
\STATE Pre-train $\phi_{\zeta}$ using $\tau \sim \pi_{\theta}^0$ and $\tau \sim \mathcal{D}$ (Sec. \ref{subsec:non-linear-cost-functions})
\FOR{$i \leftarrow 0$ {\bfseries to} $\eta$}
    \STATE Learn $\pi_{\theta}$ maximizing eq.~\eqref{eq:lagrangian} w.r.t. to the policy parameters $\theta$ (Sec. \ref{subsec:soft-polic-iteration} and \ref{subsec:soft-policy-gradient}).
    \STATE Perform gradient descent on $\lambda$ and $\zeta$ minimizing eq.~\eqref{eq:lagrangian} w.r.t. $\lambda$ (Sec. \ref{subsec:optimizing-the-dual-variables}) and $\zeta$ (Sec.~\ref{subsec:non-linear-cost-functions}).
\ENDFOR
\end{algorithmic}
\end{algorithm}

\subsection{Policy Iteration}
\label{subsec:soft-polic-iteration}
Maximizing the Lagrangian w.r.t. the policy corresponds with solving a planning problem.
We derive a policy iteration algorithm which alternates between a policy evaluation and a policy improvement step.
We will show that this algorithm converges to the optimal policy (Theorem \ref{theorem:soft-policy-iteration}).
\begin{lemma}
    \label{lemma:optimal-policy}
    The optimal policy $\pi^*$ maximizing eq.~\eqref{eq:lagrangian} is given by
    \begin{equation}
    \label{eq:opt-policy}
        \pi^*(a_t \mid s_t) = \exp\left(\dfrac{1}{\beta} \big(Q^{*}(s_t,a_t) - V^{*}(s_t)\big)\right).
    \end{equation}
    With the action-value function defined by
    \begin{equation}
        \label{eq:q}
        Q(s_t,a_t) = R(s_t,a_t) - \lambda \cdot \phi(s_t,a_t) + \gamma \, \mathbb{E}_{p} \left[V(s_{t+1}) \right],
    \end{equation}
    and the state-value function by
    \begin{equation}
        V(s_t) = \beta \log \int \exp\left(\dfrac{1}{\beta} Q(s_t, a_t)\right) \, \mathrm{d}a.
    \end{equation}
    Proof. See Section \ref{sec:proof-tabular} in the appendix.
\end{lemma}
The Q-function in eq.~\ref{eq:q} resembles the Q-function of the standard entropy-regularized RL objective \cite{haarnoja2017reinforcement} with an additional term that subtracts $\lambda \cdot \phi(s,a)$ from the reward.
In the policy evaluation step, our goal is to compute the action-value function for an arbitrary policy $\pi$.
We can obtain the Q-function, for a fixed policy, by iteratively applying a modified Bellman backup operator $\mathcal{T}^\pi$ given by
\begin{multline}
    \label{eq:bellman-operator}
    \mathcal{T}^\pi Q(s_t,a_t) \triangleq R(s_t,a_t) - \lambda \cdot \phi(s_t,a_t) + \gamma \, \mathbb{E}_{p} \left[V(s_{t+1}) \right],
\end{multline}
with $V(s) = \mathbb{E}_{\pi} \left[Q(s,a) - \beta\log \pi(a \mid s)\right]$ (derived from eq.~\eqref{eq:opt-policy}).
During the policy improvement step, we update the policy for each state towards the optimal policy (eq.~\eqref{eq:opt-policy}) using the computed Q-function according to
%We use the information projection defined in terms of the Kullback-Leibler divergence to project the new policy into the valid set of policies $\Pi$.
%Thus, the policy improvement corresponds with updating the policy for each state according to
\begin{equation}
    \label{eq:kl}
    \pi_{\textrm{new}} = \arg\min_{\pi^{\prime} \in \Pi} \textrm{D}_{\textrm{KL}} \left(\pi^{\prime}(\cdot \mid s_t) \,  \biggr\rvert  \biggr\rvert \, \pi^*_{\textrm{old}} \right).
\end{equation}
\begin{theorem}[Policy Iteration]
    \label{theorem:soft-policy-iteration}
    Continually applying policy evaluation (using eq.~\eqref{eq:bellman-operator}) and policy improvement (using eq.~\eqref{eq:kl}) starting from any $\pi \in \Pi$, converges to the optimal policy $\pi^*$ for which $Q^{\pi^*} (s_t, a_t) \geq Q^{\pi} (s_t, a_t)$ for all $\pi \in \Pi$ and $(s_t,a_t) \in \mathcal{S} \times \mathcal{A}$, assuming $|\mathcal{A}| < \infty$. \\
    Proof. See Section \ref{sec:proof-soft-policy-iteration} in the appendix.
\end{theorem}

\subsection{Optimizing the Dual Variables}
\label{subsec:optimizing-the-dual-variables}
\begin{lemma}
    \label{lemma:dual-convex}
    The Lagrange dual problem (eq.~\eqref{eq:dual}) is a convex optimization problem. \\
    Proof. See Section \ref{sec:proofs-dual-convex} in the appendix.
\end{lemma}
Because of lemma \ref{lemma:dual-convex}, the dual variables will converge to a global minimum when optimizing using gradient descent.
Then, theorem \ref{lemma:dual-convergence} and \ref{theorem:soft-policy-iteration} and lemma \ref{lemma:dual-convex} prove convergence for algorithm \ref{alg:algorithm} in the tabular setting.
\\
The step we need to take on the dual variable vector is obtained by taking the gradient of the Lagrangian with respect to $\lambda$:
\begin{multline}
    \label{eq:lag_lambda}
    \triangledown_{\lambda} \mathcal{L}(\theta, \lambda)= \mathbb{E}_{\tau \sim \mathcal{D}}  \left[\phi(\tau) \right] - \mathbb{E}_{\tau \sim \pi_{\theta}} \left[\phi(\tau)\right] - \alpha.
\end{multline}
After each update, negative values of $\lambda$ are clamped to zero since all values of $\lambda$ should be non-negative.
Feature expectations under the nominal policy are calculated by iterating over a set of trajectories sampled from the learned policy.
To get the feature expectations under the expert policy we iterate over the set of given expert trajectories.
Intuitively, when interpreting $\lambda \cdot \phi(s,a)$ as a cost function in eq~\eqref{eq:q}, we update $\lambda$ such that features which occur in trajectories sampled from the nominal policy and not in the expert data result in a higher cost.

\subsection{Policy Gradient}
\label{subsec:soft-policy-gradient}
The algorithm presented in Sec.~\ref{subsec:soft-polic-iteration} only considers a tabular setting with a limited state-action space.
Because of this, we present a practical algorithm which scales to continuous state-action spaces.
We parameterize the policy with a neural network with parameters $\theta$.
To recover the optimal policy, we learn the optimal parameters using a policy gradient algorithm.
\begin{lemma}
\label{lemma:gradient-theta-objective}
Suppose that the policy $\pi_{\theta}$ is differentiable with respect to its parameters $\theta$. Then 
\begin{multline}
    \label{eq:lag_theta}
    \triangledown_{\theta} \mathcal{L}(\theta, \lambda) = \mathbb{E}_{\tau \sim \pi_{\theta}} \bigg[\sum_{t=0}^{T-1}\triangledown_{\theta} \log \pi_{\theta}(a_t \mid s_t) \\
    \big(Q(s_t, a_t) - \beta \log \pi_{\theta}(a_t \mid s_t) - \sum_{t^{\prime}=t}^{T-1}\gamma^{t^{\prime}-t}\beta \big) \bigg].
\end{multline}
\textit{Proof.} See Section \ref{sec:proof-lag} in the appendix.
\end{lemma}
The gradient in eq.~\eqref{eq:lag_theta} resembles the standard policy gradient \cite{williams1992simple}, with the difference that a term $-\beta \log \pi_{\theta}(a_t \mid s_t) - \sum_{t^{\prime}=t}^{T-1}\gamma^{t^{\prime}-t}\beta$ was added to the action-value function.
The vanilla policy gradient is characterized by no bias but high variance.
Variance can be reduced while keeping bias low, by introducing a  state-dependent baseline \cite{williams1992simple}.
\begin{lemma}
    \label{thm:baseline}
    In the gradient of the Lagrangian (eq.~\eqref{eq:lag_theta}), we can replace $\sum_{t^{\prime}=t}^{T-1}\gamma^{t^{\prime}-t}\beta$ with a state dependent baseline $b(s_t)$. \\
    \textit{Proof.} See Section \ref{sec:proofs-baseline} in the appendix.
\end{lemma}
Because of lemma~\ref{thm:baseline} we can replace $\sum_{t^{\prime}=t}^{T-1}\gamma^{t^{\prime}-t}\beta$ with the state-value function $V^{\pi_{\theta}}(s_t)$ which is a commonly used baseline.
The gradient can then be written as
\begin{multline}
    \triangledown_{\theta} \mathcal{L}(\theta, \lambda) = \mathbb{E}_{\tau \sim \pi_{\theta}} \bigg[\sum_{t=0}^{T-1}\triangledown_{\theta} \log \pi_{\theta}(a_t \mid s_t) \\
    \big(Q^{\pi_{\theta}}(s_t,a_t) - \beta \log \pi_{\theta}(a_t \mid s_t) - V^{\pi_{\theta}}(s_t) \big)\bigg] \\
    = \mathbb{E}_{\tau \sim \pi_{\theta}} \bigg[\sum_{t=0}^{T-1}\triangledown_{\theta} \log \pi_{\theta}(a_t \mid s_t)A^{\pi_{\theta}}(s_t,a_t)\bigg]
\end{multline}
with $A^{\pi_{\theta}}(s,a)$ the advantage function.
We adopt generalized advantage estimation (GAE) \cite{schulman2015high} to obtain an approximation of the advantage.

\subsection{Non-Linear Cost Functions}
\label{subsec:non-linear-cost-functions}
We defined the cost function as the dot product of the transposed Lagrange multiplier $\lambda$ and the feature representation $\phi(s,a)$.
This means we can only learn cost functions which are linear with respect to $\phi$.
In order to learn non-linear cost functions, we replace the hard-coded feature representation with a neural network with parameters $\zeta$.
We treat $\zeta$ as one of the dual variables and thus update them together with $\lambda$.
To do this we derive the gradient of the Lagrangian w.r.t. $\zeta$:
\begin{multline}
    \label{eq:lag_zeta}
    \triangledown_{\zeta}\mathcal{L}(\theta, \lambda, \zeta) = \\
    \lambda \cdot \left(\mathbb{E}_{\tau \sim \mathcal{D}} \left[\triangledown_{\zeta}\phi_{\zeta}(\tau)\right] -  \mathbb{E}_{\tau \sim \pi_{\theta}}  \left[\triangledown_{\zeta}\phi_{\zeta}(\tau)\right]\right).
\end{multline}
The outputs of $\phi_{\zeta}$ are passed through a sigmoid activation function to assure positive feature values.
\\
We obtained better results by pre-training the feature encoder network $\phi_{\zeta}$.
To do this, we define a feature decoder network and train the encoder and decoder as an autoencoder trying to minimize the reconstruction error between the original data (i.e. encoder input) and the reconstruction (i.e. decoder output).
The autoencoder is trained on nominal and expert data.

\section{Experiments}
\label{sec:experiments}
We validated our approach on a gridworld environment and the ICRL benchmark proposed by \citet{liu2022benchmarking}, which comprises 5 robotic domains and a realistic traffic environment.
Expert trajectories are sampled from a policy trained using reward constrained policy optimization \cite{tessler2018reward} given ground truth constraints.
We calculate the nominal policy using Proximal Policy Optimization (PPO) \cite{schulman2017proximal}.
We also perform a study on the transferability of the learned cost function to other agents, an ablation study on the pre-training of the feature encoder and the influence of the hyperparameter $\beta$.
An overview of the used hyperparameters and other details about the experiments are presented in appendix \ref{sec:experimental-settings}.
We compare our method, which we denote by MCE-ICRL, with three baseline methods: 
\begin{itemize}
    \item \textbf{Generative Adversarial Constraint Learning (GACL)}: This method is based on the well-established imitation learning method: generative adversarial imitation learning (GAIL) \cite{ho2016generative}.
    Following \citet{malik2021inverse}, GAIL can be adopted for ICRL by training the discriminator $D(s,a)$ to assign 0's to constrained state-action pairs and 1's to valid ones.
    During the forward step, the policy is trained by maximizing $\bar{r}(s,a) = r(s,a) + \log D(s,a)$.
    \item \textbf{Maximum Entropy Inverse Constrained Reinforcement Learning (ME-ICRL)} \cite{malik2021inverse}:
    %: They extend the work of \citet{scobee2019maximum} to continuous state and action spaces which enables them to learn cost functions in realistic environments. 
    This method adopts the principle of maximum entropy \cite{ziebart2008maximum} for modeling the probability of trajectories.
    \item \textbf{Variational Maximum Entropy Inverse Constrained Reinforcement Learning (VME-ICRL)} \cite{liu2022benchmarking}: This method also builds on the principle of maximum entropy but models constraints as a $\beta-$distribution.
    %The authors claim inferring a distribution of constraints extends the applicability of ICRL to environments with stochastic dynamics.
\end{itemize}
We evaluate the nominal policy at different points during training by sampling trajectories from it and reporting the average obtained reward and the average constraint violation rate.
The reward denotes the total reward an agent has received during a trajectory.
When a constraint is violated the trajectory is terminated immediately (only during evaluation).
The constraint violation rate is the fraction of all timesteps of a trajectory that violate at least one constraint.
\begin{figure}
\begin{center}
\subfigure{\includegraphics[width=0.4\columnwidth]{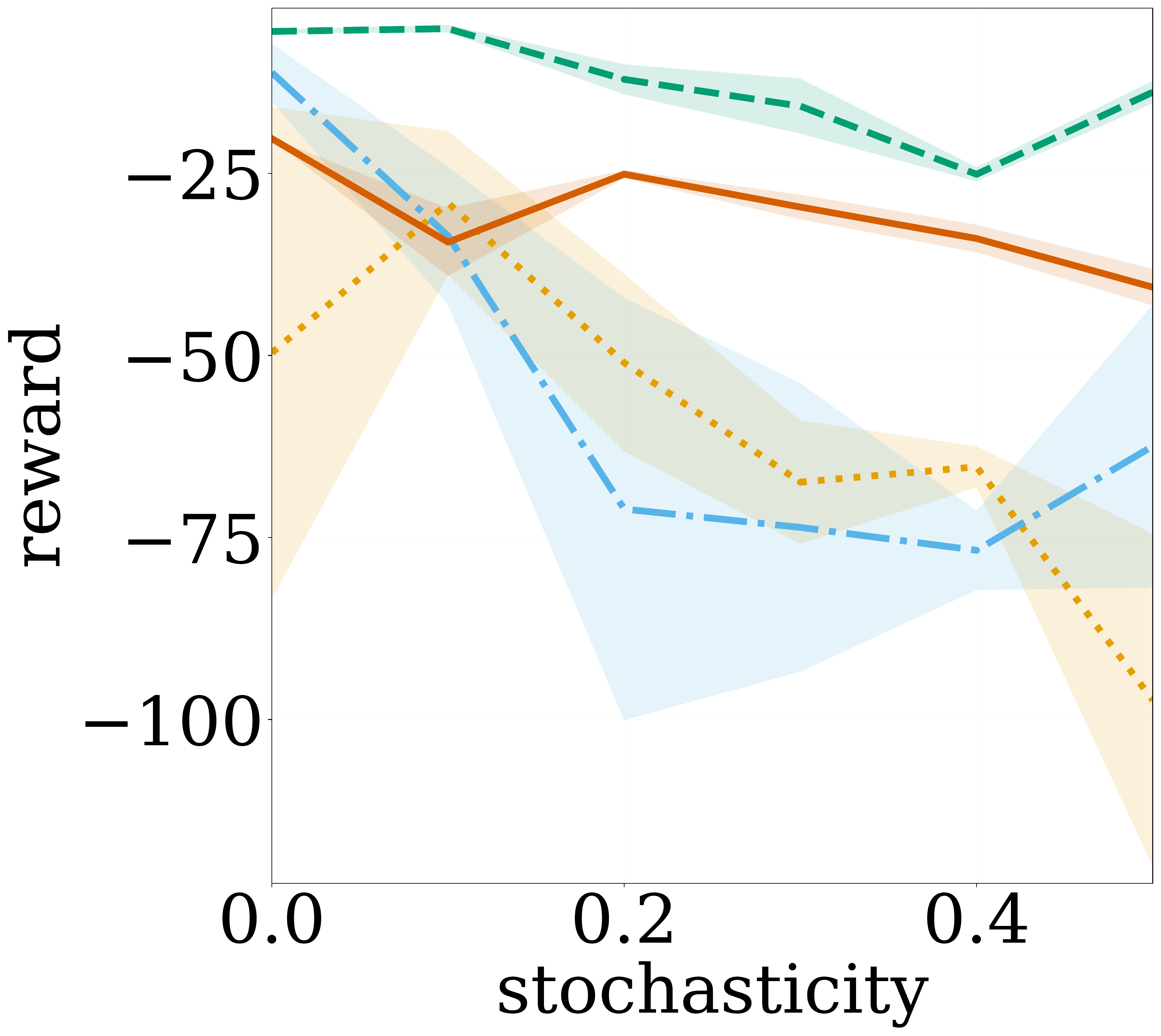}}\hspace{0.1in}
    \subfigure{\includegraphics[width=0.4\columnwidth]{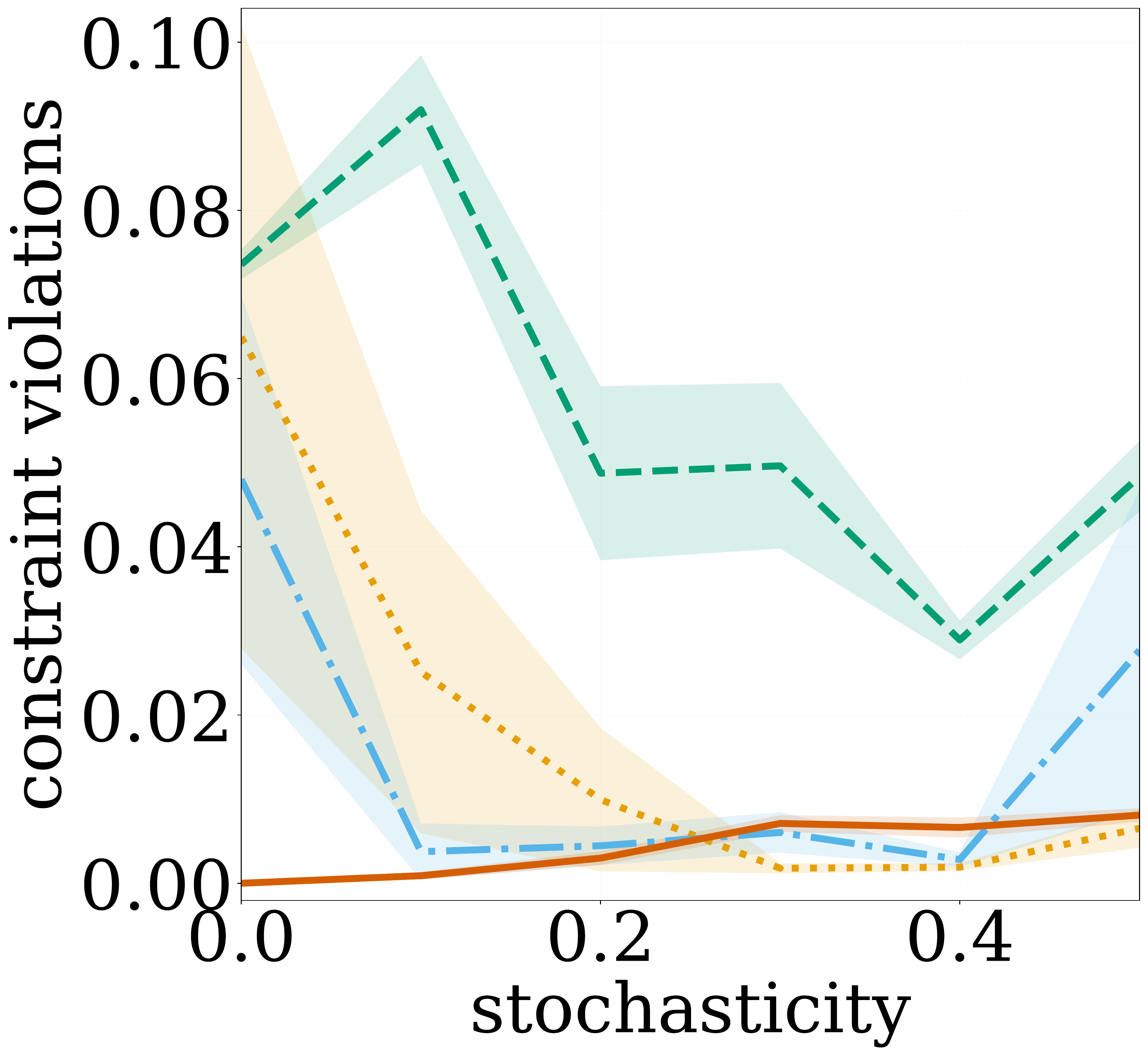}}\\
    \subfigure{\includegraphics[width=\columnwidth]{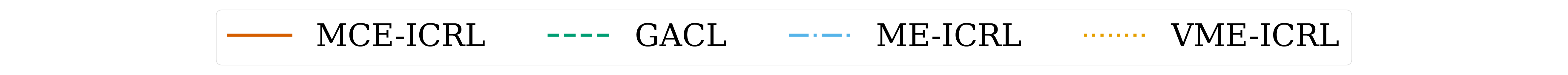}}\\
    \caption{Evaluation of the different methods in the gridworld environment for increasing stochasticity: reward (left) and constraint violation rate (right) of trajectories sampled from the nominal policy after training. Results are averaged over 5 random seeds. The x-axis corresponds with the stochasticity. The shaded regions correspond to the standard error.}
    \label{fig:gridworld}
\end{center}
\vskip -0.2in
\end{figure}
\begin{figure*}[t]
\begin{center}
    \hspace{0.01cm}
    \subfigure{\includegraphics[width=0.18\textwidth]{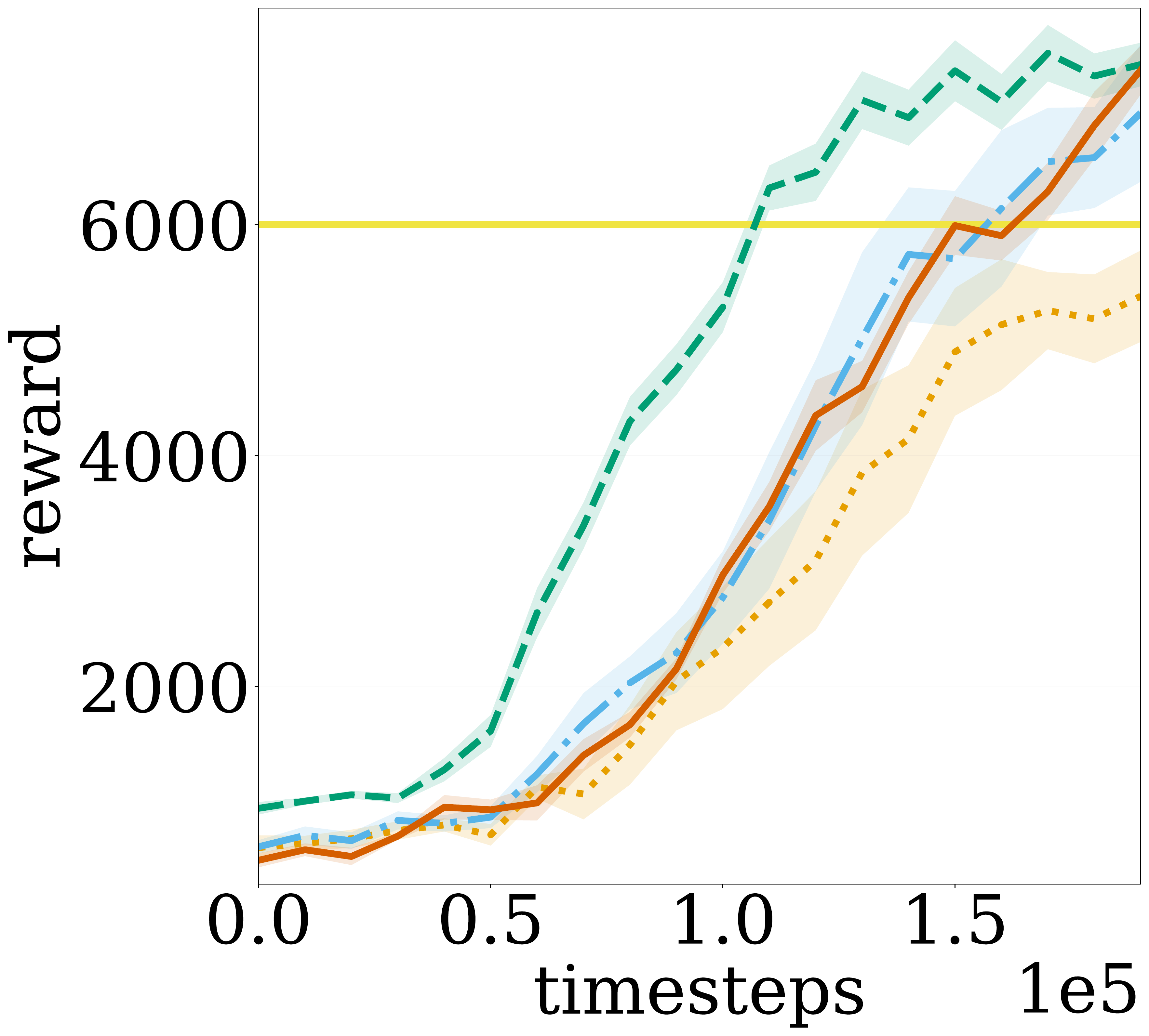}} %\hspace{0.2in}
    \subfigure{\includegraphics[width=0.17\textwidth]{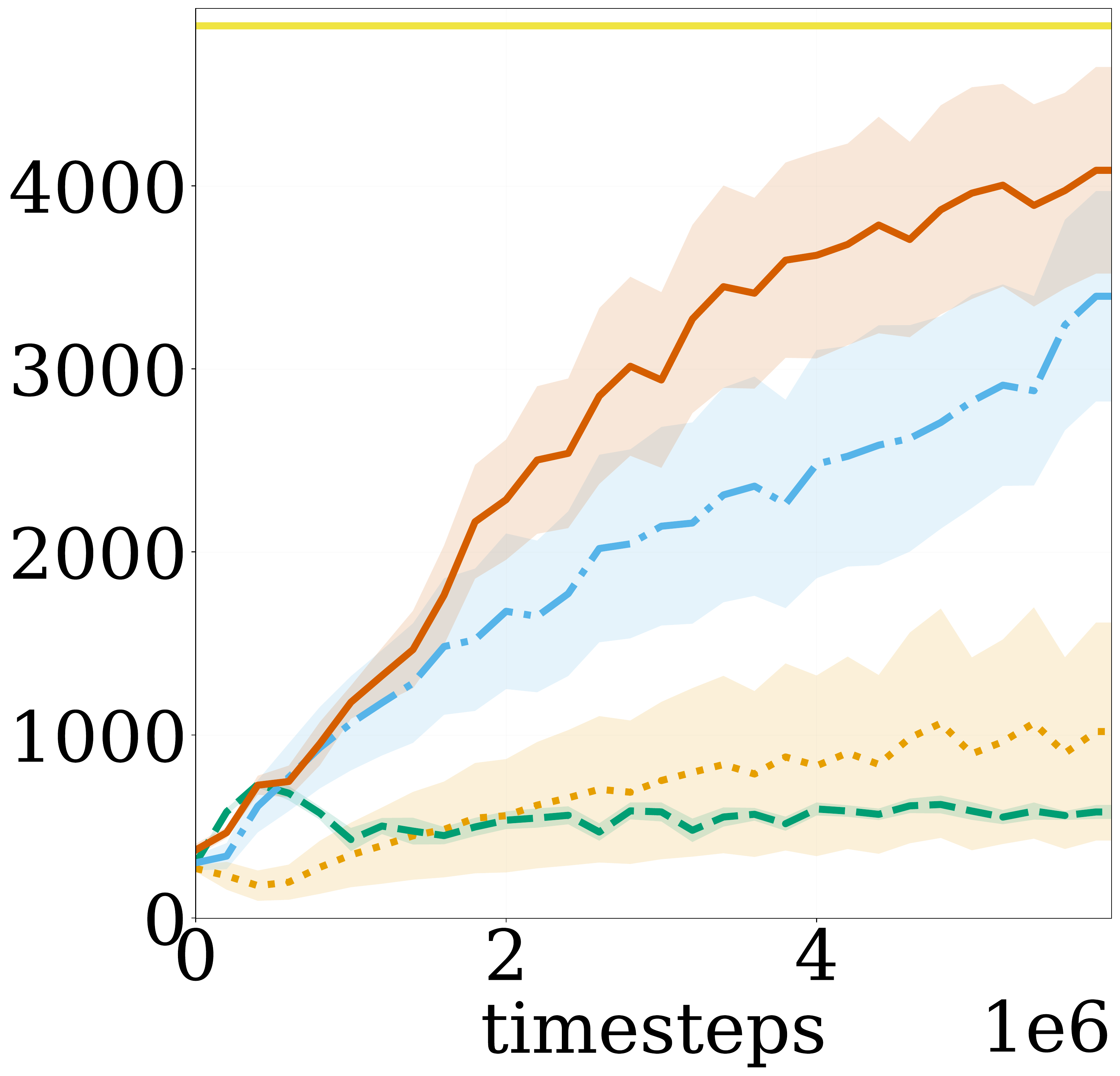}}
    \subfigure{\includegraphics[width=0.165\textwidth]{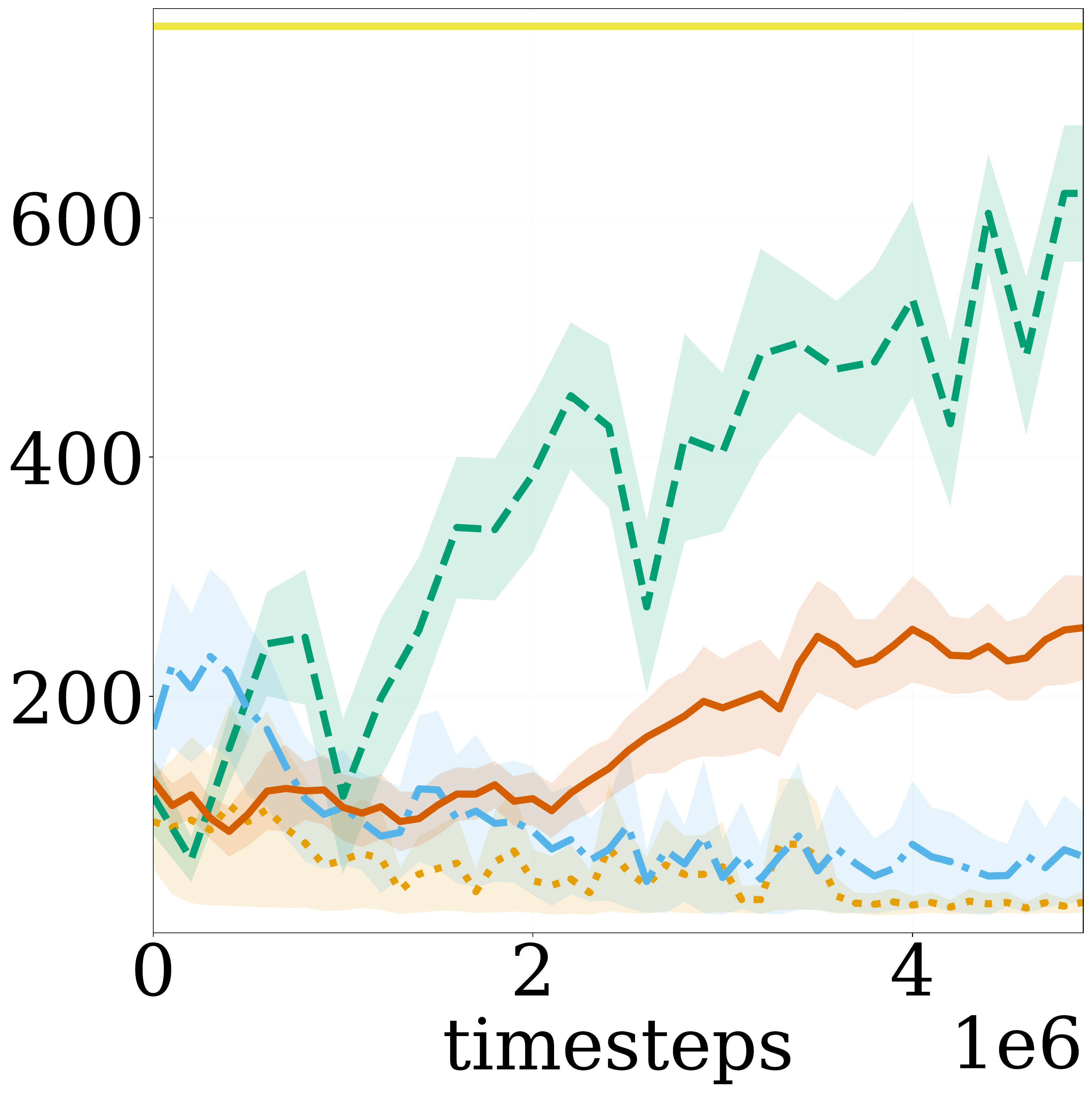}}
    \subfigure{\includegraphics[width=0.172\textwidth]{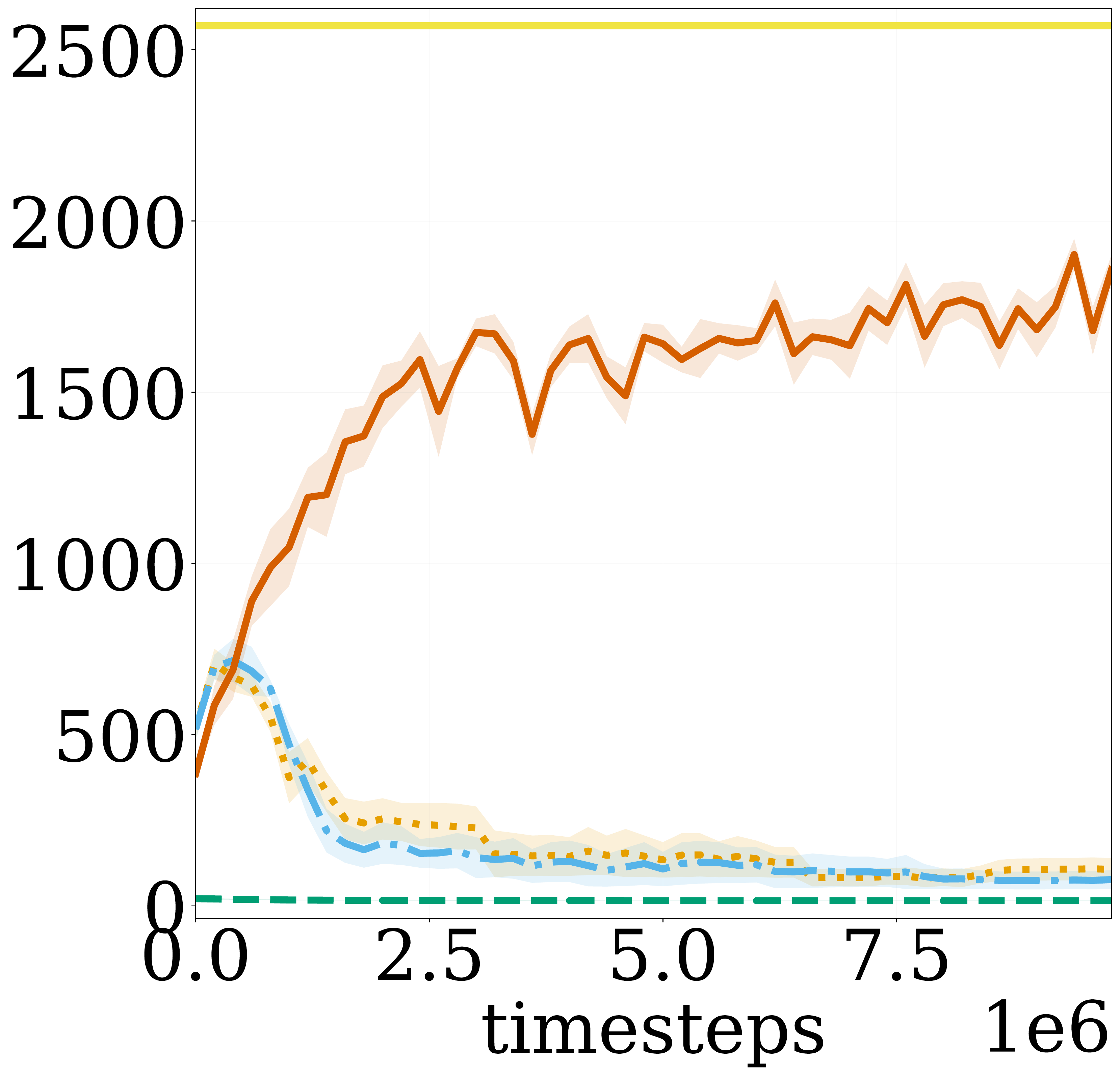}}
    \subfigure{\includegraphics[width=0.16\textwidth]{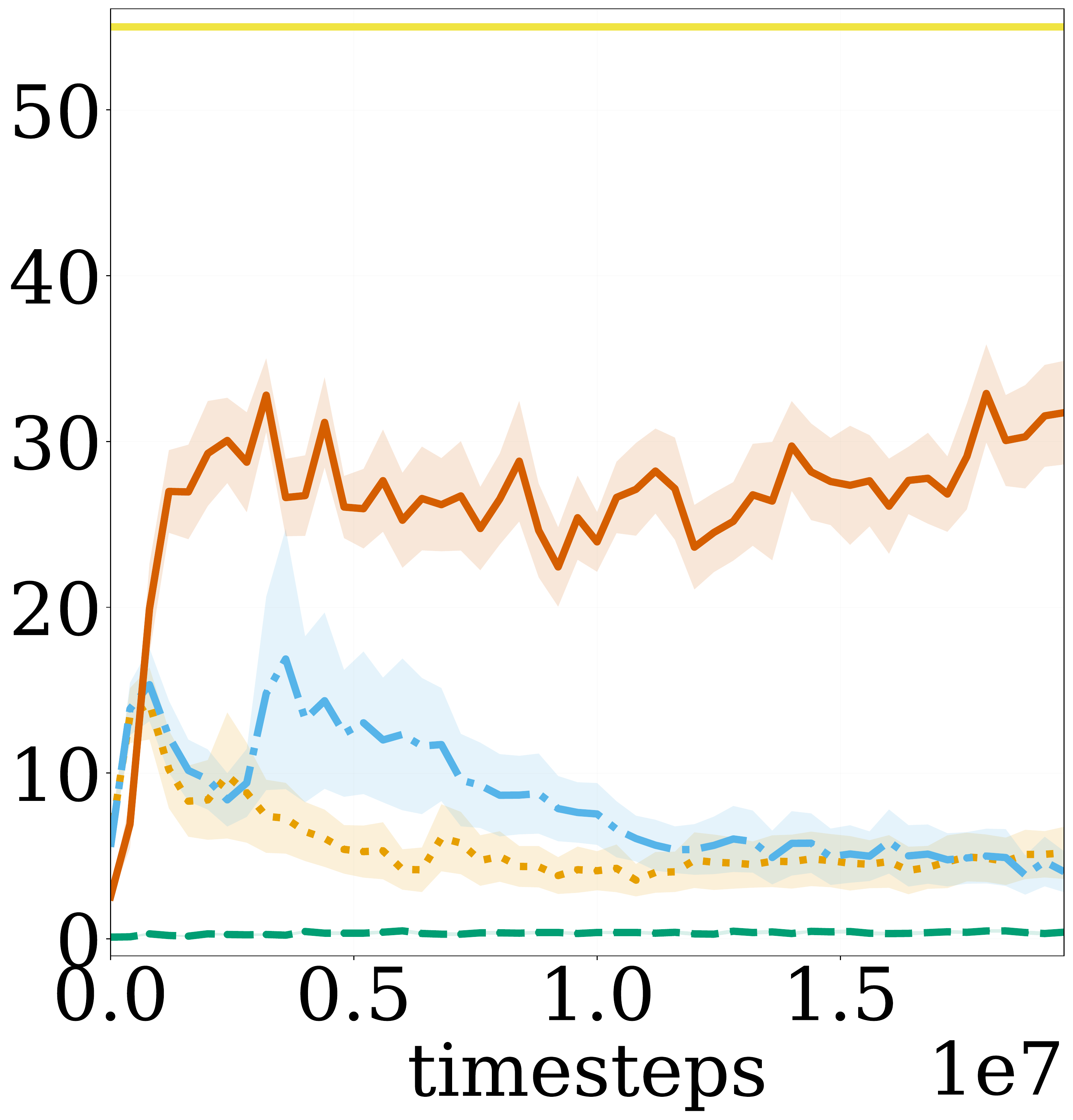}}\\
    \setcounter{subfigure}{0}
    \subfigure[Ant]{\includegraphics[width=0.185\textwidth]{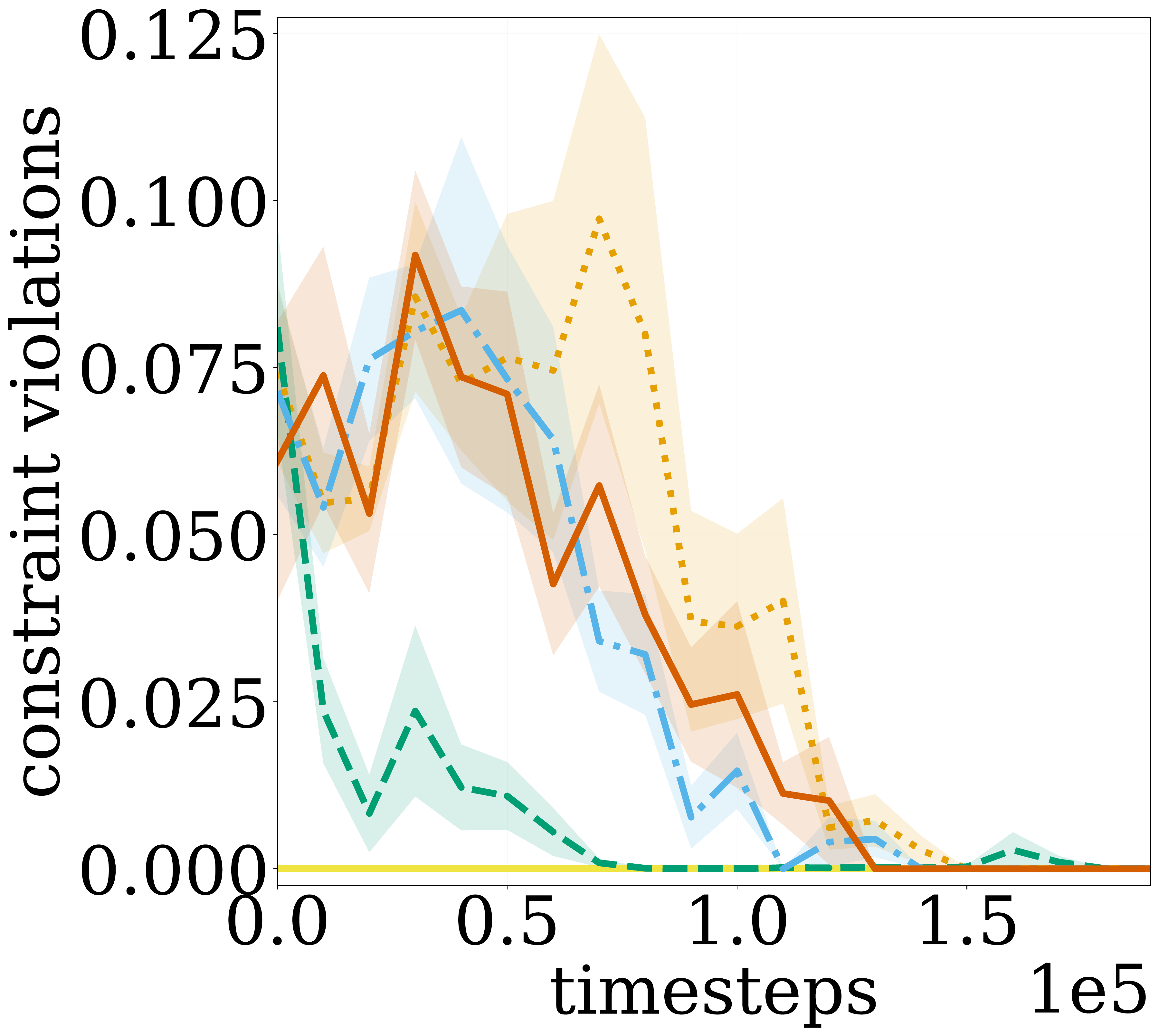}} 
    \hspace{0.12cm}
    \subfigure[Half-cheetah]{\includegraphics[width=0.16\textwidth]{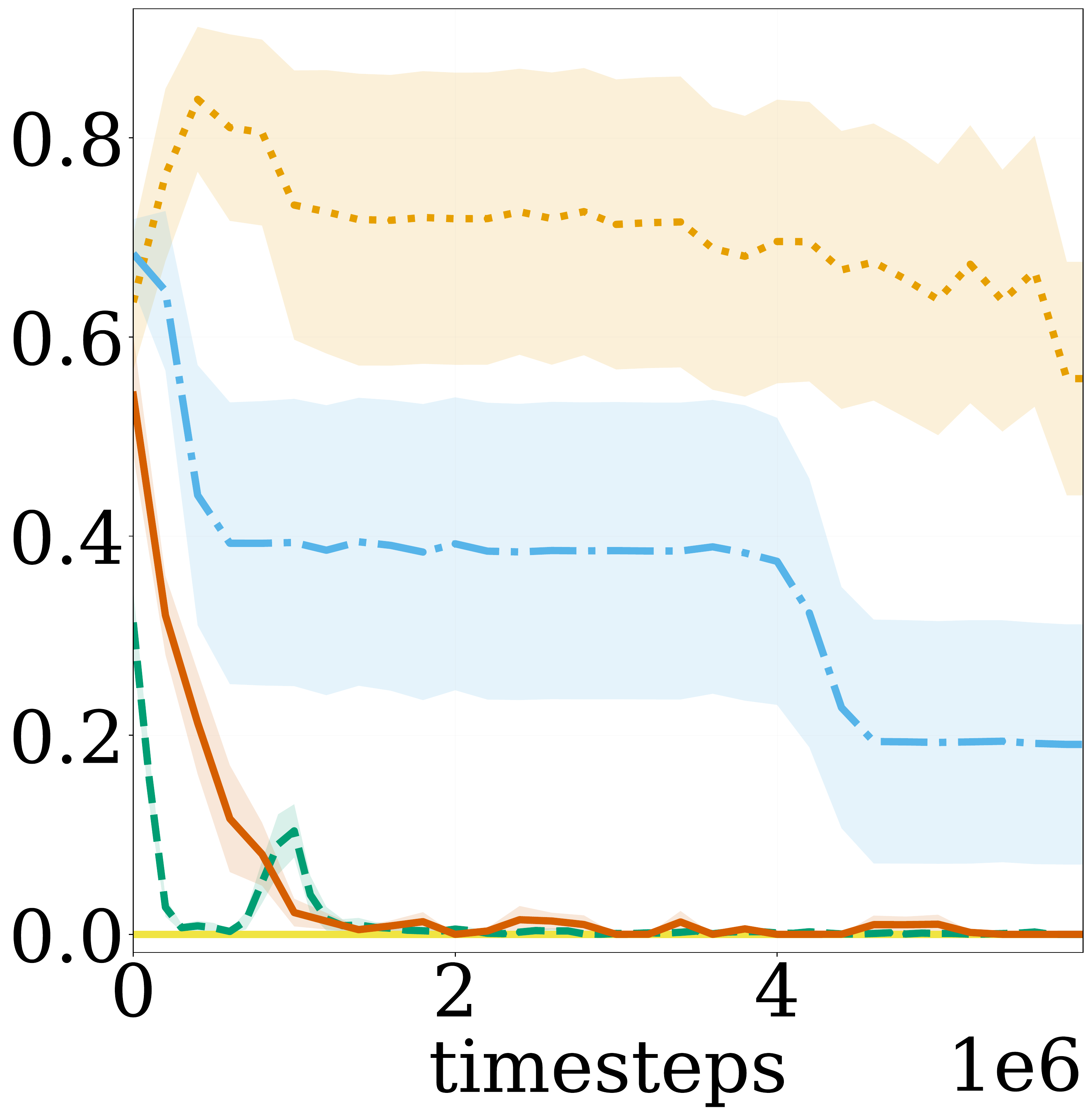}}
    \hspace{0cm}
    \subfigure[Swimmer]{\includegraphics[width=0.16\textwidth]{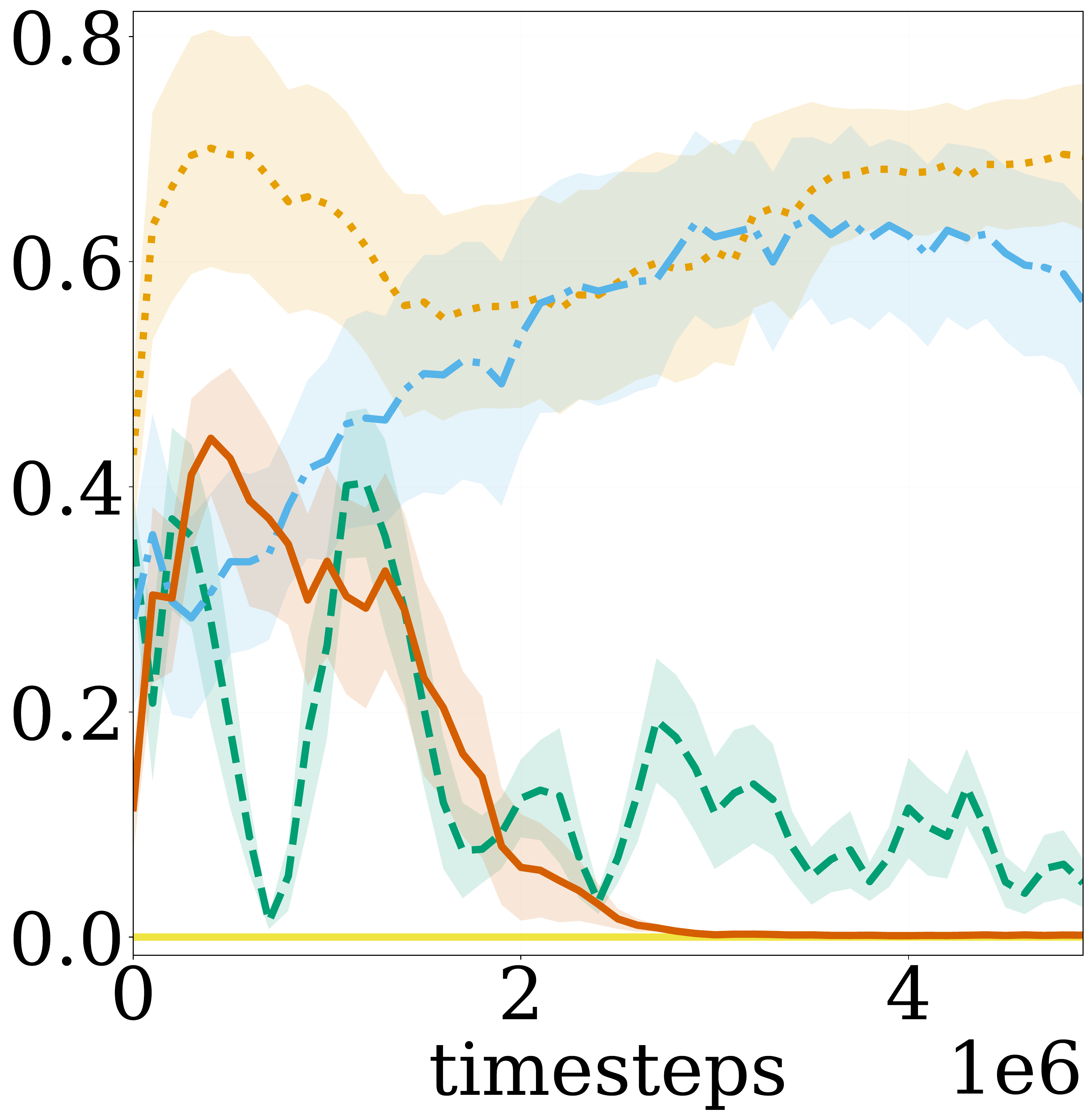}}
    \hspace{0.05cm}
    \subfigure[Walker]{\includegraphics[width=0.16\textwidth]{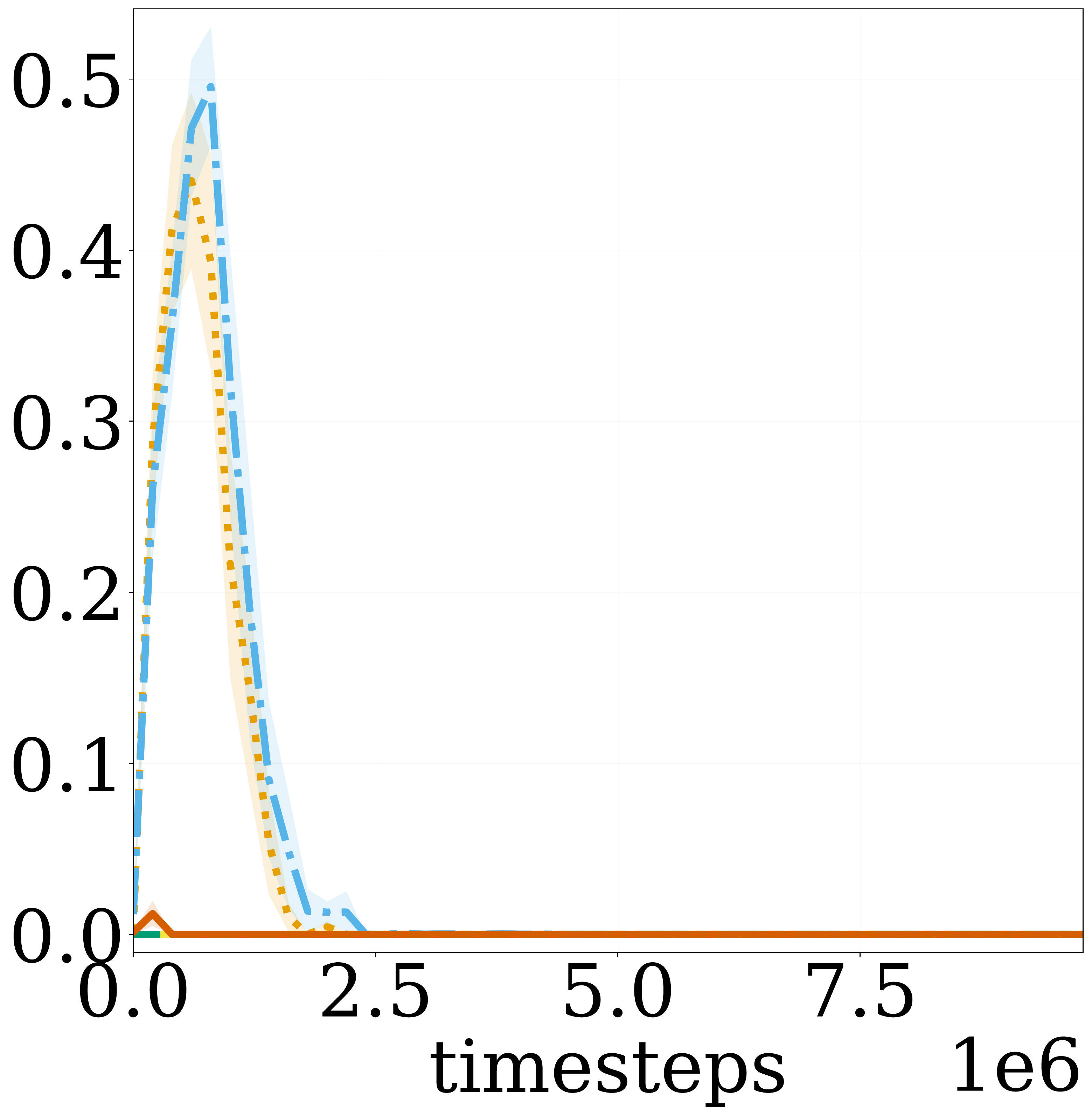}}
    \subfigure[Inverted pendulum]{\includegraphics[width=0.16\textwidth]{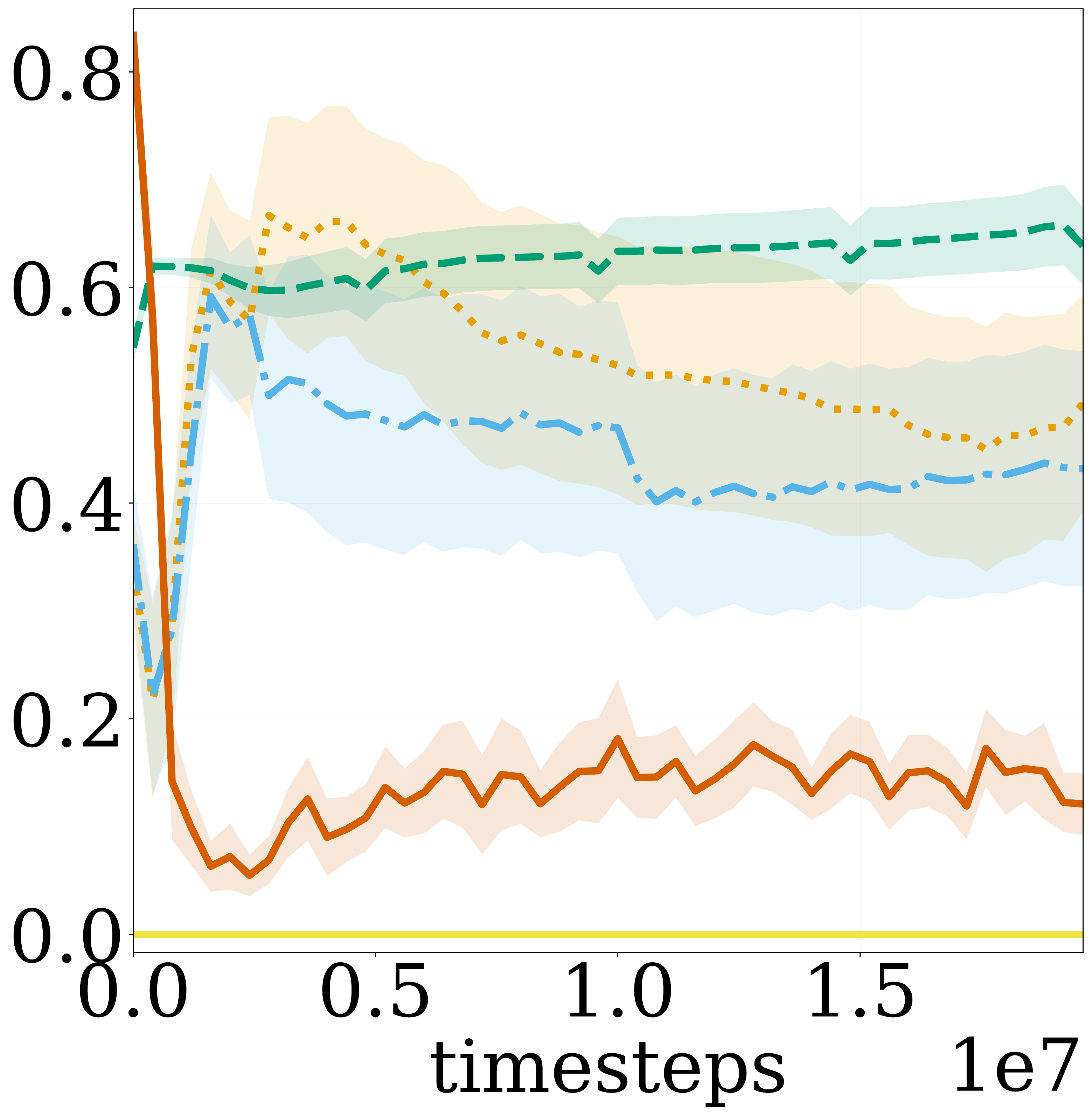}}
    \subfigure{\includegraphics[width=0.6\textwidth]{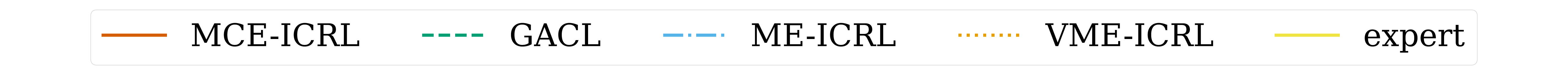}}\\
    \caption{Evaluation of the different methods in the virtual robotics environments: reward (top) and constraint violation rate (bottom) of trajectories sampled from the nominal policy during training. Results are averaged over 10 random seeds. The x-axis is the number of timesteps taken in the environment during training. The shaded regions correspond with the standard error.}
    \label{fig:resultsmujoco}
\end{center}
\vskip -0.2in
\end{figure*}

\subsection{Gridworld}
We designed a gridworld environment where the agent's goal is to travel from a fixed initial location to a fixed goal location while avoiding constrained locations.
A screenshot of this environment is shown in Sec. \ref{sec:env-details} in the appendix (Figure~\ref{fig:gt-constraints-gw}).
Every step the agent receives a reward of $-1$ except when it reaches the goal state, the agent gets a reward of $1$.
Each episode lasts a maximum of 200 timesteps.
We use this environment to examine the influence of stochastic environment dynamics on the performance of learning a cost function and retrieving the optimal policy.
With a particular probability, the environment ignores the action taken by the agent and instead transitions it to a randomly chosen neighboring cell.
We call this probability the stochasticity of the environment.
Figure \ref{fig:gridworld} shows the obtained reward and constraint violations at test time for increasing stochasticity.
We observed a small decrease of reward and small increase of constraint violations for MCE-ICRL for increasing stochasticity.
Non-causal entropy approximates causal entropy for small stochasticity rates but fails when the randomness increases, this is reflected in the results of ME-ICRL and VME-ICRL.
GACL receives always the largest rewards but also at the highest cost because this method was unable to learn a policy which takes the constraints into account.
This supports our claim that IRL methods are not suitable for learning constraints.
In Sec. \ref{sec:exp-results-gw} of the appendix, we report extended results of experiments in the gridworld environment.

\subsection{Virtual Robotics Environments}
The virtual robotic environments are based on the MuJoCo environments from OpenAI Gym \cite{brockman2016openai} but augmented with constraints as proposed by \citet{malik2021inverse} and extended by \citet{liu2022benchmarking}.
We evaluate on five simulated robots: ant, half-cheetah, swimmer and inverted pendulum.
Figure \ref{fig:robotic-env-screenshots} (Sec. \ref{sec:env-details-mujoco} in the appendix) depicts a screenshot of each environment.
This environment is characterized by a continuous state-action space and deterministic dynamics.
The reward of the ant agent is proportional to the distance traveled from the starting position.
The reward of the half-cheetah, swimmer and walker agents is determined by the distance it moves each step.
Because of their morphology, each of these agent can move ``easier'' in one direction than in all other directions, but the ground truth constraints invalidate moving in the ``easy'' direction.
The agent controlling the inverted pendulum receives higher rewards when it is to the left of the starting position.
However, the ground truth cost function assigns high costs to these high reward locations.
The agent should learn to balance the pendulum to the right from the origin receiving lower rewards but also lower costs.
Results are depicted in figure \ref{fig:resultsmujoco}.
In the ant environment MCE-ICRL performs comparable to the other methods.
In the half-cheetah, walker and inverted pendulum domain our method results in higher rewards and less constraint violations.
In the swimmer environment MCE-ICRL results in the lowest cost and higher rewards than ME-ICRL and VME-ICRL. 
GACL obtains higher rewards but with more constraint violations.
For every task, we also plot the reward and constraint violation rate obtained during the expert trajectories.
Note there is often still a gap between the performance of the learning agent and the expert.

\subsection{Realistic Traffic Environment}
\begin{figure}
\vskip 0.2in
\begin{center}
    \hspace{0.01cm}    \subfigure{\includegraphics[width=0.18\textwidth]{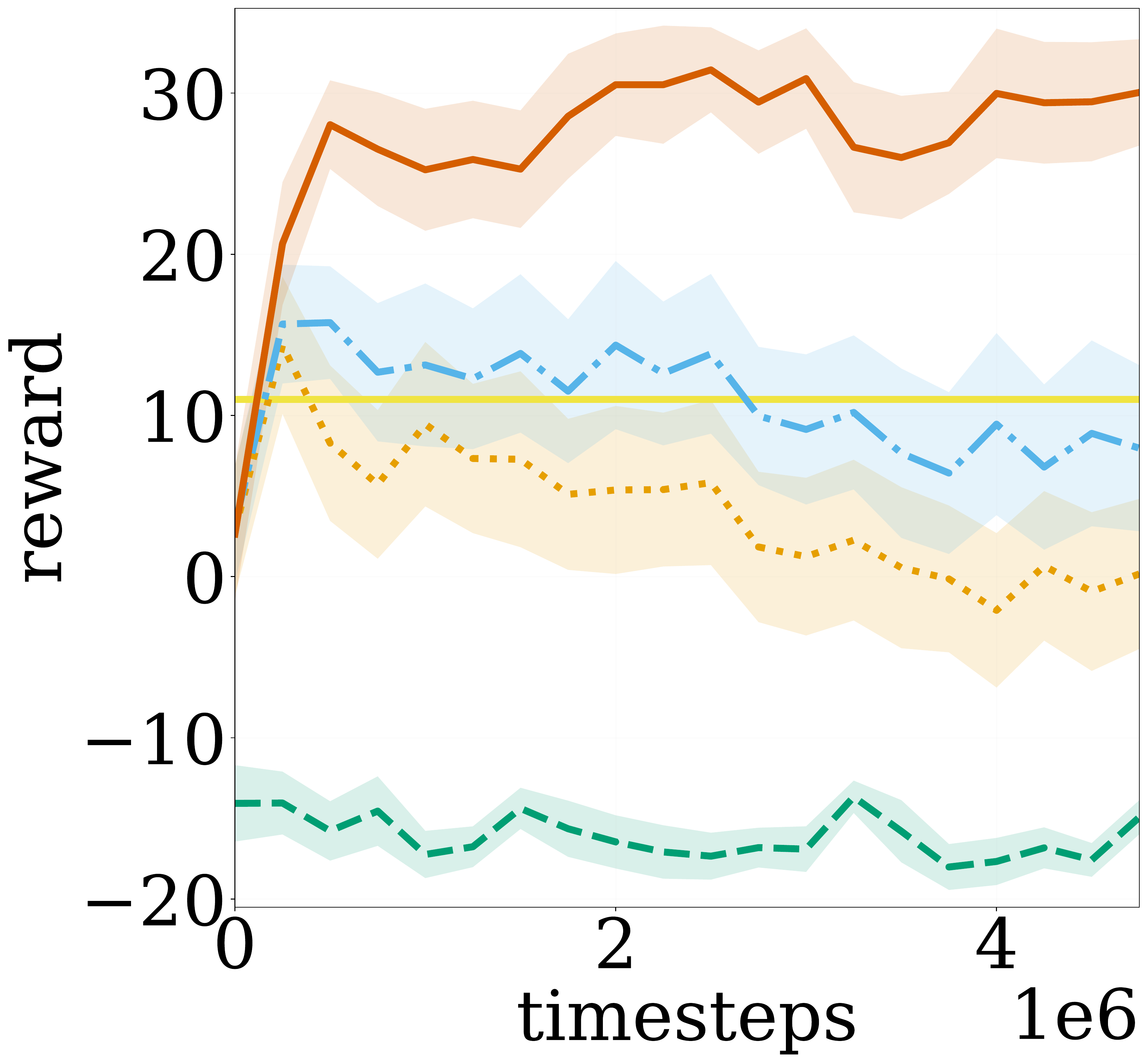}} \hspace{0.2cm}
    \subfigure{\includegraphics[width=0.18\textwidth]{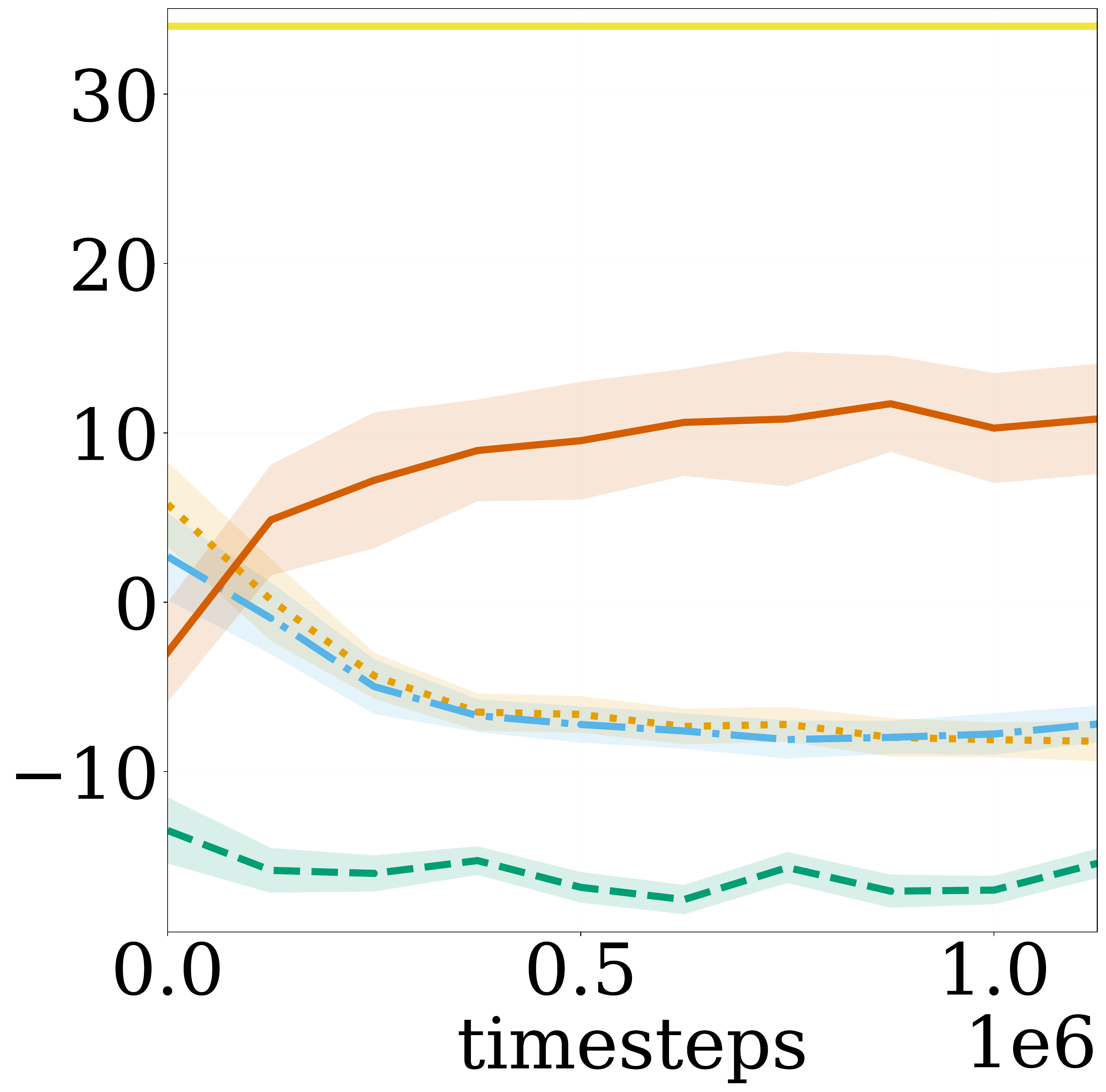}}\\
    \setcounter{subfigure}{0}
    \subfigure[Distance constraint]{\includegraphics[width=0.185\textwidth]{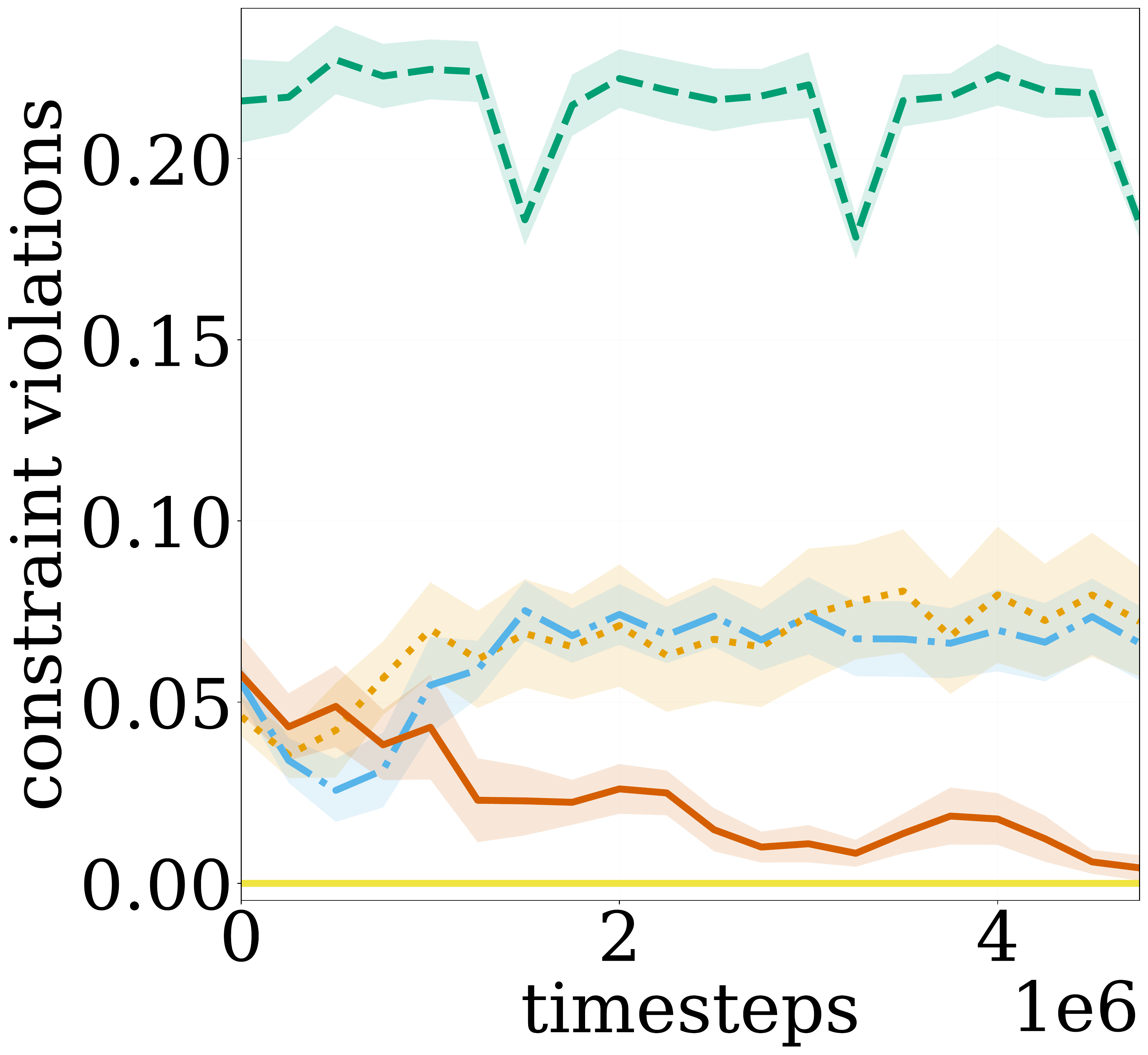}} 
    \hspace{0.12cm}
    \subfigure[Velocity constraint]{\includegraphics[width=0.18\textwidth]{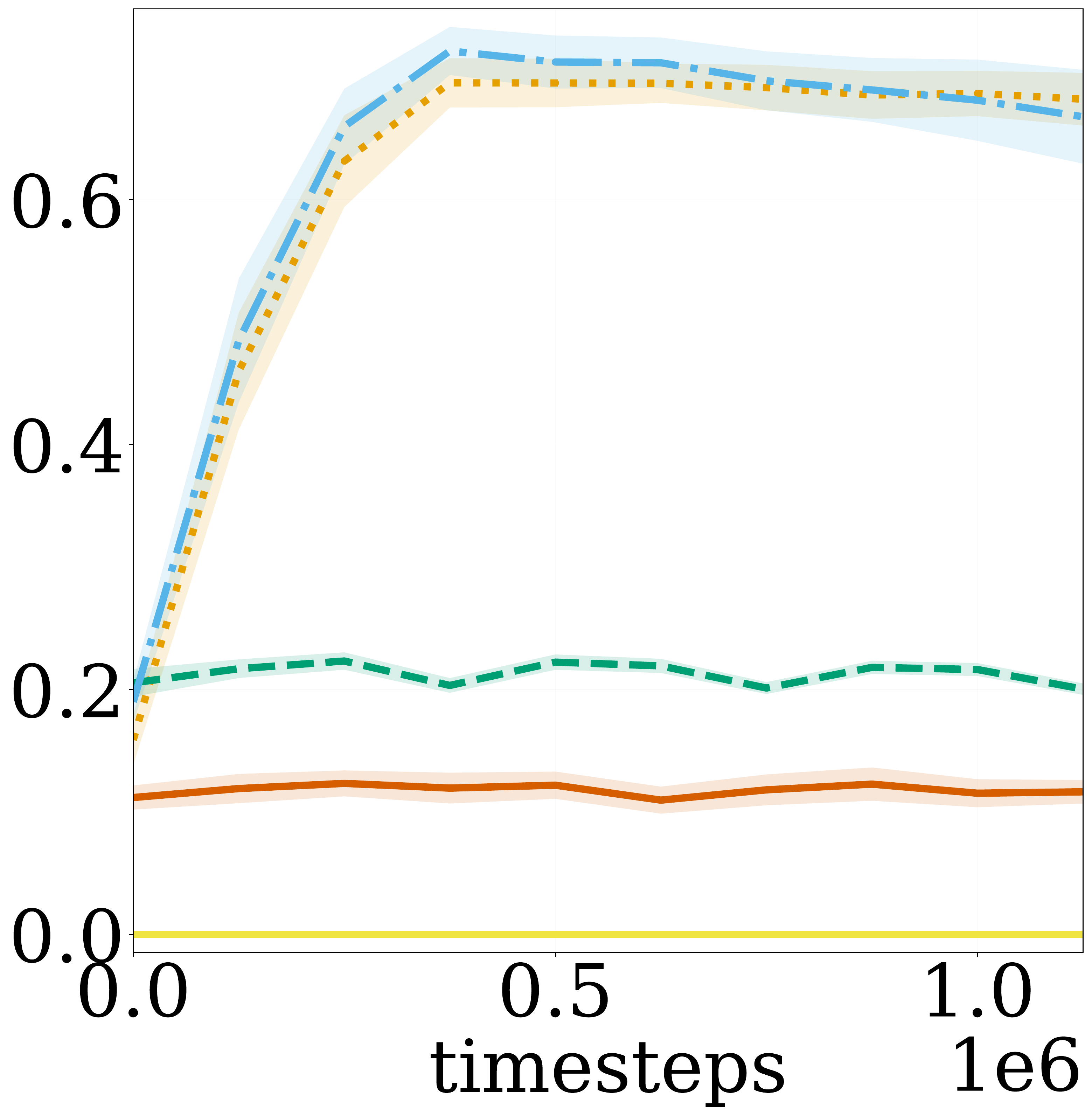}}
    \subfigure{\includegraphics[width=0.5\textwidth]{figures/legend.pdf}}\\
    \caption{Evaluation of the different methods in the realistic traffic environment: reward (top) and constraint violation rate (bottom) of trajectories sampled from the nominal policy during training. Results are averaged over 10 random seeds. The x-axis is the number of timesteps taken in the environment during training. The shaded regions correspond with the standard error.}
    \label{fig:highd}
\end{center}
\vskip -0.3in
\end{figure}
The realistic traffic environment comprises a highway driving task.
This task was proposed by \citet{liu2022benchmarking} and is part of their proposed benchmark.
This environment is constructed from the highD dataset \cite{highDdataset} which is a dataset of naturalistic trajectories of vehicles on German highways.
A scenario and an ego agent are randomly selected from the dataset for control and CommonRoad-RL \cite{commonroadrl} is used to get a state description of the ego car and its surroundings.
This environment is characterized by a continuous state-action space and stochastic dynamics as the behavior of other agents in the environment is unpredictable and different expert agents act differently based on their preferences.
The goal of the agent is to reach its destination at the end of the highway.
Figure \ref{fig:traffic-env} (Sec. \ref{sec:appendix-traffic} in the appendix) visualizes the environment.
The environment contains two scenarios, one with a minimum distance constraint between the ego car and other vehicles and one with a maximum velocity constraint.
The results are depicted in Figure~\ref{fig:highd}.
Our method outperforms the other methods by a large margin which is not surprising as the traffic domain is characterized by high stochasticity.
The ground truth constraints were arbitrarily chosen by \citet{liu2022benchmarking} and are in some cases violated in the starting state of the ego agent.
Because of this, no method, even not the expert agent trained given the ground truth constraints, is able to reduce the cost to zero.

\subsection{Transfer Learning}
\begin{figure}
\vskip 0.2in
\begin{center}
    \hspace{0.01cm}
    \subfigure{\includegraphics[width=0.18\textwidth]{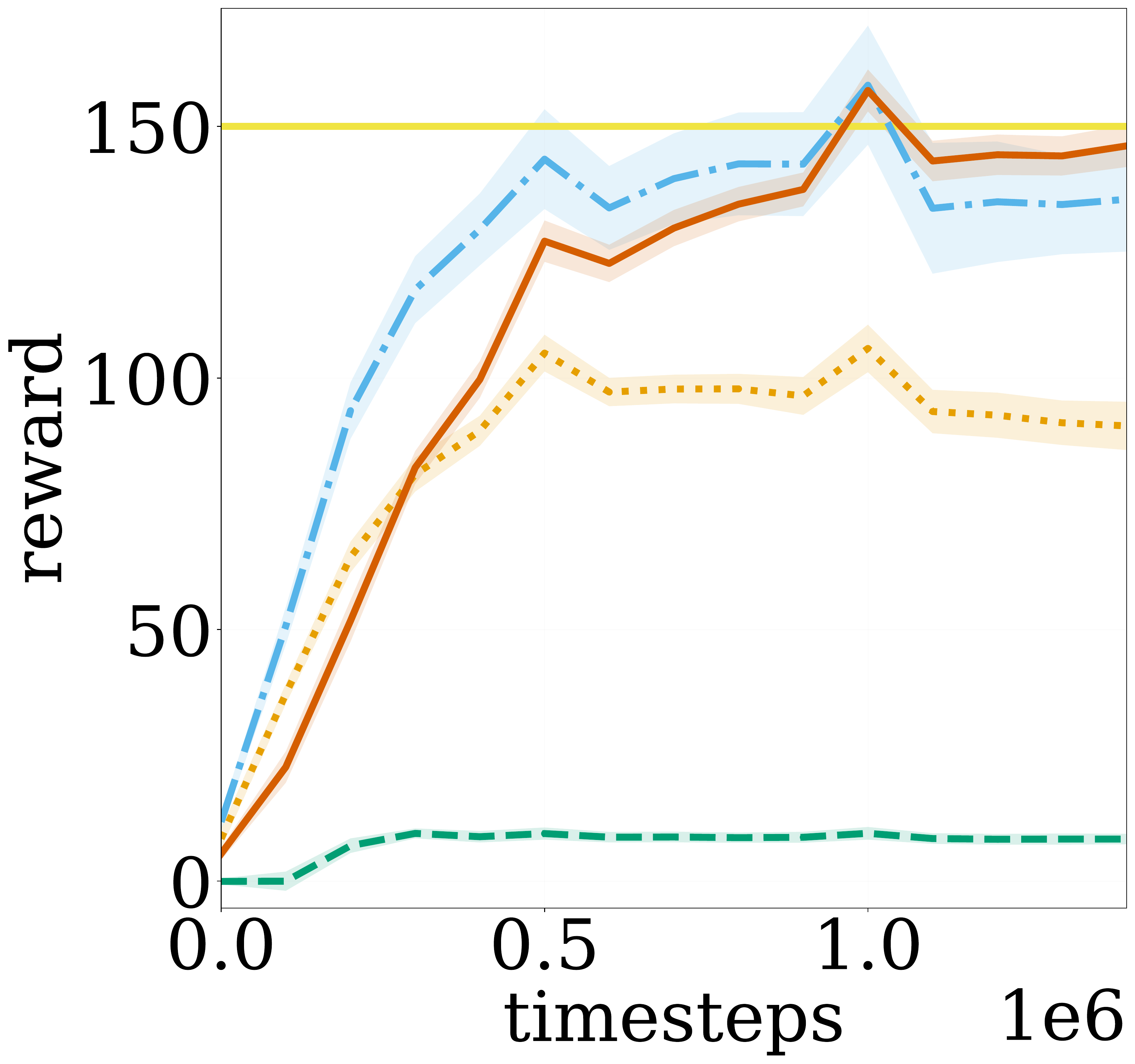}} \hspace{0.2cm}
    \subfigure{\includegraphics[width=0.18\textwidth]{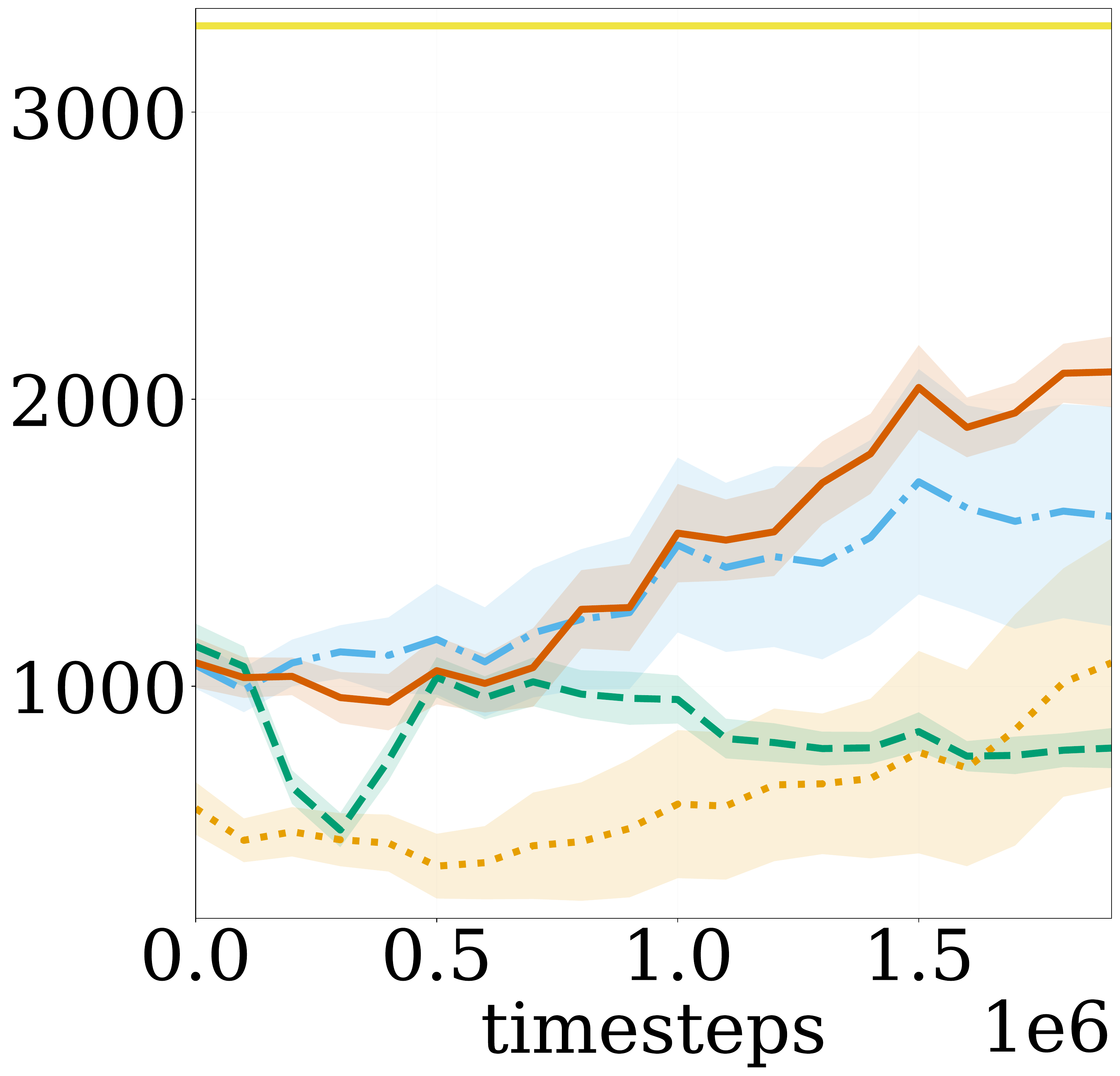}}\\
    \setcounter{subfigure}{0}
    \subfigure[Point]{\includegraphics[width=0.185\textwidth]{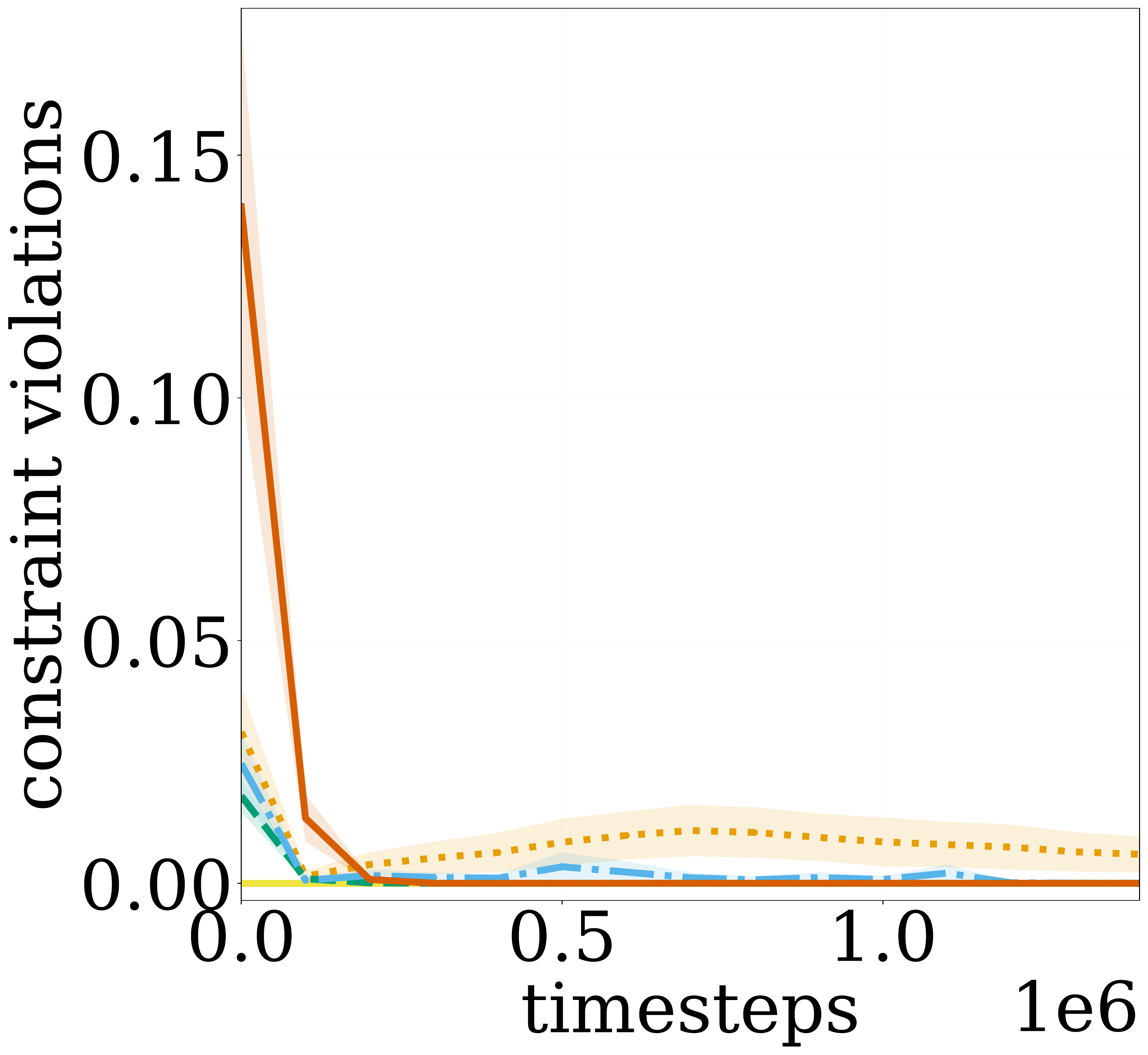}} 
    \hspace{0.12cm}
    \subfigure[Ant broken]{\includegraphics[width=0.18\textwidth]{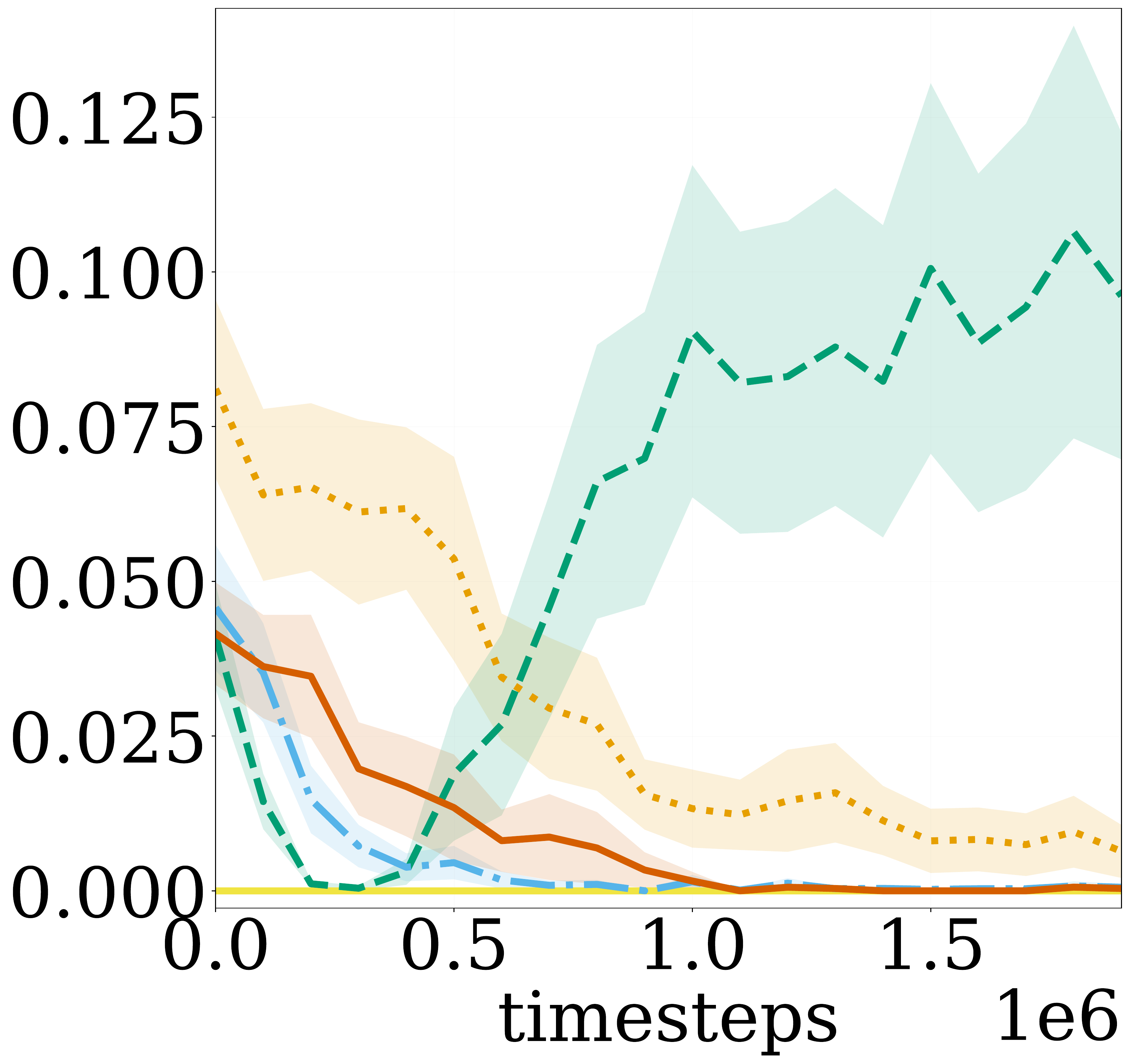}}
    \subfigure{\includegraphics[width=0.5\textwidth]{figures/legend.pdf}}\\
    \caption{Evaluation of the transferability of the cost function obtained by the different methods: reward (top) and constraint violation rate (bottom) of trajectories sampled from the nominal policy during training. Results are averaged over 10 random seeds. The x-axis is the number of timesteps taken in the environment during training. The shaded regions correspond with the standard error.}
    \label{fig:transfer}
\end{center}
\vskip -0.2in
\end{figure}
When a cost function is learned for a specific agent, it could be beneficial when that cost function can be transferred to other types of agents operating in the same environment.
This would prevent us from learning a cost function for every kind of agent.
Note that constraints are often environment specific and apply to different types of agents, e.g. traffic rules apply to all vehicles on the road.
To this end, we assess the transferability of a learned cost function to other types of agents which possibly have other morphologies and reward functions.
We perform the transfer learning experiments proposed by \citet{malik2021inverse}.
The cost function learned for the ant agent is transferred to a ``broken ant'' agent for which two of its legs are disabled.
We also transfer this cost function to the point agent from OpenAI Gym \cite{brockman2016openai}.
The reward function of the point robot encourages the agent to move in counter clockwise in a circle at a distance of 10 units from its initial location.
We adopt reward constrained policy optimization \cite{tessler2018reward} to train agents on the given reward function and the transferred cost function.
The results are depicted in Figure \ref{fig:transfer}.
For both agents our method performs slightly better than ME-ICRL and outperforms VME-ICRL and GACL by a large margin in both reward and number of constraint violations.
%We use the cost function learned in the ant environment to train an agents similar to the ant but with its 

\subsection{Ablation Studies}
\begin{table*}[t]
\caption{Ablation study on the pre-training of the feature encoder and the entropy coefficient $\beta$, the table contains for every agent the received reward and the constraint violation rate at test time $\pm$ standard error, averaged over 10 random seeds and 10 trajectories per seed.}
\hspace{0.1in}
\setlength{\tabcolsep}{4pt}
\centering
\begin{tabular}{lcccccc}
\toprule
& & Ant & Half-cheetah & Swimmer & Walker & Inverted pendulum \\
\hline
\multirow{5}{*}{Reward} & $\beta=1\mathrm{e-}5$ & 5985$\pm$259 & 3895$\pm$295 & 177$\pm$52 & 
 1647$\pm$100 & 27$\pm$3 \\
 & $\beta=1\mathrm{e-}4$ & \textbf{7338$\pm$211} & \textbf{4085$\pm$565} &\textbf{257$\pm$44} & \textbf{1858$\pm$44} & 26$\pm$3 \\
& $\beta=1\mathrm{e-}3$ & 6057$\pm$313 & 3499$\pm$492 & 253$\pm$54 & 1746$\pm$81 & \textbf{32$\pm$3} \\
& $\beta=1\mathrm{e-}2$ & 6228$\pm$316 & 2377$\pm$631 & 192$\pm$43 & 1732$\pm$77 & 29$\pm$4 \\
& no pre-training & 7131$\pm$313 & 3609$\pm$490 & 165$\pm$59 & 1656$\pm$71 & 21$\pm$1\\
\hline
\multirow{5}{*}{\shortstack[l]{Constraint\\ violation\\ rate}} & $\beta=1\mathrm{e-}5$ & \textbf{0$\pm$0} & \textbf{0$\pm$0} & 0.33$\pm$0.09 & \textbf{0$\pm$0} & 0.22$\pm$0.05 \\
& $\beta=1\mathrm{e-}4$ & \textbf{0$\pm$0} & \textbf{0$\pm$0} & \textbf{0.002$\pm$0.001} & \textbf{0$\pm$0} & \textbf{0.11$\pm$0.03} \\
& $\beta=1\mathrm{e-}3$ & \textbf{0$\pm$0} & 0.002$\pm$0.002 & 0.12$\pm$0.04 & \textbf{0$\pm$0} & .12$\pm$0.03 \\
& $\beta=1\mathrm{e-}2$ & \textbf{0$\pm$0} & 0.024$\pm$0.012 & $0.11\pm$0.05 & \textbf{0$\pm$0} & 0.17$\pm$0.02 \\
& no pre-training & \textbf{0$\pm$0} & \textbf{0$\pm$0} & 0.22$\pm$0.09 & \textbf{0$\pm$0} & 0.23$\pm$0.04\\
\bottomrule
\end{tabular}
\label{table:bootstrap}
\end{table*}

\subsubsection{Influence of the Hyperparameter $\beta$}
The entropy coefficient $\beta$ can be interpreted as a regularization coefficient by enforcing randomness into the learned policy $\pi$.
Table \ref{table:bootstrap} reports the reward and the constraint violation rate of trajectories sampled from a policy learned using MCE-ICRL during testing for the virtual robotic environments.
It is self-evident, that a policy which is more random will result in smaller rewards since the policy will be less optimal.
However, when $\beta$ reaches a certain limit, the performance starts to decrease which could be attributed to the agent overfitting on the expert trajectories.

\subsubsection{Pre-Training the Feature Encoder}
To assess the importance of pre-training the feature encoder, we train a version of MCE-ICRL without pre-training.
Table \ref{table:bootstrap} reports the received reward and constraint violation rate of the nominal policy at test time for the virtual robotic environments.
From these results it is clear that pre-training the feature encoder provides a significant increase of the received reward. 
Section~\ref{sec:appendix-more-results} in the appendix provides additional results on the effect of pre-training the feature encoder and $\beta$ during training and for the other environments.

\section{Related Work}
\label{sec:related}
\subsection{Maximum Entropy RL}
In Maximum Entropy RL (MaxEnt RL), the standard maximum reward objective is augmented with a maximum entropy objective \cite{ziebart2008maximum,toussaint2009robot,fox2015taming}.
This augmentation has several advantages, such as improved exploration by acquiring diverse behaviours~\cite{haarnoja2017reinforcement} and better convergence and computational properties~\cite{gu2016q}.
\citet{ziebart2010modeling} extended the principle of maximum entropy to settings were information is revealed over time, i.e. principle of maximum causal entropy.
This extension enables applicability in problems with partial observability, feedback and stochastic influences of the environment.
\subsection{Learning From Experts}
A variety of methods have leveraged human expertise to improve the performance of autonomous agents.
Behavioral cloning uses expert demonstrations to learn a policy using supervised learning \cite{bain1995framework,ross2011reduction}.
Although simple and efficient in some applications, it is possible there will be a mismatch between the training and testing state distribution which could cause a considerable decrease in performance.
In preference learning, the goal is to learn a mapping between inputs and a ranking of the inputs which represents the user's (or expert's) preferences \cite{chu2005preference,lee2021b}.
For humans, it is often easier to provide preferences instead of complete demonstrations although the latter contains more information.
Most closely related to ICRL is IRL which has the goal of learning and subsequently optimizing a reward function \cite{abbeel2004apprenticeship,ho2016generative}.
But as stated in the introduction and confirmed in the experiment section,  IRL cannot be applied for learning constraints.
We refer to the introduction for a review on ICRL literature.

%\section{Related Work}
%\label{sec:related}
%\input{related_work}
%
\section{Conclusion and Future Work}
\label{sec:conclusion}
We proposed a novel ICRL method which is able to learn constraints in environments with stochastic dynamics.
We provide a theoretical proof of convergence in a tabular setting.
Next, we presented an approximation which enables our method to scale to complex problems with a continuous state-action space.
Our method outperforms the state-of-the-art ICRL methods and the learned cost functions can be successfully transferred to other types of agents with different reward functions.
The empirical evaluation shows that our method still does not meet the performance of expert agents in many scenarios, leaving room for improvement.
The ablation study highlights the importance of a well-chosen value for $\beta$, and automatically tuning this parameter as proposed by \citet{haarnoja2018soft} could be a valuable extension to our method.
Another interesting direction is to learn constraints in an offline fashion, i.e. when no nominal model of the environment is available.
Current benchmark domains only consider relatively simple constraints, often specified as location constraints.
Future work is required to develop challenging benchmarks considering more complex constraints, e.g constraints based on the state of multiple objects.


\bibliography{main}
\bibliographystyle{icml2023}

%%%%%%%%%%%%%%%%%%%%%%%%%%%%%%%%%%%%%%%%%%%%%%%%%%%%%%%%%%%%%%%%%%%%%%%%%%%%%%%
%%%%%%%%%%%%%%%%%%%%%%%%%%%%%%%%%%%%%%%%%%%%%%%%%%%%%%%%%%%%%%%%%%%%%%%%%%%%%%%
% APPENDIX
%%%%%%%%%%%%%%%%%%%%%%%%%%%%%%%%%%%%%%%%%%%%%%%%%%%%%%%%%%%%%%%%%%%%%%%%%%%%%%%
%%%%%%%%%%%%%%%%%%%%%%%%%%%%%%%%%%%%%%%%%%%%%%%%%%%%%%%%%%%%%%%%%%%%%%%%%%%%%%%
\newpage
\appendix
\onecolumn
\section{Proofs}
\subsection{Lemma \ref{lemma:optimal-policy}}
\label{sec:proof-tabular}
The primal objective is defined as
\begin{equation}
    \max_{\pi \in \Pi} \mathcal{L}(\pi,\lambda)
\end{equation}
with $\mathcal{L}(\pi,\lambda)$ the Lagrangian defined in eq.~\eqref{eq:lagrangian}, and thus the optimal policy
\begin{equation}
    \pi^{*} = \arg\max_{\pi \in \Pi}\mathcal{L}(\pi,\lambda).
\end{equation}
The simplex $\Pi$ which restricts the valid policies is defined by two constraints (see eq.~\eqref{eq:theta-search-space}).
The first constraint requires that $\pi(a \mid s)$ is positive for all states and timesteps.
We treat this constraint as implicit since the causal entropy is not defined for negative probabilities.
The second constraint requires that $\pi$ is normalized over all actions for all states and timesteps.
We assume, for now, the state-action space is discrete.
Augmenting the normalization constraint on the current objective gives us:
\begin{align}
    \label{eq:primal-with-norm}
    \begin{split}
    & \quad \pi^{*} = \arg\max_{\pi \in \Pi}\mathcal{L}(\pi,\lambda) \\
    & \textrm{s.t.} \: \sum_{t=0}^{T-1} \sum_{s_t \in \mathcal{S}} \left( \sum_{a \in \mathcal{A}} \pi(a \mid s_t) \, - 1 \right) = 0.
    \end{split}
\end{align}
Then, the Lagrangian of the optimization problem in eq.~\eqref{eq:primal-with-norm} is defined as
\begin{equation}
    \label{eq:lag-with-norm}
    \Psi(\pi,\lambda,\mu)  = \mathcal{L}(\pi,\lambda) + \sum_{t=0}^{T-1} \sum_{s_t \in \mathcal{S}} \mu_{s_t} \left( \sum_{a \in \mathcal{A}} \pi(a \mid s) \, - 1 \right) 
\end{equation}
with $\{\mu_{s_t}\}_{s_t \in \mathcal{S}, 0 \leq t \leq T-1}$ the dual variables for the normalization constraint.
Then the optimal policy is
\begin{equation}
    \pi^{*} = \arg\max_{\pi}\Psi(\pi,\lambda,\mu).
\end{equation}
The optimal policy $\pi^{*}$ satisfies the KKT condition:
\begin{equation}
    \triangledown_{\pi} \Psi(\pi^{*}, \lambda, \mu) = 0.
\end{equation}
To obtain an expression for the optimal policy, we take the gradient of the Lagrangian w.r.t. the policy, equate $\triangledown_{\pi}\Psi(\pi,\lambda,\mu)$ to zero and solve for $\pi$.
\begin{lemma}
    \label{lemma:gradient-lag}
    The gradient of the Lagrangian (eq.~\eqref{eq:lag-with-norm}) w.r.t. to the policy $\pi(a_t \mid s_t)$ is given by
    \begin{multline}
        \triangledown_{\pi(a_t \mid s_t)} \, \Psi(\pi,\lambda,\mu) = \mu_{s_t} + \rho_{\pi}(s_t) \bigg(R(s_t, a_t) - \lambda \cdot \phi(s_t,a_t) - \beta \log \pi(a_t \mid s_t) - \beta \\
        + \mathbb{E}_{\pi}\left[ \sum_{t^{\prime}=t+1}^{T-1} \gamma^{t^{\prime}-t} \big(R(S_{t^{\prime}},A_{t^{\prime}}) -\lambda \cdot \phi(S_{t^{\prime}},A_{t^{\prime}}) - \beta \log \pi(A_{t^{\prime}} \mid S_{t^{\prime}})\big) \biggr\rvert {s_t, a_t} \right] \bigg).
    \end{multline}
    Proof.
    We take the gradient of each term of the Lagrangian (eq.~\eqref{eq:lag-with-norm}) w.r.t. to the policy $\pi(a_t \mid s_t)$ separately.
    Uppercase characters $S_t, A_t$ represent random variables while lower case characters $s_t, a_t$ denote individual observations.
    \begin{align}
    & \triangledown_{\pi(a_t \mid s_t)}{\mathbb{E}_{\pi}\left[ 
    R(\tau) \right]} \\
    &= \triangledown_{\pi(a_t \mid s_t)}\mathbb{E}_{\pi}\left[ \sum_{t^{\prime}=0}^{T-1} \gamma^{t^{\prime}} R(S_{t^{\prime}}, A_{t^{\prime}}) \right] \\
    &= \mathbb{E}_{S_t \sim \pi} \left[\triangledown_{\pi(a_t \mid s_t)}{ \mathbb{E}_{\pi} \left[ \sum_{t^{\prime}=0}^{T-1} \gamma^{t^{\prime}} R(S_{t^{\prime}}, A_{t^{\prime}}) \biggr\rvert S_t  \right]} \right] \\
    &= \mathbb{E}_{S_t \sim \pi} \left[\gamma^t \mathbb{I}{[S_t=s_t]} \triangledown_{\pi(a_t \mid s_t)} \mathbb{E}_{\pi}\left[ \sum_{t^{\prime}=t}^{T-1} \gamma^{t^{\prime}-t} R(S_{t^{\prime}}, A_{t^{\prime}}) \biggr\rvert {S_t = s_t} \right] \right] \\
    &= \rho_{\pi}(s_t) \triangledown_{\pi(a_t \mid s_t)} \mathbb{E}_{\pi}\left[ \sum_{t^{\prime}=t}^{T-1} \gamma^{t^{\prime}-t} R(S_{t^{\prime}}, A_{t^{\prime}}) \biggr\rvert {S_t = s_t} \right] \\ 
    &= \rho_{\pi}(s_t) \triangledown_{\pi(a_t \mid s_t)} \left( \pi(a_t \mid s_t) \; \mathbb{E}_{\pi}\left[ \sum_{t^{\prime}=t}^{T-1} \gamma^{t^{\prime}-t} R(S_{t^{\prime}}, A_{t^{\prime}}) \biggr\rvert {S_t = s_t, A_t = a_t} \right] \right. \nonumber\\ & \qquad \left. + \left( \sum_{a^{\prime}_t \neq a_t} \pi(a^{\prime}_t \mid s_t) \right) \mathbb{E}_{\pi}\left[ \sum_{t^{\prime}=t}^{T-1} \gamma^{t^{\prime}-t} R(S_{t^{\prime}}, A_{t^{\prime}}) \biggr\rvert {S_t = s_t, A_t \neq a_t} \right] \right)\\
    &= \rho_{\pi}(s_t) \mathbb{E}_{\pi}\left[ \sum_{t^{\prime}=t}^{T-1} \gamma^{t^{\prime}-t} R(S_{t^{\prime}}, A_{t^{\prime}}) \biggr\rvert {S_t = s_t, A_t = a_t} \right] \\
    &= \rho_{\pi}(s_t) \left[ R(s_t, a_t) + \mathbb{E}_{\pi}\left[ \sum_{t^{\prime}=t+1}^{T-1} \gamma^{t^{\prime}-t} R(S_{t^{\prime}}, A_{t^{\prime}}) \biggr\rvert {s_t, a_t} \right] \right],
    \end{align}
    where $\rho_{\pi}(s_t)$ represents the discounted probability that an agent acting according to policy $\pi$ will be in state $s_t$ at time $t$
    \begin{equation}
        \rho_{\pi}(s_t) = \mathbb{E}_{S_t \sim \pi} \left[\gamma^t \mathbb{I}{[S_t=s_t]}\right].
    \end{equation}
    The gradient of the entropy term:
    \begin{align}
    & \triangledown_{\pi(a_t \mid s_t)}{\mathbb{E}_{\pi}\left[ 
    H(\tau) \right]} \\
    &= \triangledown_{\pi(a_t \mid s_t)}\mathbb{E}_{\pi}\left[ \sum_{t^{\prime}=0}^{T-1} \gamma^{t^{\prime}} H(A_{t^{\prime}} \mid S_{t^{\prime}}) \right] \\
    &= -\triangledown_{\pi(a_t \mid s_t)} \mathbb{E}_{\pi} \left[\sum_{t^{\prime}=0}^{T-1} \gamma^{t^{\prime}} \log \pi(A_{t^{\prime}} \mid S_{t^{\prime}}) \right] \\
    &= -\rho_{\pi}(s_t) \triangledown_{\pi(a_t \mid s_t)} \mathbb{E}_{\pi} \left[\sum_{t^{\prime}=t}^{T-1} \gamma^{t^{\prime}-t} \log \pi(A_{t^{\prime}} \mid S_{t^{\prime}}) \biggr\rvert S_t=s_t \right] \\
    &= -\rho_{\pi}(s_t) \triangledown_{\pi(a_t \mid s_t)} \mathbb{E}_{\pi} \left[\log \pi(A_t \mid S_t) + \sum_{t^{\prime}=t+1}^{T-1} \gamma^{t^{\prime}-t} \log \pi(A_{t^{\prime}} \mid S_{t^{\prime}}) \biggr\rvert S_t=s_t \right] \\
    &= -\rho_{\pi}(s_t) \left(1 + \log \pi(a_t \mid s_t) + \mathbb{E}_{\pi} \left[ \sum_{t^{\prime}=t+1}^{T-1} \gamma^{t^{\prime}-t} \log \pi(A_{t^{\prime}} \mid S_{t^{\prime}}) \biggr\rvert s_t, a_t \right] \right).
    \end{align}
    The gradient of the feature matching term:
    \begin{align}
    & \triangledown_{\pi(a_t \mid s_t)} \lambda \cdot \left(\mathbb{E}_{\mathcal{D}} \left[\sum_{t^{\prime}=0}^{T-1} \gamma^{t^{\prime}} \phi(S_t,A_t)\right] - \mathbb{E}_{\pi}  \left[\sum_{t^{\prime}=0}^{T-1} \gamma^{t^{\prime}} \phi(S_{t^{\prime}},A_{t^{\prime}})\right] \right) \\
    &= -\triangledown_{\pi(a_t \mid s_t)} \lambda \cdot \mathbb{E}_{\pi} \left[\sum_{t^{\prime}=0}^{T-1} \gamma^{t^{\prime}} \phi(S_{t^{\prime}},A_{t^{\prime}})\right] \\
    &= -\rho_{\pi}(s_t) \lambda \cdot \triangledown_{\pi(a_t \mid s_t)}  \mathbb{E}_{\pi} \left[\sum_{t^{\prime}=0}^{T-1} \gamma^{t^{\prime}-t} \phi(S_{t^{\prime}},A_{t^{\prime}}) \biggr\rvert S_t=s_t \right] \\
    &= -\rho_{\pi}(s_t) \lambda \cdot \left[ \phi(s_t, a_t) + \mathbb{E}_{\pi}\left[ \sum_{t^{\prime}=t+1}^{T-1} \gamma^{t^{\prime}-t} \phi(S_{t^{\prime}},A_{t^{\prime}}) \biggr\rvert {s_t, a_t} \right] \right].
    \end{align}
    The gradient of the normalization constraint term:
    \begin{align}
         & \triangledown_{\pi(a_t \mid s_t)} \sum_{t=0}^{T-1} \sum_{s_t \in \mathcal{S}} \mu_{s_t} \left( \sum_{a \in \mathcal{A}} \pi(a \mid s) \, - 1 \right)  \\
         & = \mu_{s_t}.
    \end{align}
\end{lemma}
After re-arranging the gradient from Lemma \ref{lemma:gradient-lag}, we get an expression for the optimal policy $\pi^{*}$
\begin{multline}
    \label{eq:pi-opt}
    \pi^*(a_t \mid s_t) = \exp \bigg(  \dfrac{1}{\beta} \bigg( R(s_t,a_t) - \lambda \cdot \phi(s_t,a_t) + \dfrac{\mu_{s}}{\rho_{\pi}(s_t)} - \beta
    \\ + \mathbb{E}_{\pi}\left[ \sum_{t^{\prime}=t+1}^{T-1} \gamma^{t^{\prime}-t} \big(R(S_{t^{\prime}},A_{t^{\prime}}) -\lambda \cdot \phi(S_{t^{\prime}},A_{t^{\prime}}) - \beta \log \pi(A_{t^{\prime}} \mid S_{t^{\prime}})\big) \biggr\rvert {s_t, a_t} \right] \bigg) \bigg).
\end{multline}
We define $V(s_t)$ and $Q(s_t, a_t)$ as
\begin{align}
    &V(s_t) \triangleq \dfrac{-\mu_{s_t}}{\rho_{\pi}(s_t)} + \beta \\
    &Q(s_t,a_t) \triangleq R(s_t,a_t) - \lambda \cdot \phi(s_t,a_t) + \mathbb{E}_{\pi}\left[ \sum_{t^{\prime}=t+1}^{T-1} \gamma^{t^{\prime}-t} \big(R(S_{t^{\prime}},A_{t^{\prime}}) -\lambda \cdot \phi(S_{t^{\prime}},A_{t^{\prime}}) - \beta \log \pi(A_{t^{\prime}} \mid S_{t^{\prime}}) \biggr\rvert {s_t, a_t} \right].
\end{align}
The optimal policy can then be described as
\begin{equation}
    \pi^*(a_t \mid s_t) = \exp\left(\dfrac{1}{\beta} \big(Q(s_t,a_t) - V(s_t)\big)\right).
\end{equation}
In eq.~\eqref{eq:theta-search-space} we defined the set of valid policies as the set of normalized policies $\Pi$ and assume all values of $\mu$ are chosen such that $\sum_{a \in \mathcal{A}} \pi(a \mid s) = 1$ $\forall s \in \mathcal{S}$.
Because of this, we can say that $\forall s \in \mathcal{S}$:
\begin{align}
   1 &= \sum_{a \in \mathcal{A}} \pi(a \mid s) \\
   &=   \sum_{a \in \mathcal{A}} \exp \left(\dfrac{1}{\beta}(Q(s, a)-V(s))\right) \\
   \exp(\dfrac{1}{\beta}V(s)) &=  \sum_{a \in \mathcal{A}}\exp \left(\dfrac{1}{\beta}(Q(s, a)-V(s))\right) \, \exp(\dfrac{1}{\beta}V(s)) \\
   &= \sum_{a \in \mathcal{A}} \exp\left(\dfrac{1}{\beta}Q(s, a)\right) \\
   V(s) &= \beta \log  \sum_{a \in \mathcal{A}} \exp\left(\dfrac{1}{\beta} Q(s, a)\right).
\end{align}
For a continuous action space we get
\begin{align}
    V(s) &= \beta \log \int \exp\left(\dfrac{1}{\beta} Q(s, a)\right) \, \mathrm{d}a.
\end{align}
Similarly, it can be demonstrated that
\begin{align}
    Q(s_t, a_t) &= R(s_t,a_t) - \lambda \cdot \phi(s_t,a_t) + \mathbb{E}_{\pi}\left[ \sum_{t^{\prime}=t+1}^{T-1} \gamma^{t^{\prime}-t} \big(R(S_{t^{\prime}},A_{t^{\prime}}) -\lambda \cdot \phi(S_{t^{\prime}},A_{t^{\prime}}) - \beta \log \pi(A_{t^{\prime}} \mid S_{t^{\prime}}) \biggr\rvert {s_t, a_t} \right] \\
    &= R(s_t,a_t) - \lambda \cdot \phi(s_t,a_t) + \gamma \, \mathbb{E}_{p} \left[ \mathbb{E}_{\pi}\left[Q(S_{t+1}, A_{t+1}) - \beta \log \pi(A_{t+1} \mid S_{t+1}) \right] \biggr\rvert {s_t, a_t} \right] \\
    &= R(s_t,a_t) - \lambda \cdot \phi(s_t,a_t) + \gamma \,  \mathbb{E}_{p} \left[ \mathbb{E}_{\pi}\left[Q(S_{t+1}, A_{t+1}) - \left(Q(S_{t+1}, A_{t+1}) - V(S_{t+1})\right) \right] \biggr\rvert {s_t, a_t} \right] \\
    &= R(s_t,a_t) - \lambda \cdot \phi(s_t,a_t) + \gamma \, \mathbb{E}_{p} \left[V(S_{t+1}) \biggr\rvert {s_t, a_t} \right].
\end{align}
\subsection{Theorem \ref{theorem:soft-policy-iteration}}
\label{sec:proof-soft-policy-iteration}
First, we will prove the convergence of the policy evaluation step.
\begin{lemma}[Policy Evaluation]
    \label{lemma:soft-policy-evaluation}
    Starting with an initial Q-function $Q^0: \mathcal{S} \times \mathcal{A} \rightarrow \mathbb{R}$ with $|\mathcal{A}| < \infty$, where $Q^{k+1} = \mathcal{T}^{\pi}Q^k$.
    Then the sequence $Q^k$ will converge to the Q-value of $\pi$ as $k \rightarrow \infty$. \\
    \\
    Proof. \\
    We define an augmented reward signal as 
        \begin{equation}
        R_{\pi}(s_t,a_t) \triangleq R(s_t,a_t) - \lambda \cdot \phi(s_t,a_t) + \beta H(a_t \mid s_t).
    \end{equation}
    Applying the standard convergence results for policy evaluation \cite{sutton2018reinforcement} proofs the lemma.
    It is necessary to have $|\mathcal{A}| < \infty$ in order to ensure that the augmented reward is bounded.
\end{lemma}
For the projection presented in eq.~\eqref{eq:kl}, we can prove that the new, projected policy has a higher value than the old policy according to the objective in eq.~\eqref{eq:lagrangian}.
\begin{lemma}[Policy Improvement]
    \label{lemma:soft-policy-improvement}
    Given $\pi_{\textrm{old}} \in \Pi$ and $\pi_{\textrm{new}}$ be the policy obtained by solving the minimization problem defined in eq.~\eqref{eq:kl}.
    Then $Q^{\pi_{\textrm{new}}}(s_t, a_t) \geq Q^{\pi_{\textrm{old}}}(s_t, a_t)$ for all $(s_t,a_t) \in \mathcal{S} \times \mathcal{A}$ with $|\mathcal{A}| < \infty.$ \\
    \\
    Proof.\\
    Let $Q^{\pi_{\textrm{old}}}$ and $V^{\pi_{\textrm{old}}}$ be the action-value and state-value function corresponding to $\pi_{\textrm{old}} \in \Pi$.
We can define $J_{\pi_{\textrm{old}}}$ as
\begin{equation}
    J_{\pi_{\textrm{old}}}\big(\pi^{\prime}(\cdot \mid s_t)\big) = \textrm{D}_{\textrm{KL}}\left(\pi^{\prime} \mid \mid \exp(\dfrac{1}{\beta}\left(Q^{\pi_{\textrm{old}}}(s_t \mid \cdot)-V^{\pi_{\textrm{old}}}(s_t)\right))\right).
\end{equation}
Then $\pi_{\textrm{new}}$ can be defined as
\begin{equation}
    %\pi_{\textrm{new}}(\cdot \mid s_t) &= \arg\min_{\pi^{\prime} \in \Pi} \textrm{D}_{\textrm{KL}}\big(\pi^{\prime} \mid \mid \exp(Q^{\pi_{\textrm{old}}}(s_t \mid \cdot)-V^{\pi_{\textrm{old}}}(s_t)\big) \\
    \pi_{\textrm{new}}(\cdot \mid s_t) = \arg\min_{\pi^{\prime} \in \Pi} J_{\pi_{\textrm{old}}}\big(\pi^{\prime}(\cdot \mid s_t)\big).
\end{equation}
The inequality $J_{\pi_{\textrm{old}}}\big(\pi_{\textrm{new}}(\cdot \mid s_t)\big) \leq J_{\pi_{\textrm{old}}}\big(\pi_{\textrm{old}}(\cdot \mid s_t)\big)$ must hold, as we have the option of always selecting $\pi_{\textrm{new}} = \pi_{\textrm{old}}$ from the set of policies $\Pi$.
Hence
\begin{align}
    \mathbb{E}_{\pi_{\textrm{new}}} \left[\log \pi_{\textrm{new}} (a_t \mid s_t) - \dfrac{1}{\beta}(Q^{\pi_{\textrm{old}}}(s_t,a_t) - V^{\pi_{\textrm{old}}}(s_t)) \right] &\leq \mathbb{E}_{\pi_{\textrm{old}}} \left[\log \pi_{\textrm{old}} (a_t \mid s_t) - \dfrac{1}{\beta}(Q^{\pi_{\textrm{old}}}(s_t,a_t) - V^{\pi_{\textrm{old}}}(s_t)) \right] \\
    \mathbb{E}_{\pi_{\textrm{new}}} \left[\log \pi_{\textrm{new}} (a_t \mid s_t) - \dfrac{1}{\beta}Q^{\pi_{\textrm{old}}}(s_t,a_t) \right] + \dfrac{1}{\beta}V^{\pi_{\textrm{old}}}(s_t) &\leq \mathbb{E}_{\pi_{\textrm{old}}} \left[\log \pi_{\textrm{old}} (a_t \mid s_t) - \dfrac{1}{\beta}Q^{\pi_{\textrm{old}}}(s_t,a_t) \right] + \dfrac{1}{\beta}V^{\pi_{\textrm{old}}}(s_t) \\
    \mathbb{E}_{\pi_{\textrm{new}}} \left[\log \pi_{\textrm{new}} (a_t \mid s_t) - \dfrac{1}{\beta}Q^{\pi_{\textrm{old}}}(s_t,a_t) \right]  &\leq \mathbb{E}_{\pi_{\textrm{old}}} \left[\log \pi_{\textrm{old}} (a_t \mid s_t) - \dfrac{1}{\beta}Q^{\pi_{\textrm{old}}}(s_t,a_t) \right] \\
    \mathbb{E}_{\pi_{\textrm{new}}} \left[Q^{\pi_{\textrm{old}}}(s_t,a_t) - \beta \log \pi_{\textrm{new}} (a_t \mid s_t)\right]  &\geq  V^{\pi_{\textrm{old}}}(s_t). 
    \label{eq:v-bound}
\end{align}
We can bring $V^{\pi_{\textrm{old}}}$ outside of the expectation since the state-value function only depends on the current state (line 2).
The last line follows from the definition of the optimal policy in eq.~\eqref{eq:opt-policy}. We can repeatedly expand the Bellman equation by applying the Bellman equation and the inequality of eq.~\eqref{eq:v-bound}:
\begin{align}
    Q^{\pi_{\textrm{old}}}(s_t,a_t) &= R(s_t,a_t) - \lambda \cdot \phi(s_t,a_t) + \gamma \, \mathbb{E}_{p} \left[V^{\pi_{\textrm{old}}}(s_{t+1})\right] \\
    &\leq R(s_t,a_t) - \lambda \cdot \phi(s_t,a_t) + \gamma \, \mathbb{E}_{p, \pi_{\textrm{new}}} \left[Q^{\pi_{\textrm{old}}}(s_{t+1}, a_{t+1}) - \beta \log \pi_{\textrm{new}}(a_{t+1} \mid s_{t+1}) \right] \\
    &\vdots \\
    &\leq Q^{\pi_{\textrm{new}}}(s_{t}, a_{t}).
\end{align}
Then, the convergence to $Q^{\pi_{\textrm{new}}}$ follows from Lemma \ref{lemma:soft-policy-evaluation}.
\end{lemma}
At each iteration $i$, the policy $\pi_i$ will reach an optimum due to Lemma \ref{lemma:soft-policy-improvement}. 
Because the sequence $Q^{\pi_{i}}$ increases monotonically and is bounded for $\pi \in \Pi$, the sequence converges to some $\pi^*$.
We have to prove that $\pi^*$ is indeed optimal.
At convergence, it must be the case that $J_{\pi^*}(\pi^*(\cdot \mid s_t)) < J_{\pi^*}(\pi(\cdot \mid s_t))$ for all $\pi \in \Pi$, $\pi \neq \pi^*$. By repeatedly expanding $Q^{\pi^*}$ as done in Lemma \ref{lemma:soft-policy-improvement}, we can show that $Q^{\pi^*}(s_t,a_t) > Q^{\pi}(s_t, a_t)$ for all $(s_t,a_t) \in \mathcal{S} \times \mathcal{A}$.
Thus, the value of any other policy in $\Pi$ is lower than that of the converged policy.
This makes $\pi^*$ optimal in $\Pi$.

\subsection{Lemma \ref{lemma:dual-convex}}
\label{sec:proofs-dual-convex}
$\max_{\pi \in \Pi}\mathcal{L}(\pi,\lambda)$ can be viewed as a pointwise maximum of affine functions of $\lambda$, thus is concave. 
$\lambda \geq 0$ is an affine constraint.
Hence, the dual problem (eq.~\eqref{eq:dual}) is a convex optimization problem.

\subsection{Lemma \ref{lemma:gradient-theta-objective}}
\label{sec:proof-lag}
We calculate the gradient of each term of the Lagrangian eq.~\eqref{eq:lagrangian} separately w.r.t. the policy parameters $\theta$.
The gradient of the causal entropy term is
\\
\begin{align}
     &\triangledown_{\theta} \mathbb{E}_{\tau \sim \pi_{\theta}} \big[H(\tau) \big] \\
     &= \triangledown_{\theta} \mathbb{E}_{\tau \sim \pi_{\theta}} \bigg[\sum_{t=0}^{T-1} \gamma^t H(a_t \mid s_t) \bigg] \\
     &= -\triangledown_{\theta} \mathbb{E}_{\tau \sim \pi_{\theta}} \bigg[\sum_{t=0}^{T-1} \gamma^t \log \pi_{\theta} (a_t \mid s_t) \bigg]  \\
     &= -\triangledown_{\theta} \sum_{\tau} \bigg( P(\tau \mid \pi_{\theta})\sum_{t=0}^{T-1} \gamma^t \log \pi_{\theta} (a_t \mid s_t)\bigg) \\
     &= -\sum_{\tau} \bigg( \triangledown_{\theta}P(\tau \mid \pi_{\theta})\sum_{t=0}^{T-1} \gamma^t \log \pi_{\theta} (a_t \mid s_t) + P(\tau \mid \pi_{\theta})\sum_{t=0}^{T-1} \gamma^t \triangledown_{\theta}\log \pi_{\theta} (a_t \mid s_t) \bigg) \\
     &= -\sum_{\tau} \bigg( P(\tau \mid \pi_{\theta}) \triangledown_{\theta} \log P(\tau \mid \pi_{\theta})\sum_{t=0}^{T-1} \gamma^t \log \pi_{\theta} (a_t \mid s_t) + P(\tau \mid \pi_{\theta})\sum_{t=0}^{T-1} \gamma^t \triangledown_{\theta}\log \pi_{\theta} (a_t \mid s_t) \bigg) \\
     &= -\sum_{\tau} P(\tau \mid \pi_{\theta}) \bigg( \triangledown_{\theta} \log P(\tau \mid \pi_{\theta})\sum_{t=0}^{T-1} \gamma^t \log \pi_{\theta} (a_t \mid s_t) + \sum_{t=0}^{T-1} \gamma^t \triangledown_{\theta} \log \pi_{\theta} (a_t \mid s_t) \bigg) \\
     &= -\mathbb{E}_{\tau \sim \pi_{\theta}}\left[ \triangledown_{\theta} \log P(\tau \mid \pi_{\theta})\sum_{t=0}^{T-1} \gamma^t \log \pi_{\theta} (a_t \mid s_t) + \sum_{t=0}^{T-1} \gamma^t \triangledown_{\theta}\log \pi_{\theta} (a_t \mid s_t) \right] \\
     &= -\mathbb{E}_{\tau \sim \pi_{\theta}}\left[ \sum_{t=0}^{T-1} \triangledown_{\theta} \log \pi_{\theta}(a_t \mid s_t)\sum_{t^{\prime}=t}^{T-1} \gamma^{t^{\prime} - t}\log \pi_{\theta} (a_{t^{\prime}} \mid s_{t^{\prime}}) + \sum_{t=0}^{T-1} \gamma^t \triangledown_{\theta} \log \pi_{\theta} (a_t \mid s_t) \right] \\
     &= -\mathbb{E}_{\tau \sim \pi_{\theta}}\left[ \sum_{t=0}^{T-1} \triangledown_{\theta} \log \pi_{\theta}(a_t \mid s_t) \left( \sum_{t^{\prime}=t}^{T-1} \gamma^{t^{\prime} - t} \log \pi_{\theta} (a_{t^{\prime}} \mid s_{t^{\prime}}) + 1 \right) \right].
\end{align}
In line 2 and 3, we apply the definition of causal entropy.
$P(\tau \mid \pi)$ denotes the probability of a trajectory $\tau$ under policy $\pi$.
Line 6 follows from $\triangledown P(\tau \mid \pi) = P(\tau \mid \pi) \triangledown \log P(\tau \mid \pi)$.
Line 9 follows from the definition of the probability of a trajectory $\tau$ under a policy $\pi$: $P(\tau \mid \pi) = \mathcal{I}(s_0)\prod_{t=0}^{T-1} p(s_{t+1} \mid s_t, a_t) \pi(a_t \mid s_t)$.
Then the second sum needs to be updated since the policy at time $t^{\prime}$ cannot affect the policy at time $t$ when $t < t^{\prime}$ because of the causal relation between consecutive states.
The gradient of the reward term can be calculated similarly
\begin{align}
     &\triangledown_{\theta} \mathbb{E}_{\tau \sim \pi_{\theta}} \big[R(\tau) \big] \\
     &= \mathbb{E}_{\tau \sim \pi_{\theta}} \bigg[\sum_{t=0}^{T-1}\triangledown_{\theta} \log \pi(a_t \mid s_t) \sum_{t^{\prime}=t}^{T-1} \gamma^{t^{\prime} - t} R(s_t,a_t) \bigg],
\end{align}
and the feature matching term:
\begin{align}
     &\triangledown_{\theta} \lambda \cdot \left( \mathbb{E}_{\tau \sim \mathcal{D}}  \left[\sum_{t=0}^{T-1} \gamma^{t} \phi(s_{t},a_{t})\right] - \mathbb{E}_{\tau \sim \pi_{\theta}} \left[\sum_{t=0}^{T-1} \gamma^{t} \phi(s_t,a_t)\right] - \alpha \right) \\
     &= -\triangledown_{\theta} \lambda \cdot \mathbb{E}_{\tau \sim \pi_{\theta}} \left[\sum_{t=0}^{T-1} \gamma^{t} \phi(s_t,a_t)\right]
     \\
     &= -\mathbb{E}_{\tau \sim \pi_{\theta}} \bigg[\sum_{t=0}^{T-1}\triangledown_{\theta} \log \pi(a_t \mid s_t) \sum_{t^{\prime}=t}^{T-1} \gamma^{t^{\prime} - t} \lambda \cdot \phi(s_t,a_t) \bigg].
\end{align}
In our work, the same discount factor is used to discount rewards, costs and entropies.
Putting it all together, we get
\begin{equation}
    \label{eq:gradient-before-rewrite}
    \triangledown_{\theta} \mathcal{L}(\theta, \lambda) = \mathbb{E}_{\tau \sim \pi_{\theta}} \bigg[\sum_{t=0}^{T-1}\triangledown_{\theta} \log \pi_{\theta}(a_t \mid s_t) \sum_{t^{\prime}=t}^{T-1}  \gamma^{t^{\prime}-t} \big( R(s_{t^{\prime}},a_{t^{\prime}}) - \lambda \cdot \phi(s_{t^{\prime}}, a_{t^{\prime}}) - \beta \log \pi_{\theta}(a_{t^{\prime}} \mid s_{t^{\prime}}) - \beta \big) \bigg].
\end{equation}
We rewrite eq.~\eqref{eq:gradient-before-rewrite}
\begin{multline}
     \triangledown_{\theta} \mathcal{L}(\theta, \lambda) = \mathbb{E}_{\tau \sim \pi_{\theta}} \bigg[\sum_{t=0}^{T-1}\triangledown_{\theta} \log \pi_{\theta}(a_t \mid s_t) \biggr( R(s_{t},a_{t}) - \lambda \cdot \phi(s_{t}, a_{t}) \\ + \sum_{t^{\prime}=t+1}^{T-1}  \gamma^{t^{\prime}-t} \big( R(s_{t^{\prime}},a_{t^{\prime}}) - \lambda \cdot \phi(s_{t^{\prime}}, a_{t^{\prime}}) - \beta \log \pi_{\theta} (a_{t^{\prime}} \mid s_{t^{\prime}})\big) - \beta \log \pi_{\theta}(a_t \mid s_t) - \sum_{t^{\prime}=t}^{T-1}\gamma^{t^{\prime}-t}\beta \biggr) \bigg].     
     %\sum_{t^{\prime}=t}^{T-1}  \gamma^{t^{\prime}-t} \big( R(s_{t^{\prime}},a_{t^{\prime}}) - \lambda \, \phi(s_{t^{\prime}}, a_{t^{\prime}}) - \beta \log \pi_{\theta} (a_{t^{\prime}} \mid s_{t^{\prime}})\big) - \beta \bigg].
\end{multline}
Using the definition of the action-value function (eq.~\eqref{eq:q}) we obtain:
\begin{equation}
    \triangledown_{\theta} \mathcal{L}(\theta, \lambda) = \mathbb{E}_{\tau \sim \pi_{\theta}} \bigg[\sum_{t=0}^{T-1}\triangledown_{\theta} \log \pi_{\theta}(a_t \mid s_t) \big(Q(s_t, a_t) - \beta \log \pi_{\theta}(a_t \mid s_t) - \sum_{t^{\prime}=t}^{T-1}\gamma^{t^{\prime}-t}\beta \big) \bigg].
\end{equation}

\subsection{Lemma \ref{thm:baseline}}
\label{sec:proofs-baseline}
We can replace $\sum_{t^{\prime}=t}^{T-1}\gamma^{t^{\prime}-t}\beta$ with a state dependent baseline $b(s_t)$ in eq.~\eqref{eq:lag_theta} only when this does not affect the gradient $\triangledown_{\theta} \mathcal{L}(\theta, \lambda)$. 
We rewrite eq.~\eqref{eq:lag_theta}:
\begin{align}
    \triangledown_{\theta} \mathcal{L}(\theta, \lambda) &= \mathbb{E}_{\tau \sim \pi_{\theta}} \bigg[\sum_{t=0}^{T-1}\triangledown_{\theta} \log \pi_{\theta}(a_t \mid s_t) \big(Q(s_t, a_t) - \beta \log \pi_{\theta}(a_t \mid s_t) - \sum_{t^{\prime}=t}^{T-1}\gamma^{t^{\prime}-t}\beta \big) \bigg] \\
    &= \mathbb{E}_{\tau \sim \pi_{\theta}} \bigg[\sum_{t=0}^{T-1}\triangledown_{\theta} \log \pi_{\theta}(a_t \mid s_t) \big(Q(s_t, a_t) - \beta \log \pi_{\theta}(a_t \mid s_t)\big) \bigg]  \\
    & \qquad \qquad - \mathbb{E}_{\tau \sim \pi_{\theta}} \bigg[\sum_{t=0}^{T-1} \triangledown_{\theta} \log \pi(a_t \mid s_t) \sum_{t^{\prime}=t}^{T-1}\gamma^{t^{\prime}-t}\beta \bigg]
\end{align}
In the second term, we replace $\sum_{t^{\prime}=t}^{T-1}\gamma^{t^{\prime}-t}\beta$ with a state-dependent baseline $b(s_t)$ which gives us:
\begin{align}
    \triangledown_{\theta} \mathcal{L}(\theta, \lambda)
    &= \mathbb{E}_{\tau \sim \pi_{\theta}} \bigg[\sum_{t=0}^{T-1}\triangledown_{\theta} \log \pi_{\theta}(a_t \mid s_t) \big(Q(s_t, a_t) - \beta \log \pi_{\theta}(a_t \mid s_t)\big) \bigg]\\
    & \qquad \qquad - \mathbb{E}_{\tau \sim \pi_{\theta}} \bigg[ \sum_{t=0}^{T-1} \triangledown_{\theta} \log \pi(a_t \mid s_t) b(s_{t}) \bigg]
\end{align}
Since the introduced baseline only affects the second term it suffices to proof that 
\begin{equation}
    \mathbb{E}_{\tau \sim \pi_{\theta}} \bigg[\sum_{t=0}^{T-1} \triangledown_{\theta} \log \pi(a_t \mid s_t) \sum_{t^{\prime}=t}^{T-1}\gamma^{t^{\prime}-t}\beta \bigg] = \mathbb{E}_{\tau \sim \pi_{\theta}} \bigg[\sum_{t=0}^{T-1}\triangledown_{\theta} \log \pi(a_t \mid s_t) b(s_{t}) \bigg]
\end{equation}
\begin{align}
    & \mathbb{E}_{\tau \sim \pi_{\theta}}\bigg[\sum_{t=0}^{T-1}\triangledown_{\theta} \log \pi_{\theta}(a_t \mid s_t) b(s_t) \bigg] \\
    &= \mathbb{E}_{s_{0:T-1},a_{0:T-1}}\bigg[\sum_{t=0}^{T-1}\triangledown_{\theta} \log \pi_{\theta}(a_t \mid s_t) b(s_t) \bigg] \\
    &= \mathbb{E}_{s_{0:t},a_{0:t-1}}\bigg[\sum_{t=0}^{T-1}\mathbb{E}_{s_{t+1:T-1},a_{t:T-1}} \big[\triangledown_{\theta} \log \pi_{\theta}(a_t \mid s_t) b(s_t) \big] \bigg] \\
    &= \mathbb{E}_{s_{0:t},a_{0:t-1}}\bigg[\sum_{t=0}^{T-1} b(s_t) \, \mathbb{E}_{s_{t+1:T-1},a_{t:T-1}} \big[\triangledown_{\theta} \log \pi_{\theta}(a_t \mid s_t)\big] \bigg] \\
    &= \mathbb{E}_{s_{0:t},a_{0:t-1}}\bigg[\sum_{t=0}^{T-1} b(s_t) \,\mathbb{E}_{a_t} \big[\triangledown_{\theta} \log \pi_{\theta}(a_t \mid s_t)\big] \bigg] \\
    &= \mathbb{E}_{s_{0:t},a_{0:t-1}}\bigg[\sum_{t=0}^{T-1} b(s_t) \sum_{a_t \in \mathcal{A}}\pi_{\theta}(a_t \mid s_t) \triangledown_{\theta} \log \pi_{\theta}(a_t \mid s_t)\big] \bigg] \\
    &= \mathbb{E}_{s_{0:t},a_{0:t-1}}\bigg[\sum_{t=0}^{T-1} b(s_t) \sum_{a_t \in \mathcal{A}}\triangledown_{\theta} \pi_{\theta}(a_t \mid s_t)\big] \bigg] \\
    &= \mathbb{E}_{s_{0:t},a_{0:t-1}}\bigg[\sum_{t=0}^{T-1} b(s_t) \triangledown_{\theta} \sum_{a_t \in \mathcal{A}} \pi_{\theta}(a_t \mid s_t)\big] \bigg] \\
    &= \mathbb{E}_{s_{0:t},a_{0:t-1}}\bigg[\sum_{t=0}^{T-1} b(s_t) \triangledown_{\theta} \cdot 1 \big] \bigg] \\
    &= 0
\end{align}
$\mathbb{E}_{\tau \sim \pi_{\theta}} \bigg[ \sum_{t=0}^{T-1} \triangledown_{\theta} \log \pi(a_t \mid s_t) \sum_{t^{\prime}=t}^{T-1}\gamma^{t^{\prime}-t}\beta \bigg]$ can be solved similarly and also results in 0.
\newpage

\section{Experimental Settings}
\label{sec:experimental-settings}
All experiments were run on a NVIDIA GeForce GTX 980 GPU with 3 GB RAM.
\\
We adopted Adam \cite{kingma2014adam} to optimize all neural networks.
\\
To calculate the nominal policy, we use Proximal Policy Optimization (PPO) \cite{schulman2017proximal}.
The output layer of the policy network utilizes a soft-max activation function such that the requirements of the policy specified in eq.~\ref{eq:theta-search-space} are met.
\\
The $\phi_\zeta$ network is a standard feedforward neural network with ReLU activation on the hidden layers and Sigmoid activation on the output layer.
$\lambda$ and $\zeta$ are adjusted using possibly different learning rates.
Every iteration the learning rate for $\zeta$ is updated using an exponential decay schedule parameterized by a decay factor.
Table \ref{table:hyperparams} reports an overview of the used hyperparameters.
\begin{table}[!ht]
    \vskip -0.2in
    \caption{Overview of the used hyperparameters}
    \vskip 0.1cm
    \label{table:hyperparams}
    \setlength{\tabcolsep}{2pt}
    \centering
    \begin{tabular}{lcccccccc}
    \hline
        ~ & Ant & Cheetah & Swimmer & Walker & Pendulum & Gridworld & HighD dist. & HighD vel. \\ \hline
        General & ~ & ~ & ~ & ~ & ~ & ~ & ~ & ~ \\
        \hspace{0.25cm} $\alpha$ & 0 & 0 & 0 & 0 & 0 & 0 & 0 & 0 \\
        \hspace{0.25cm} $\beta$ & 0.0001 & 0.0001 & 0.0001 & 0.0001 & 0.0001 & 0.00001 & 0.0001 & 0.0001 \\
        \hspace{0.25cm} $\gamma$ & 0.99 & 0.99 & 0.99 & 0.99 & 0.99 & 0.99 & 0.99 & 0.99 \\
        \hspace{0.25cm} Iterations & 20 & 30 & 100 & 50 & 200 & 20 & 100 & 50 \\
        \hspace{0.25cm} Expert trajectories & 20 & 10 & 50 & 50 & 50 & 50 & 20 & 50 \\
        \hspace{0.25cm} Max. trajectory length & 500 & 1000 & 500 & 500 & 100 & 200 & 1000 & 1000 \\
        PPO & ~ & ~ & ~ & ~ & ~ & ~ & ~ & ~ \\
        \hspace{0.25cm} Batch size & 128 & 128 & 128 & 64 & 64 & 64 & 64 & 64 \\
        \hspace{0.25cm} Learning rate & 3.00E-05 & 3.00E-05 & 3.00E-05 & 0.0003 & 0.0001 & 3e-0.5 & 0.0001 & 0.0001 \\
        \hspace{0.25cm} Steps & 2048 & 2048 & 2048 & 2048 & 2048 & 2048 & 1024 & 1024 \\
        \hspace{0.25cm} Epochs & 20 & 20 & 20 & 10 & 20 & 10 & 20 & 20 \\
        \hspace{0.25cm} Timesteps & 200000 & 200000 & 50000 & 200000 & 100000 & 50000 & 50000 & 25000 \\
        \hspace{0.25cm} Gae-$\lambda$ & 0.9 & 0.9 & 0.9 & 0.95 & 0.9 & 0.9 & 0.9 & 0.9 \\
        \hspace{0.25cm} Target KL & 0.02 & 0.02 & 0.02 & 0.01 & 0.01 & 0.01 & 0.02 & 0.02 \\
        \hspace{0.25cm} Bootstrap timesteps & 200000 & 200000 & 200000 & 200000 & 100000 & 35000 & 20000 & 0 \\
        \hspace{0.25cm} Policy network & [64, 64] & [64, 64] & [64, 64] & [64, 64] & [64, 64] & [64,64] & [64, 64] & [64, 64] \\
        $\lambda$, $\zeta$ & ~ & ~ & ~ & ~ & ~ & ~ & ~ & ~ \\
        \hspace{0.25cm} Learning rate decay factor $\zeta$  & 0.9 & 0.9 & 0.9 & 1 & 0.9 & 1 & 0.9 & 0.9 \\
        \hspace{0.25cm} Batch size & 64 & 64 & 64 & 64 & 128 & 64 & 1000 & 1000 \\
        \hspace{0.25cm} Network $\phi_{\zeta}$ & [40] & [40] & [40] & [40] & [64] & [40, 40] & [20] & [40, 20, 10] \\
        \hspace{0.25cm} Feature dimensions & 64 & 64 & 64 & 64 & 16 & 64 & 10 & 8 \\
        \hspace{0.25cm} Learning rate $\zeta$ & 0.0003 & 0.0003 & 0.0003 & 0.0003 & 0.0005 & 0.0005 & 0.0001 & 1.00E-06 \\
        \hspace{0.25cm} Learning rate $\lambda$ & 0.0001 & 0.0001 & 0.0001 & 0.0001 & 0.0001 & 0.0005 & 0.0001 & 1.00E-06 \\
        \hspace{0.25cm} Bootstrap iterations & 20 & 200 & 200 & 200 & 5 & 20 & 20 & 0 \\
        \hspace{0.25cm} $\lambda$ initial values & 1 & 1 & 1 & 1 & 1 & 1 & 1 & 1 \\
        \hline
    \end{tabular}
\end{table}
\newpage

\section{Details on the Environments}
\label{sec:env-details}
In this section we provide details on and screenshots from the environments used for evaluation.

\subsection{Gridworld}
\begin{figure}[h!]
    \centering
    \includegraphics[width=0.24\linewidth]{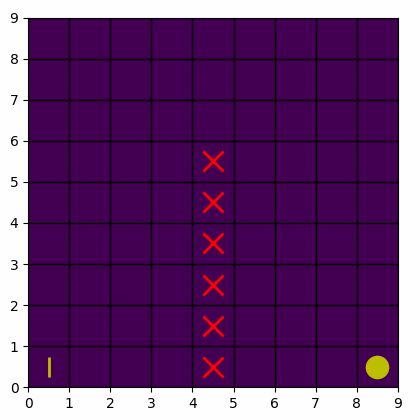}
    \caption{Gridworld environment}
    \label{fig:gt-constraints-gw}
\end{figure}

Figure \ref{fig:gt-constraints-gw} shows the gridworld used in the experiments, with the yellow ``I'' representing the initial state, the yellow ``O'' the goal state and the red crosses the constrained states.

\subsection{Virtual Robotics Environment}
\label{sec:env-details-mujoco}
\begin{figure}[ht!]
%\vskip 0in
\begin{center}
    \hspace{0.01cm}
    \subfigure[Ant]{\includegraphics[width=0.18\textwidth]{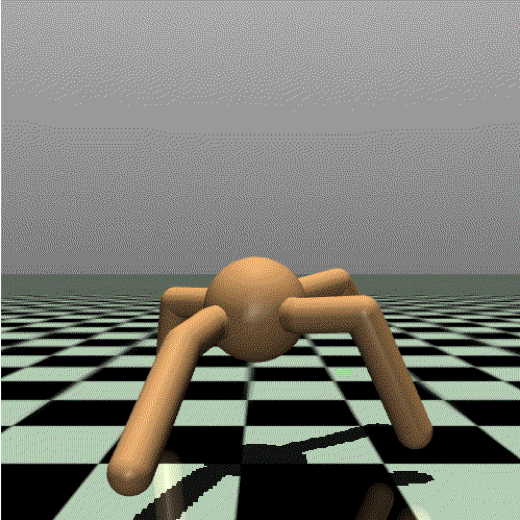} \label{fig:ant}}
    \subfigure[Half-cheetah]{\includegraphics[width=0.18\textwidth]{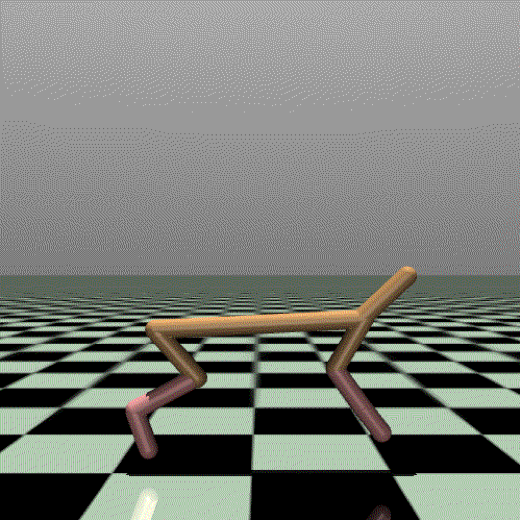} \label{fig:cheetah}}
    \subfigure[Swimmer]{\includegraphics[width=0.18\textwidth]{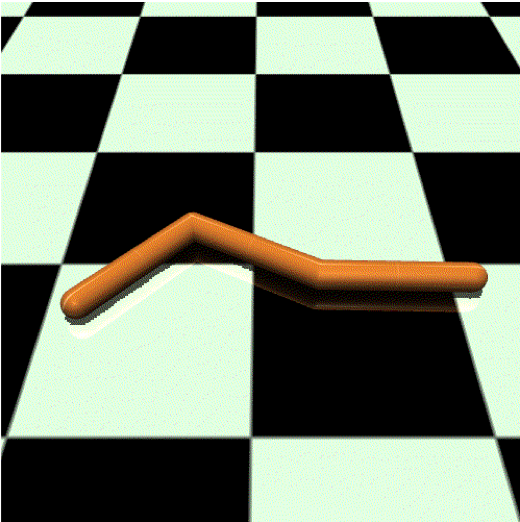} \label{fig:swimmer}}
    \subfigure[Walker]{\includegraphics[width=0.18\textwidth]{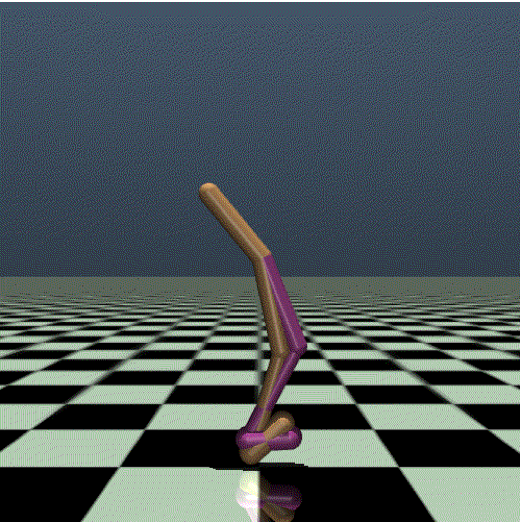} \label{fig:walker}}
    \subfigure[Inverted pendulum]{\includegraphics[width=0.18\textwidth]{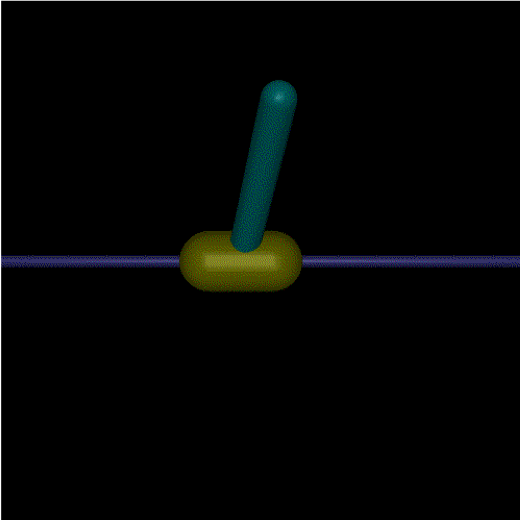} \label{fig:pendulum}}
    \caption{Screenshots from the virtual robotics environments.}
    \label{fig:robotic-env-screenshots}
\end{center}
%\vskip 0in
\end{figure}

\subsection{Realistic Traffic Environment}
\label{sec:appendix-traffic}
\begin{figure}[h!]
    \centering
    \includegraphics[width=\linewidth]{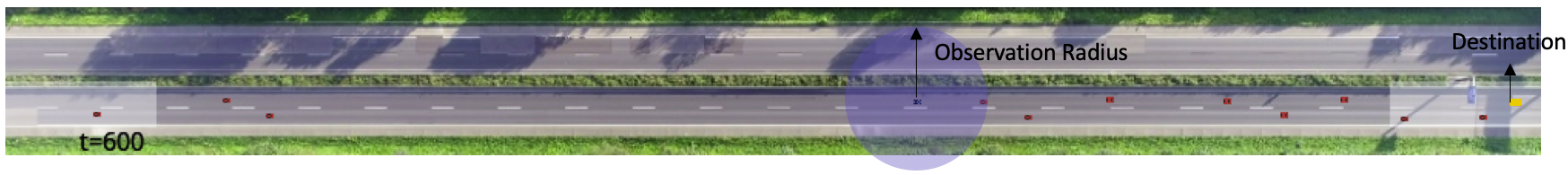}
    \caption{Realistic traffic environment}
    \label{fig:traffic-env}
\end{figure}

Figure \ref{fig:traffic-env} shows the realistic traffic environment (image from original publication \cite{liu2022benchmarking}).
The ego car is shown in blue, other cars are red.
The agent only has partial state information, observations are restricted by an observation radius (visualized in blue around the ego agent).
The agent's destination is shown in yellow.

\section{More Experimental Results}
\label{sec:appendix-more-results}

\subsection{Gridworld}
\label{sec:exp-results-gw}
\begin{figure}[h!]
\vskip 0in
\begin{center}
    \hspace{0.01cm}
    \subfigure[Expert trajectories state visitations]{\includegraphics[width=0.24\textwidth]{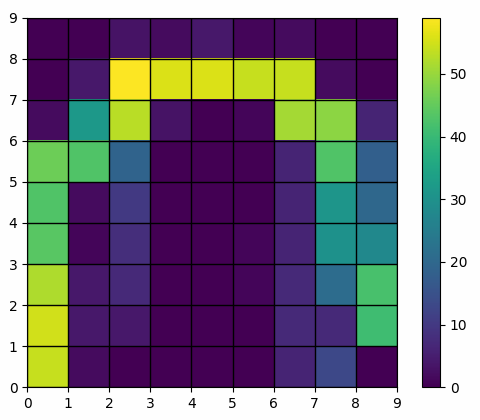} \label{fig:s-count-expert-gw}}
    \subfigure[Recovered cost function]{\includegraphics[width=0.24\textwidth]{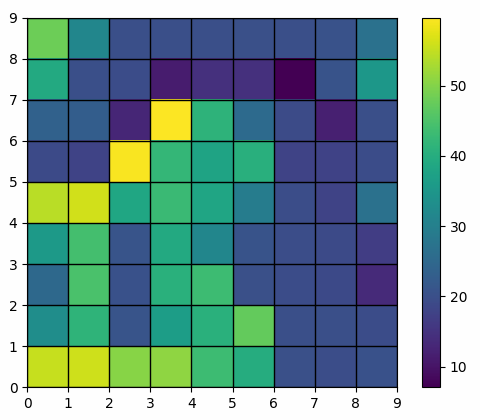} \label{fig:learned-cost-gw}}
    \subfigure[State visitations for trajectories sampled from the nominal policy]{\includegraphics[width=0.24\textwidth]{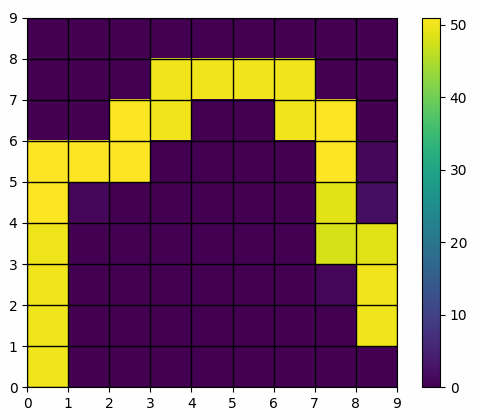} \label{fig:s-count-nominal-gw}}
    \hspace{-0.2cm}
    \caption{Results from MCE-ICRL ($\beta=0.01$) in the proposed gridworld environment.}
    \label{fig:addresultsgw}
\end{center}
\vskip 0in
\end{figure}

Figure \ref{fig:s-count-expert-gw} shows the states visited by the expert which are used as input for our method.
Figure \ref{fig:learned-cost-gw} reports the learned cost function when using MCE-ICRL with $\beta=0.01$.
Figure \ref{fig:s-count-nominal-gw} shows the state visitations of the nominal agent.

\begin{figure*}[h!]
\vskip 0in
\begin{center}
\subfigure{\includegraphics[width=0.17\textwidth]{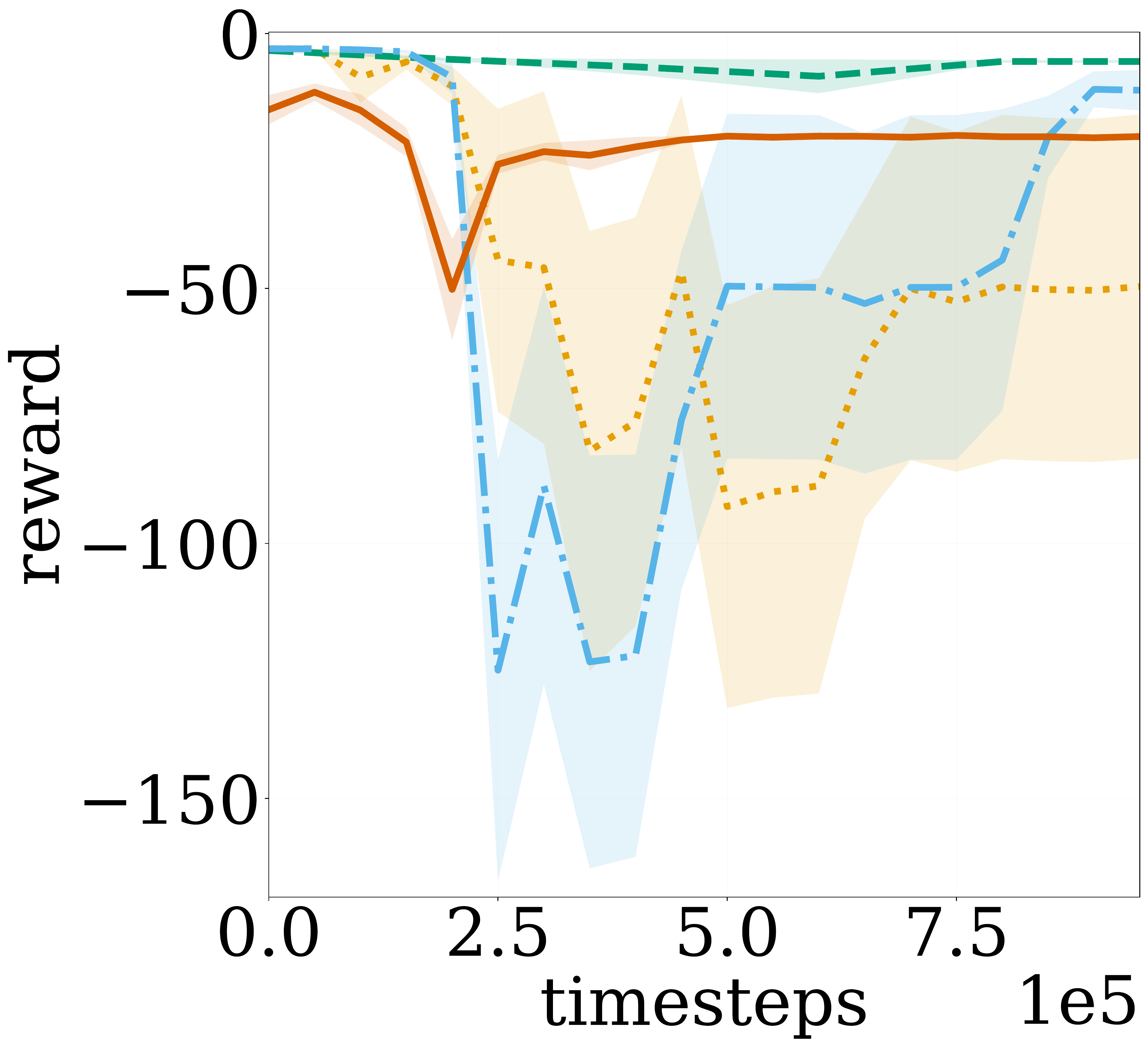}} %\hspace{0.2in}
    \subfigure{\includegraphics[width=0.16\textwidth]{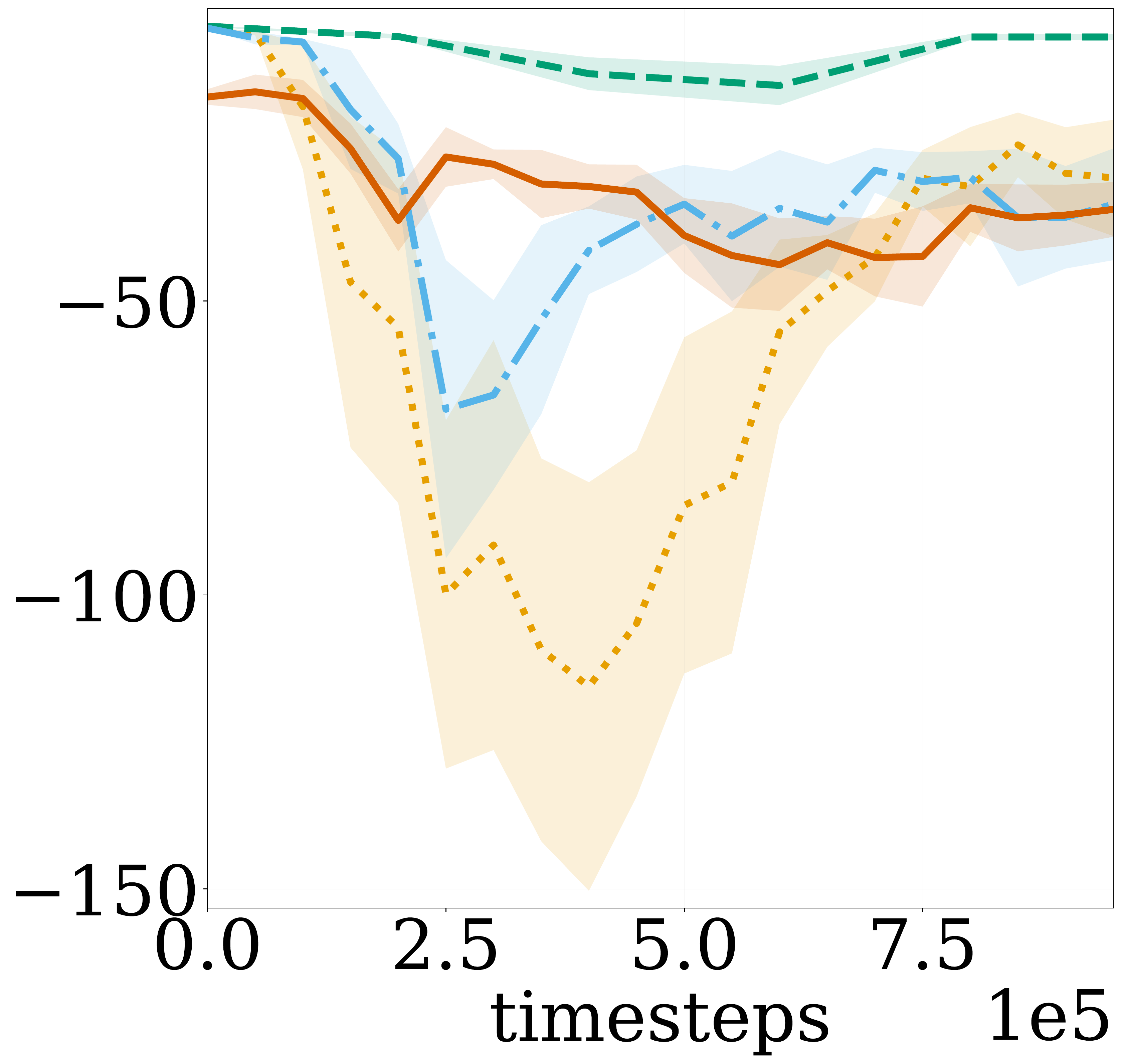}}
    \subfigure{\includegraphics[width=0.16\textwidth]{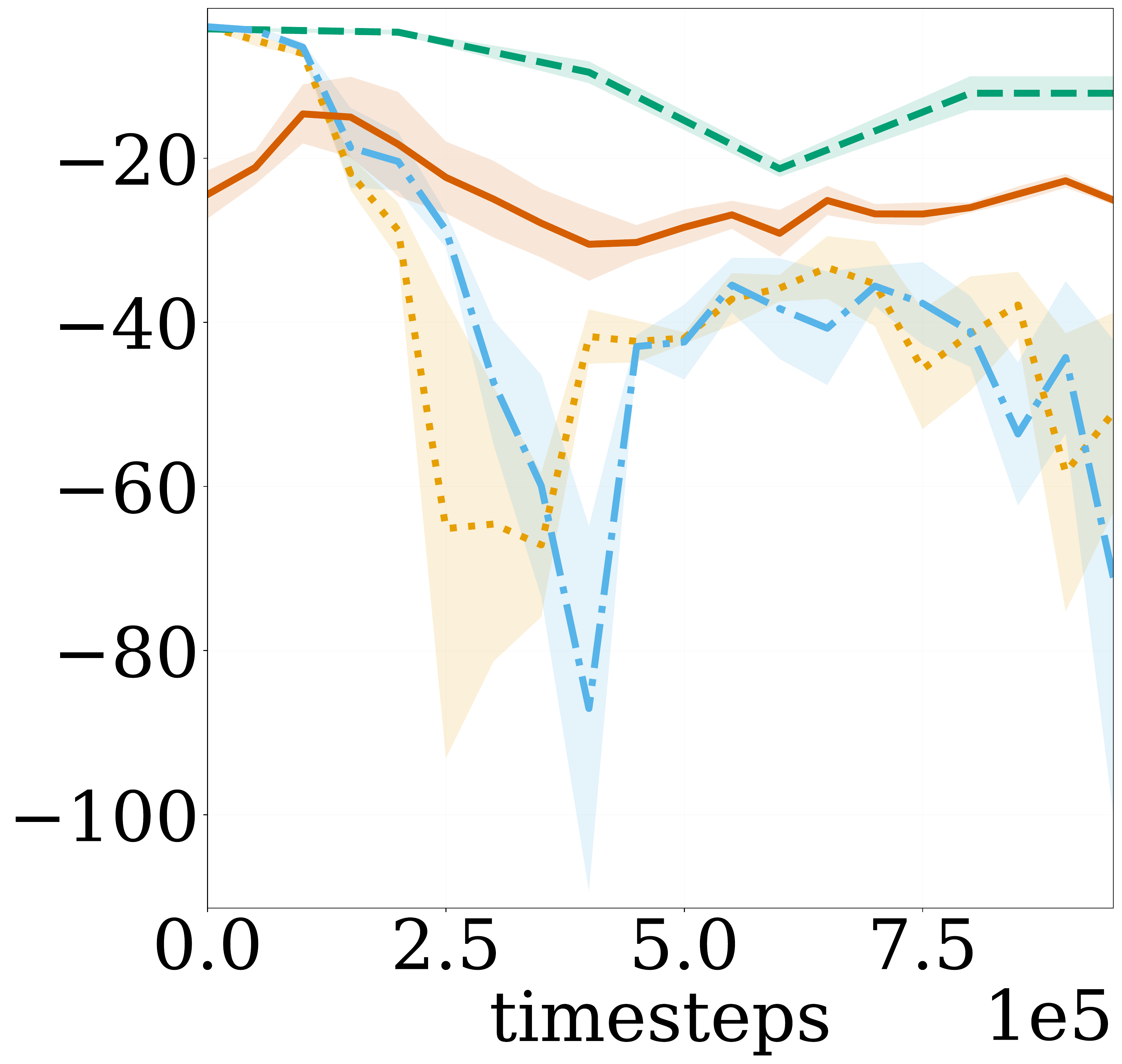}}
    \subfigure{\includegraphics[width=0.16\textwidth]{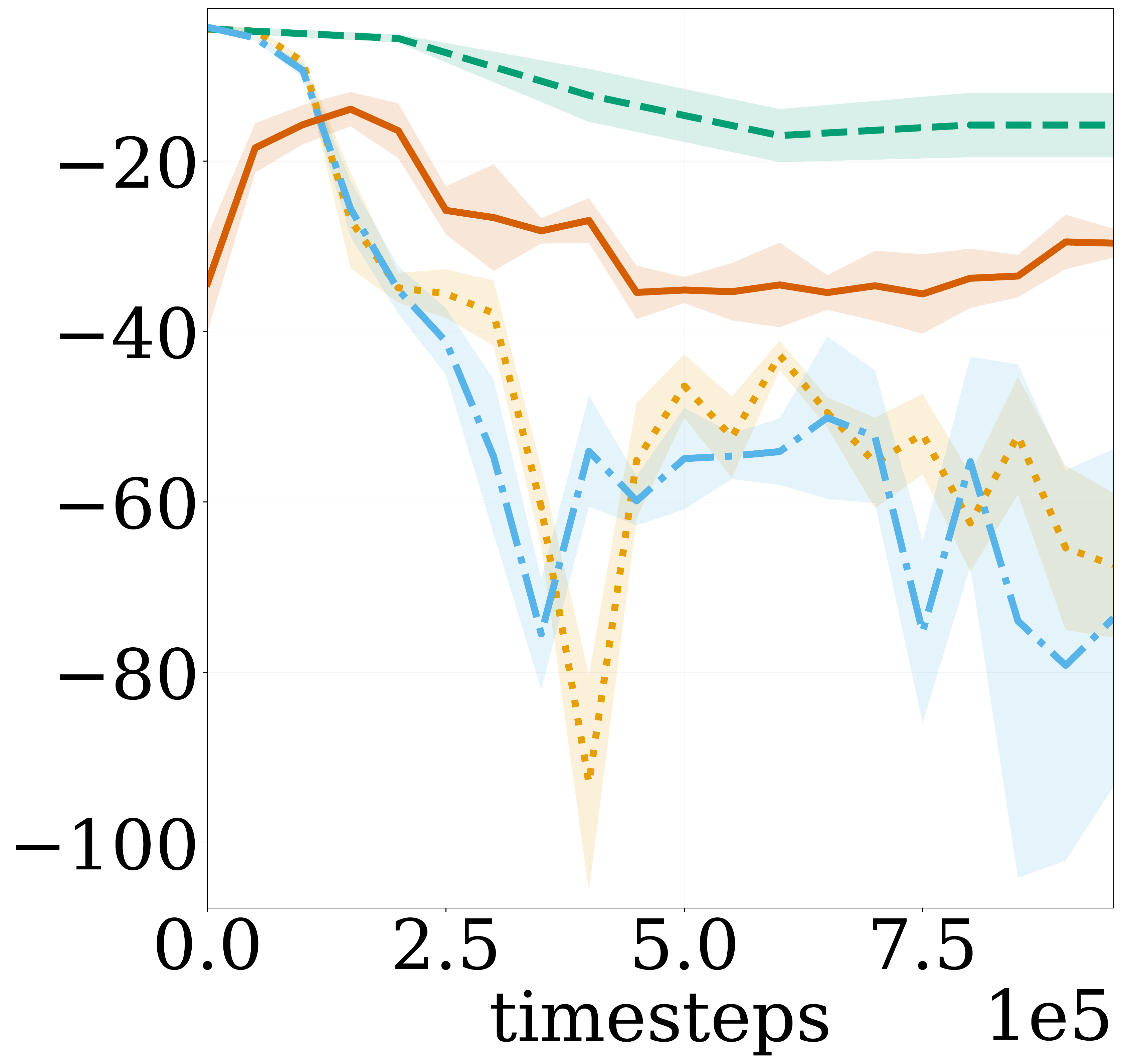}}
    \subfigure{\includegraphics[width=0.16\textwidth]{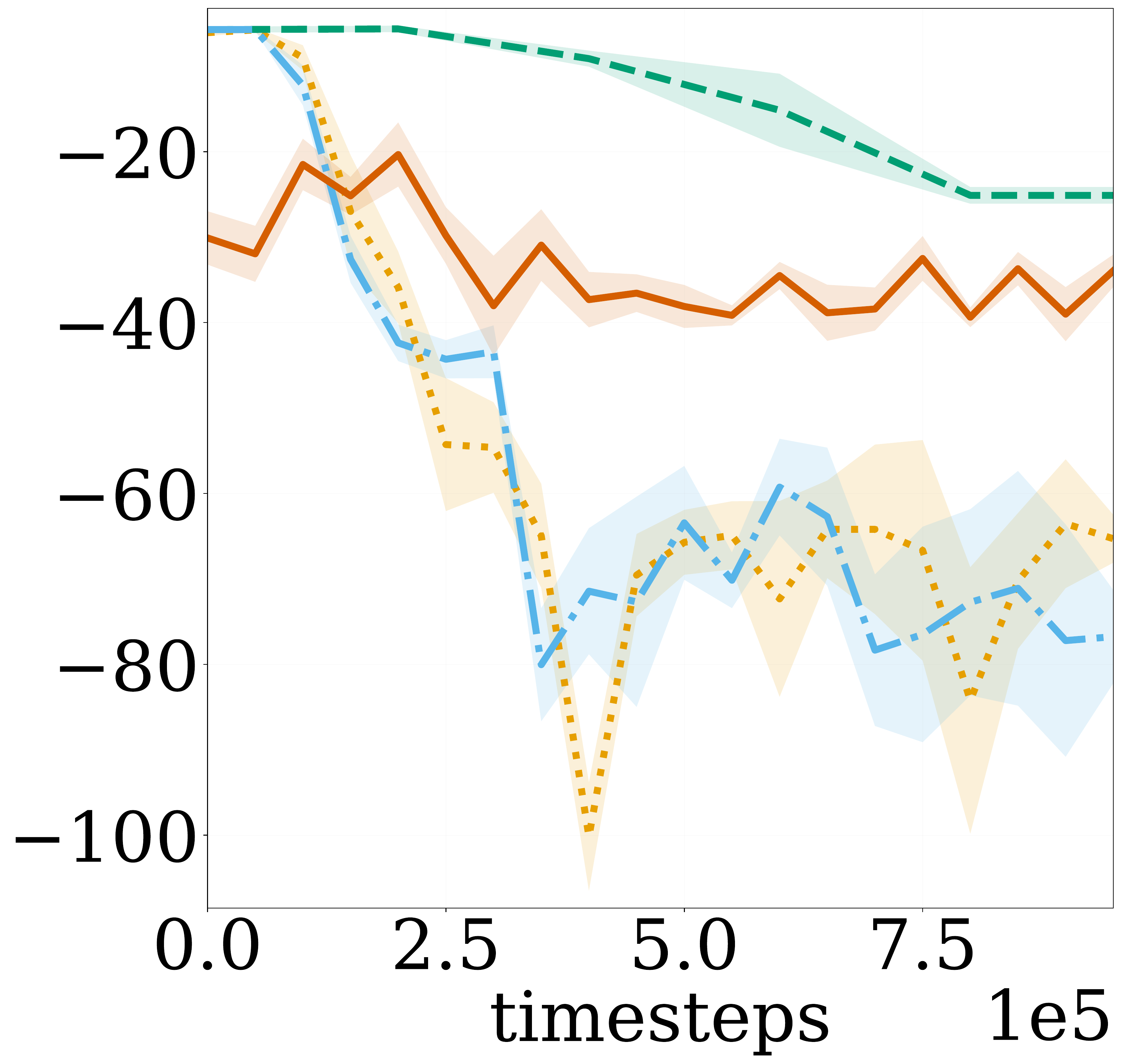}}
    \subfigure{\includegraphics[width=0.16\textwidth]{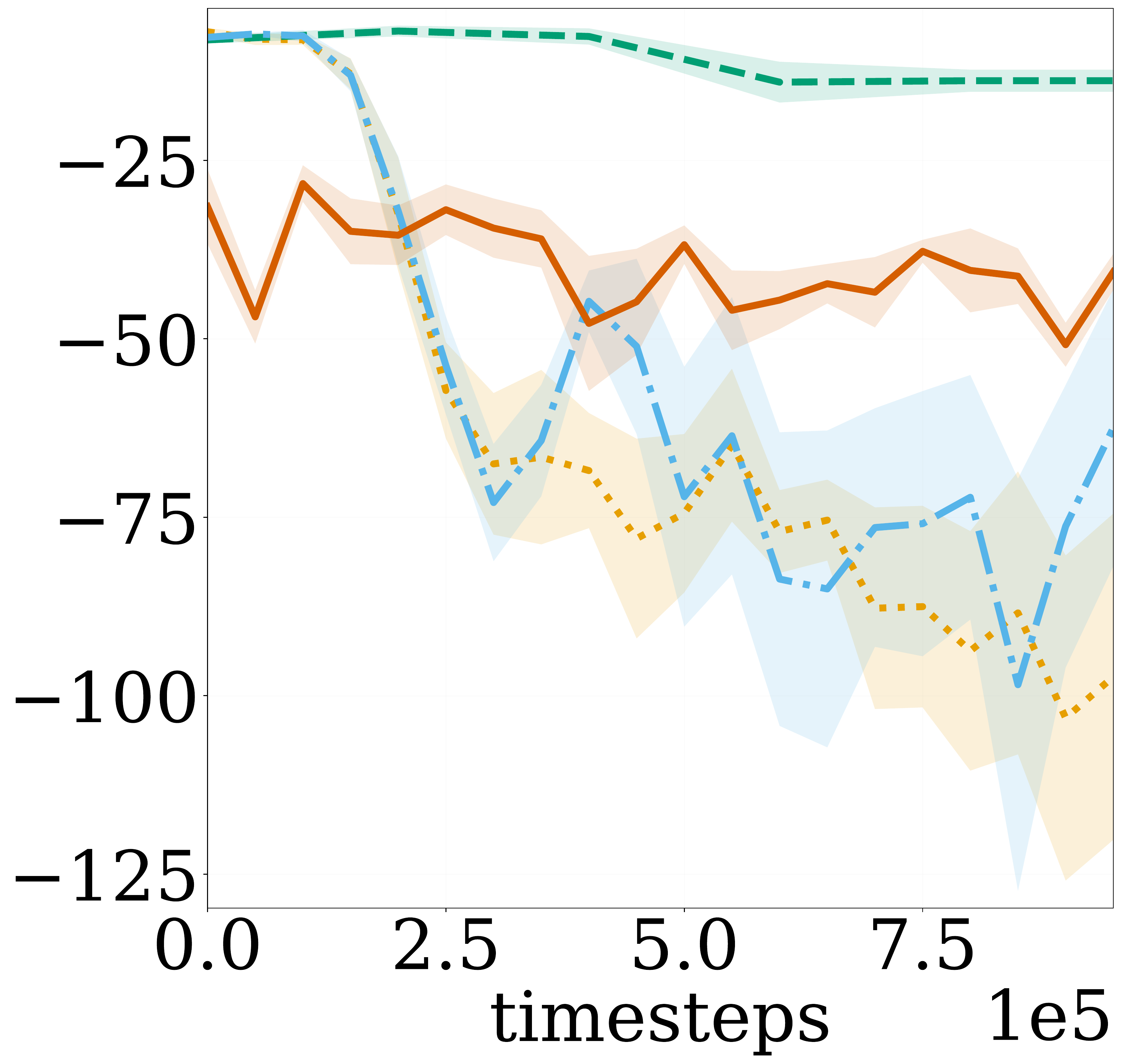}}\\
    \setcounter{subfigure}{0}
    \hspace{0.1cm}
    \subfigure[0.0]{\includegraphics[width=0.155\textwidth]{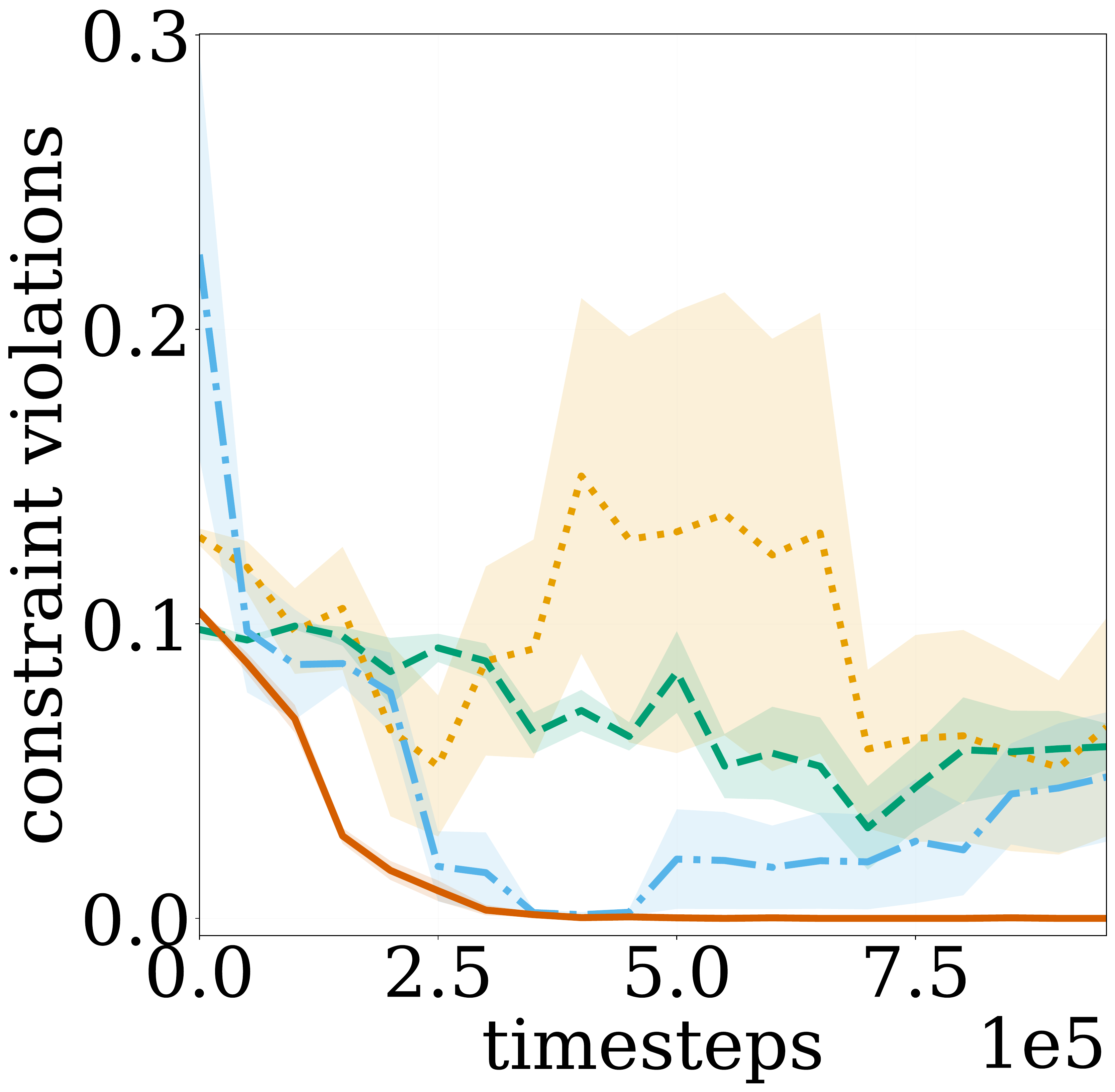}} 
    \hspace{0cm}
    \subfigure[0.1]{\includegraphics[width=0.155\textwidth]{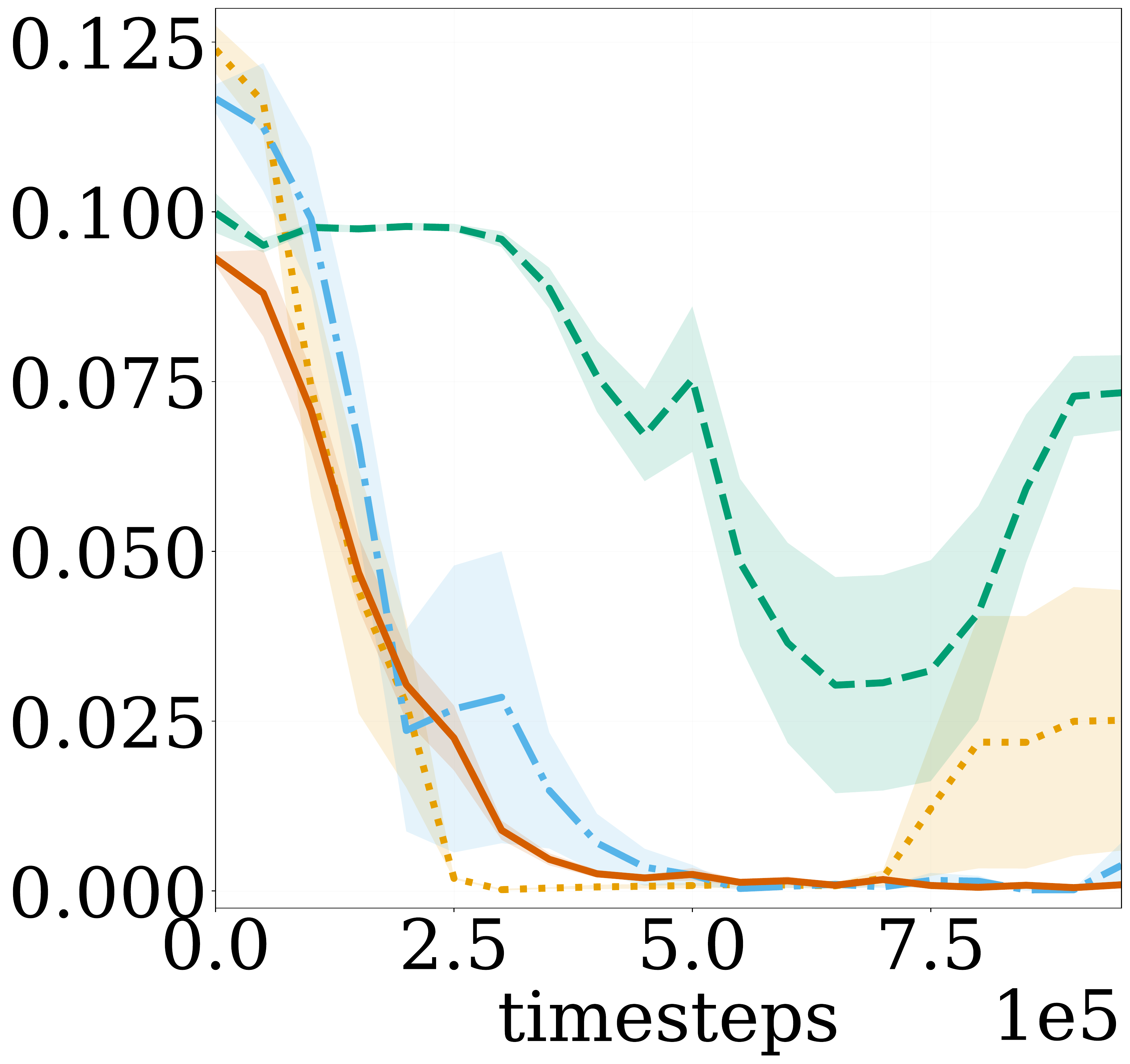}}
    \hspace{0cm}
    \subfigure[0.2]{\includegraphics[width=0.155\textwidth]{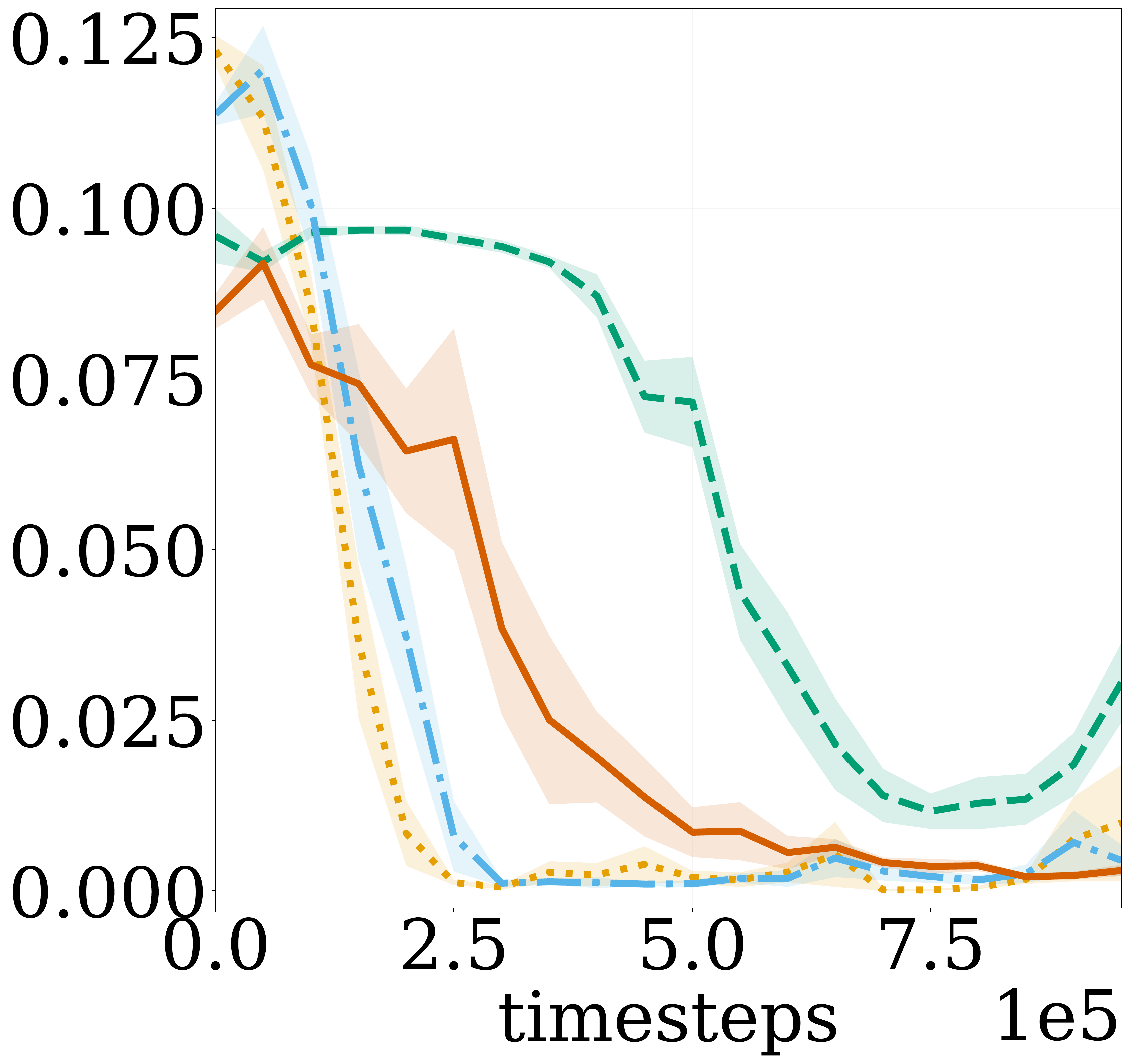}}
    \hspace{0cm}
    \subfigure[0.3]{\includegraphics[width=0.155\textwidth]{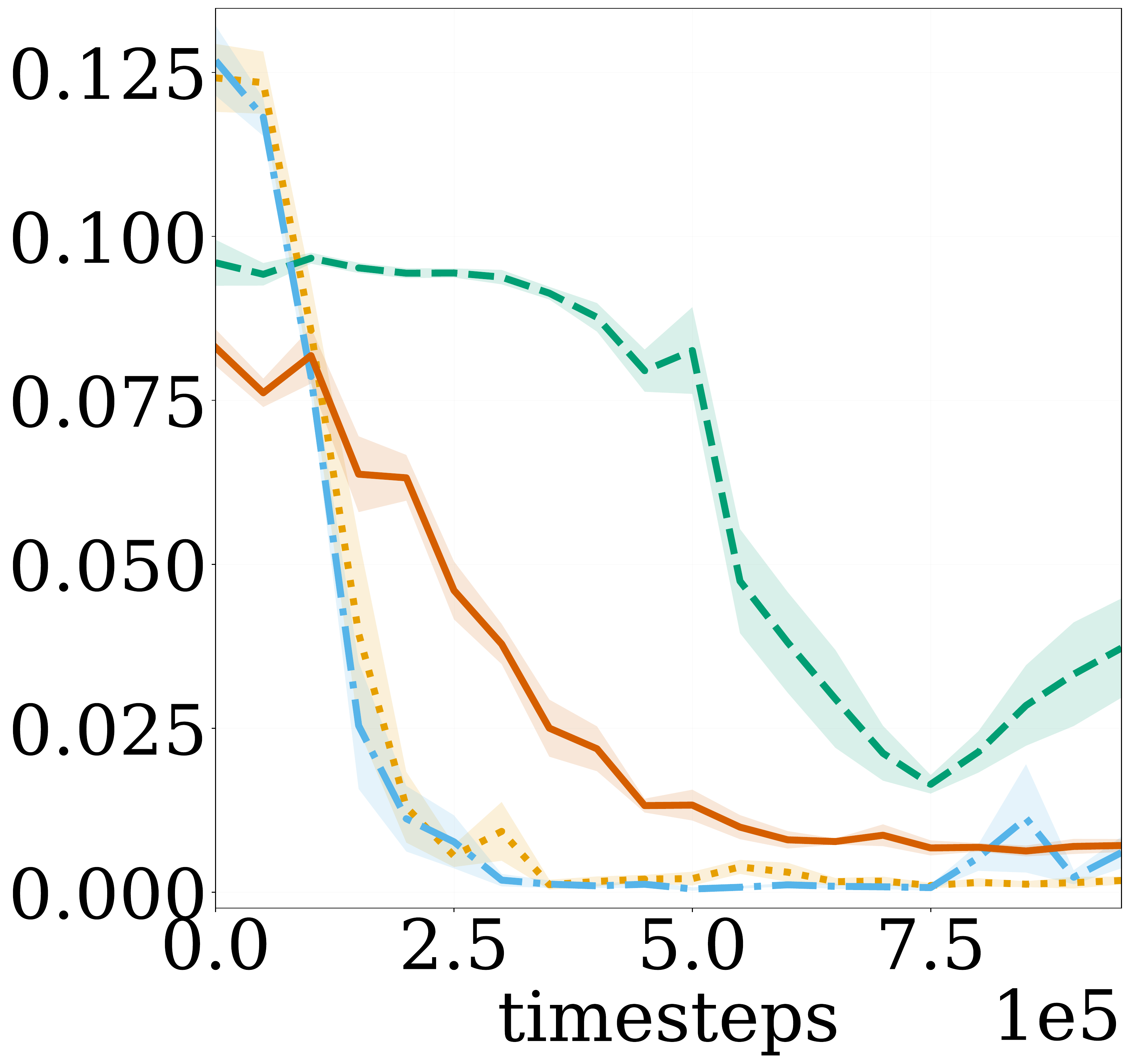}}
    \hspace{0cm}
    \subfigure[0.4]{\includegraphics[width=0.155\textwidth]{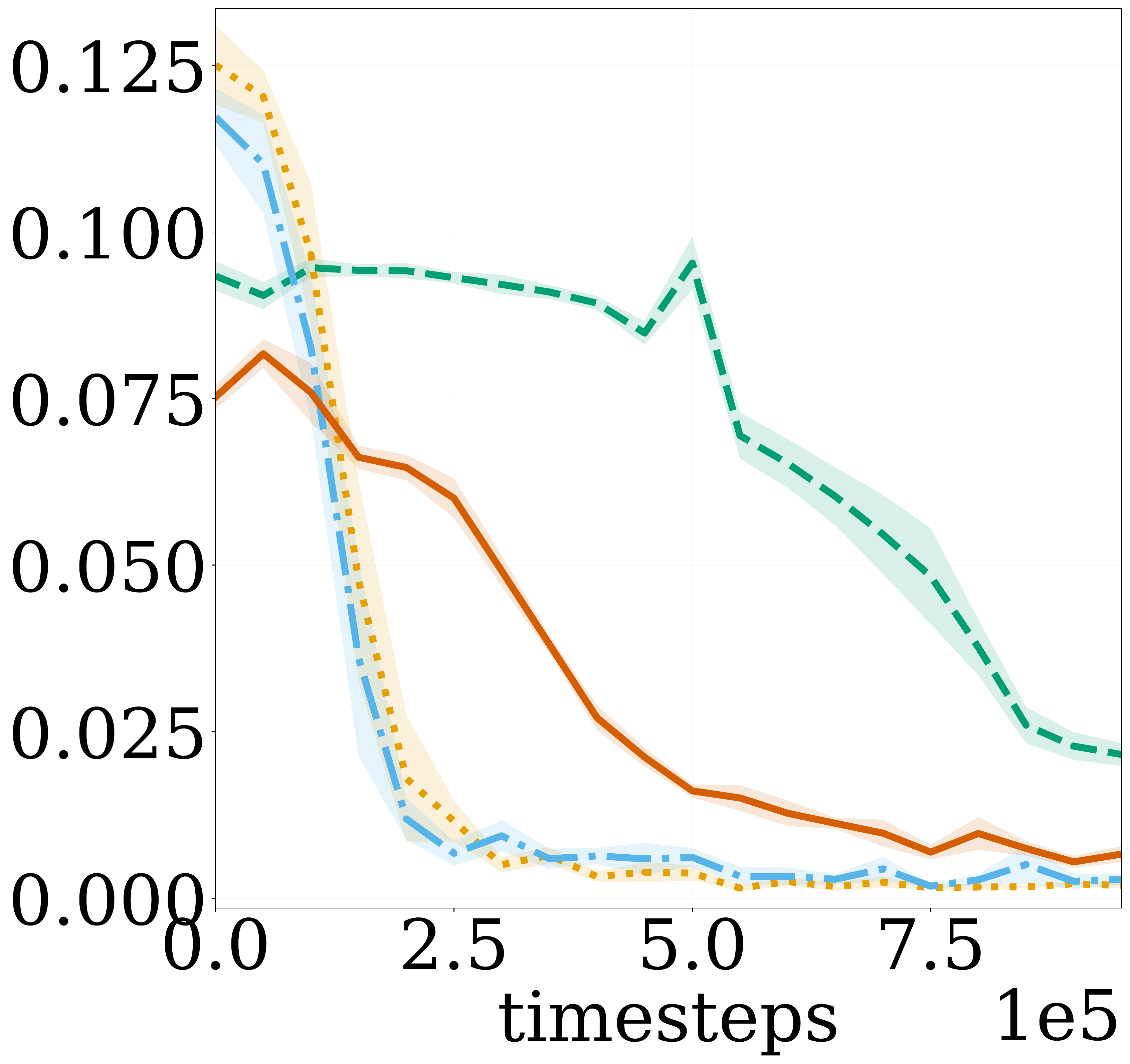}}
    \hspace{0cm}
    \subfigure[0.5]{\includegraphics[width=0.155\textwidth]{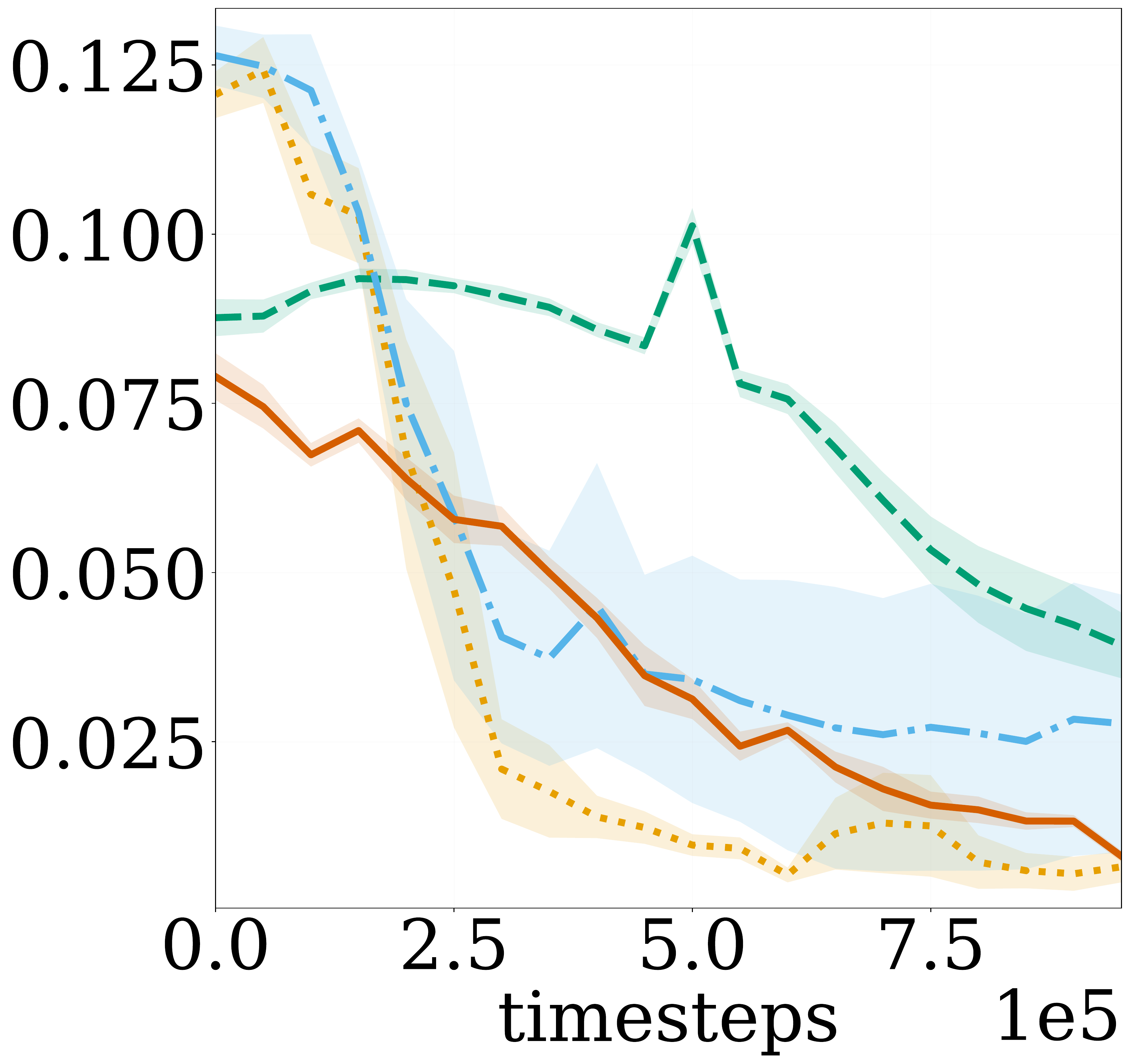}}
    \subfigure{\includegraphics[width=0.6\textwidth]{figures/legend.pdf}}\\
    \caption{Evaluation of the different methods in the gridword environment for different rates of stochasticity: reward (top) and constraint violation rate (bottom) of trajectories sampled from the nominal policy during training. Results are averaged over 5 random seeds. The x-axis is the number of timesteps taken in the environment during training. The shaded regions correspond with the standard error.}
    \label{fig:appendix-gridworld-main}
\end{center}
%\vskip -0.2in
\end{figure*}

Figure~\ref{fig:appendix-gridworld-main} depicts the reward and constraint violation rate during training at different timesteps for the different methods for different rates of stochasticity.

\begin{figure*}[h!]
%\vskip -0.2in
\begin{center}    \subfigure{\includegraphics[width=0.17\textwidth]{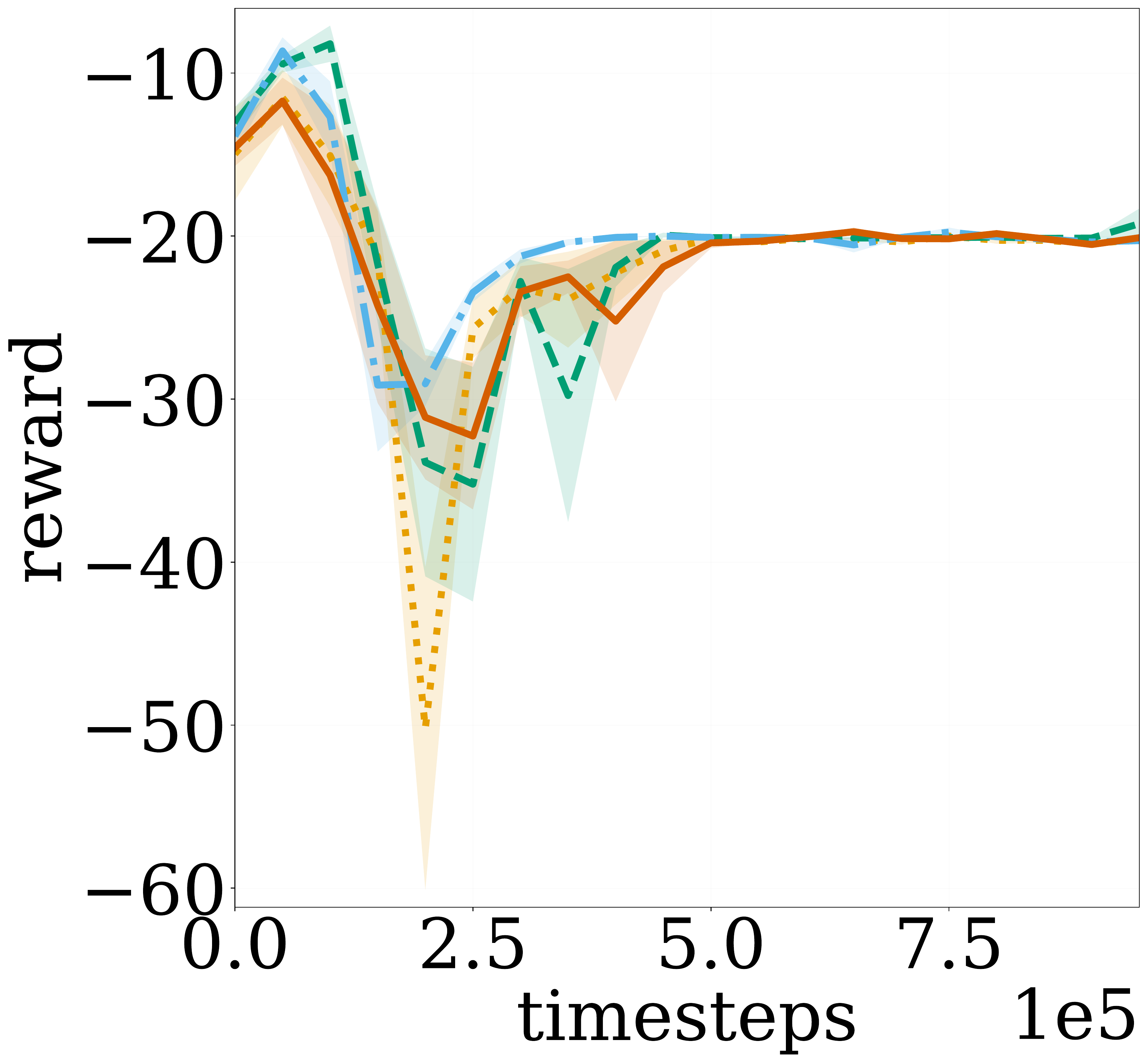}} %\hspace{0.2in}
    \subfigure{\includegraphics[width=0.16\textwidth]{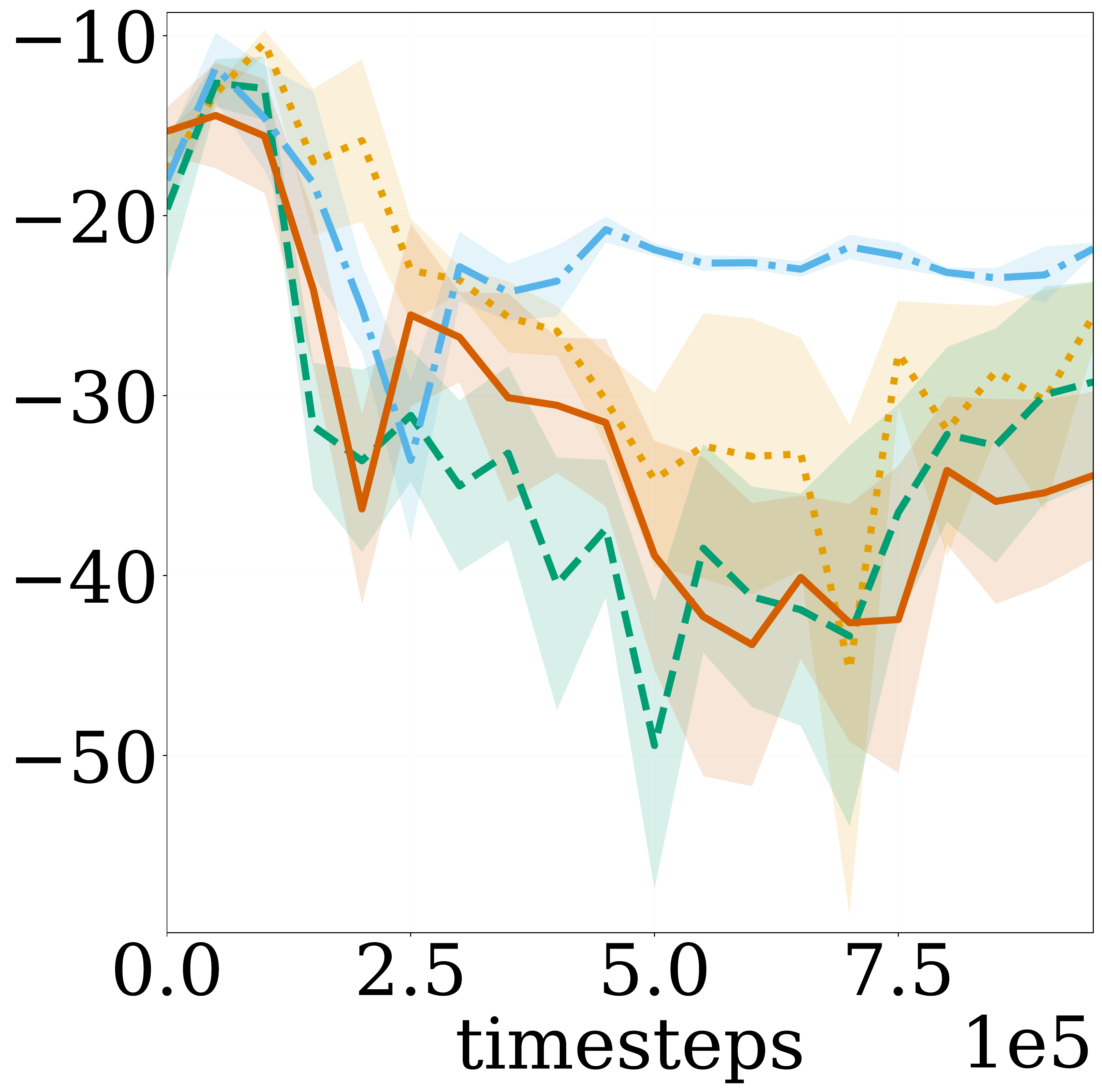}}
    \subfigure{\includegraphics[width=0.16\textwidth]{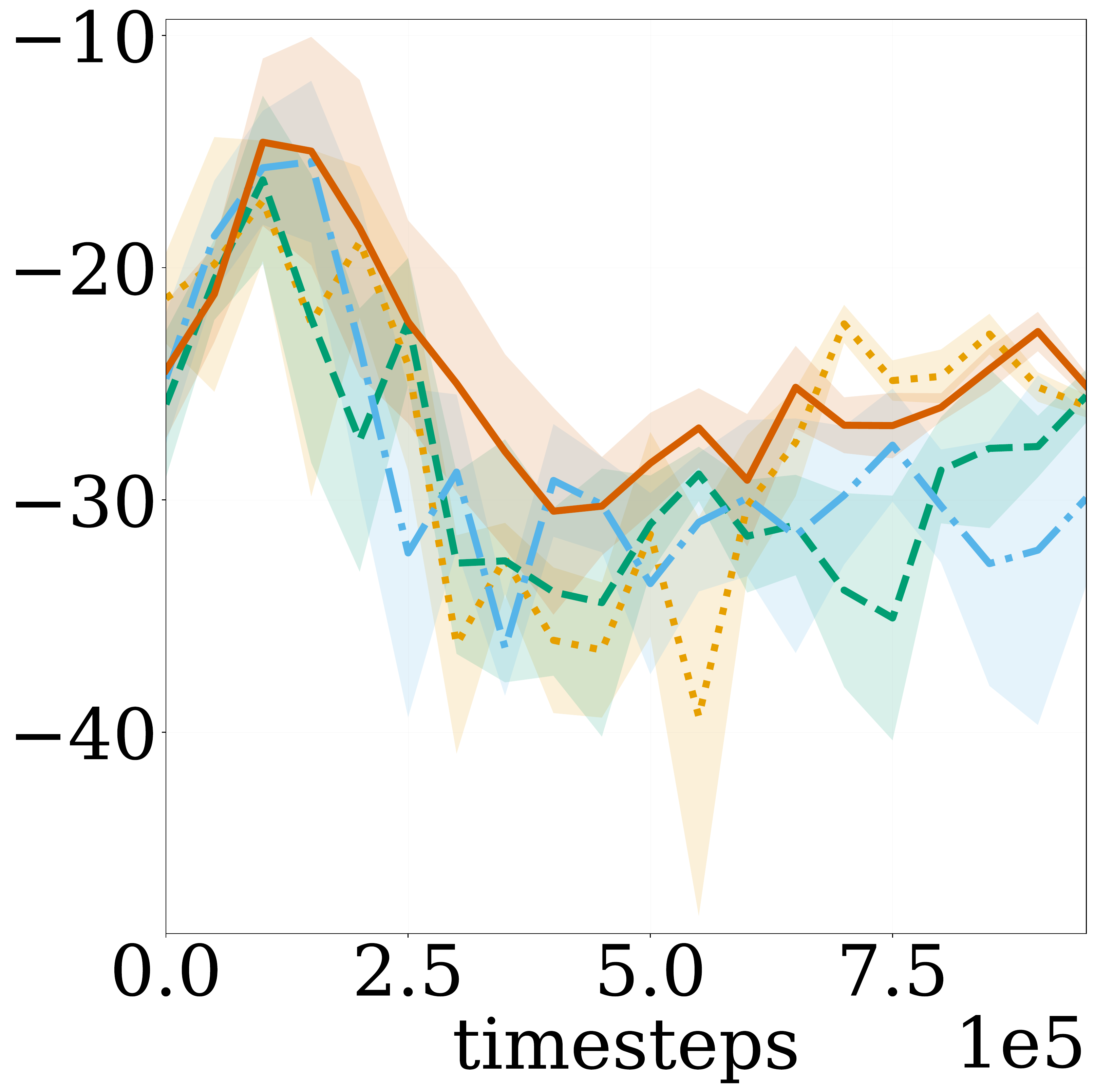}}
    \subfigure{\includegraphics[width=0.16\textwidth]{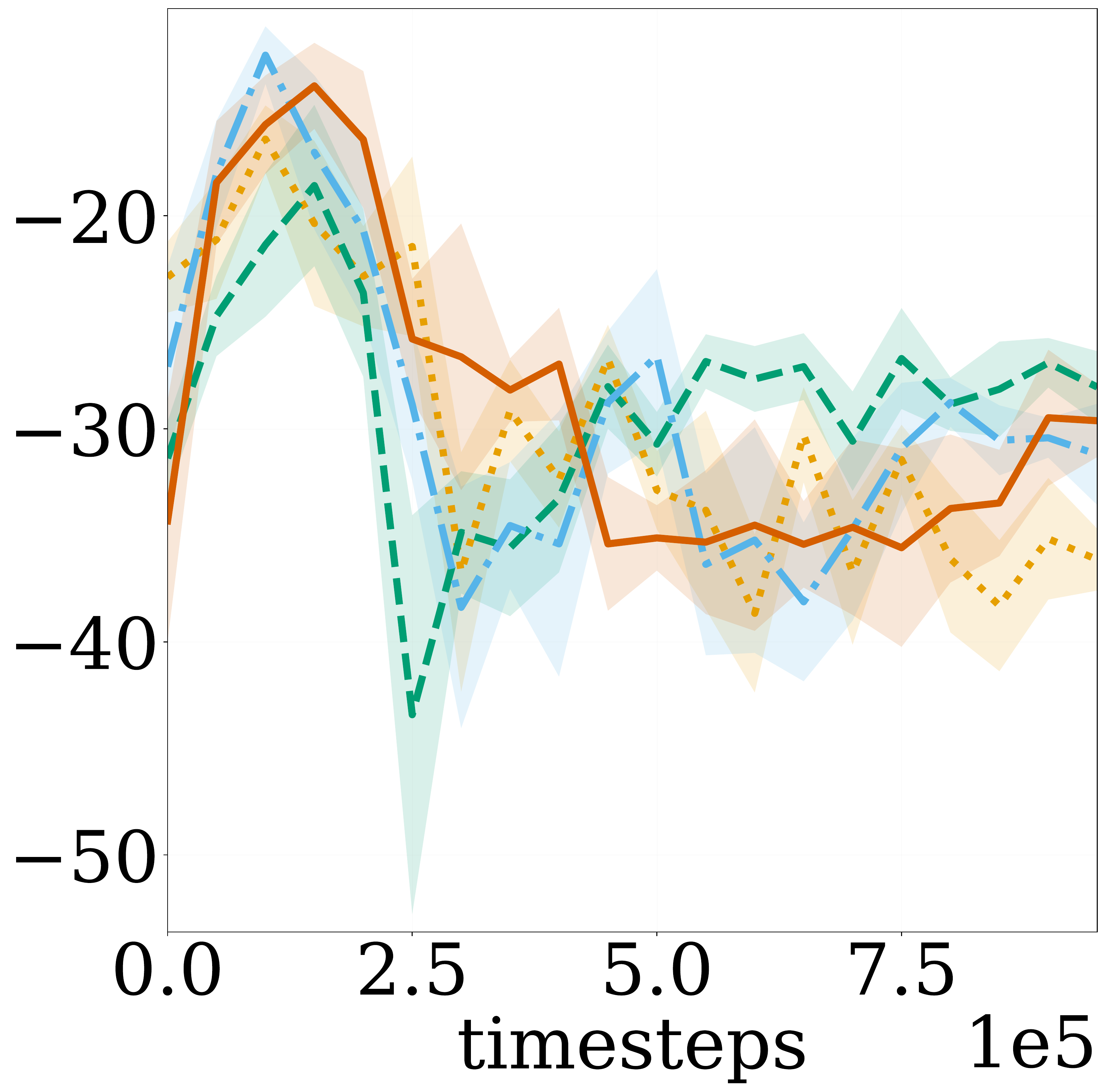}}
    \subfigure{\includegraphics[width=0.16\textwidth]{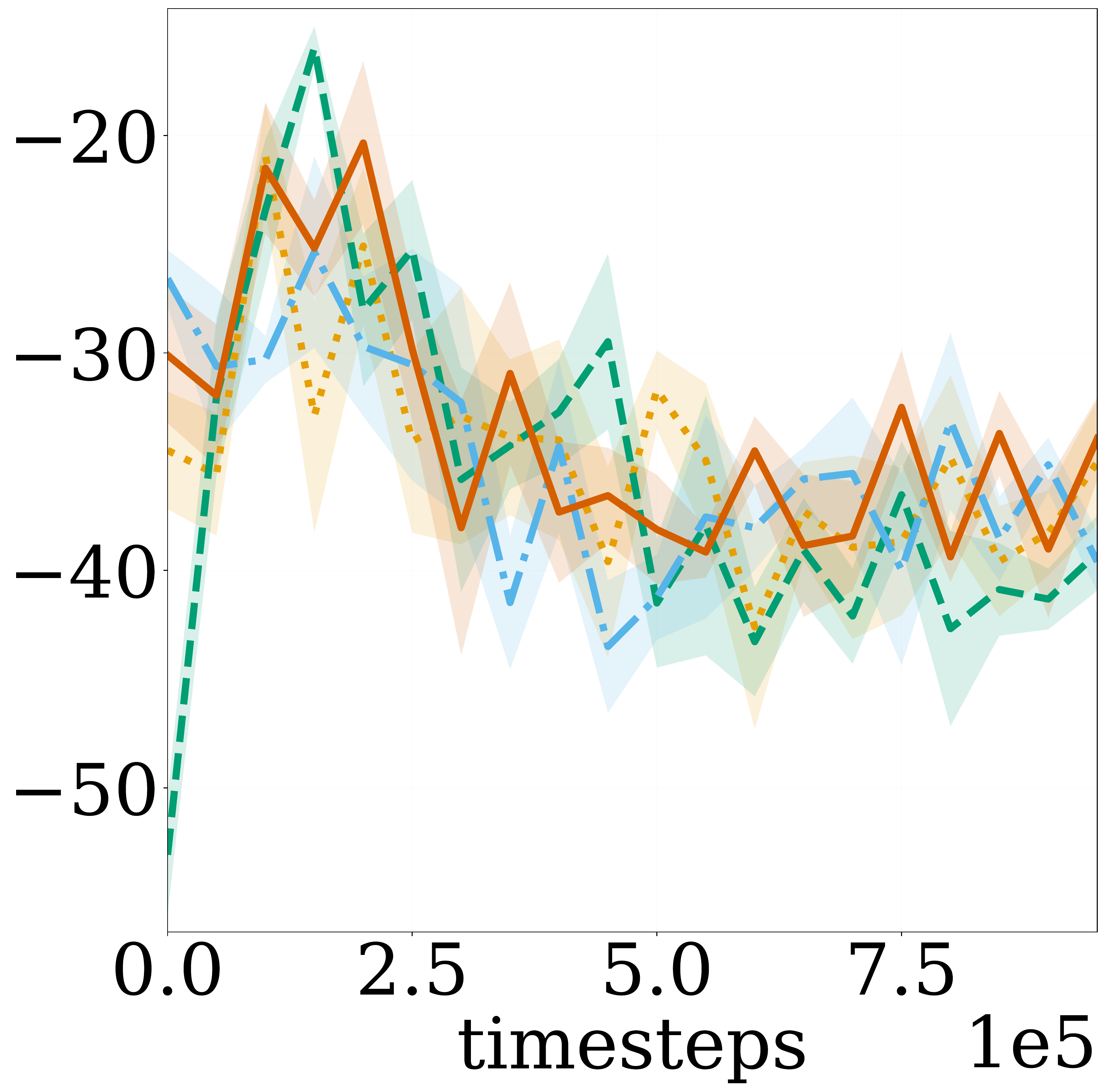}}
    \subfigure{\includegraphics[width=0.16\textwidth]{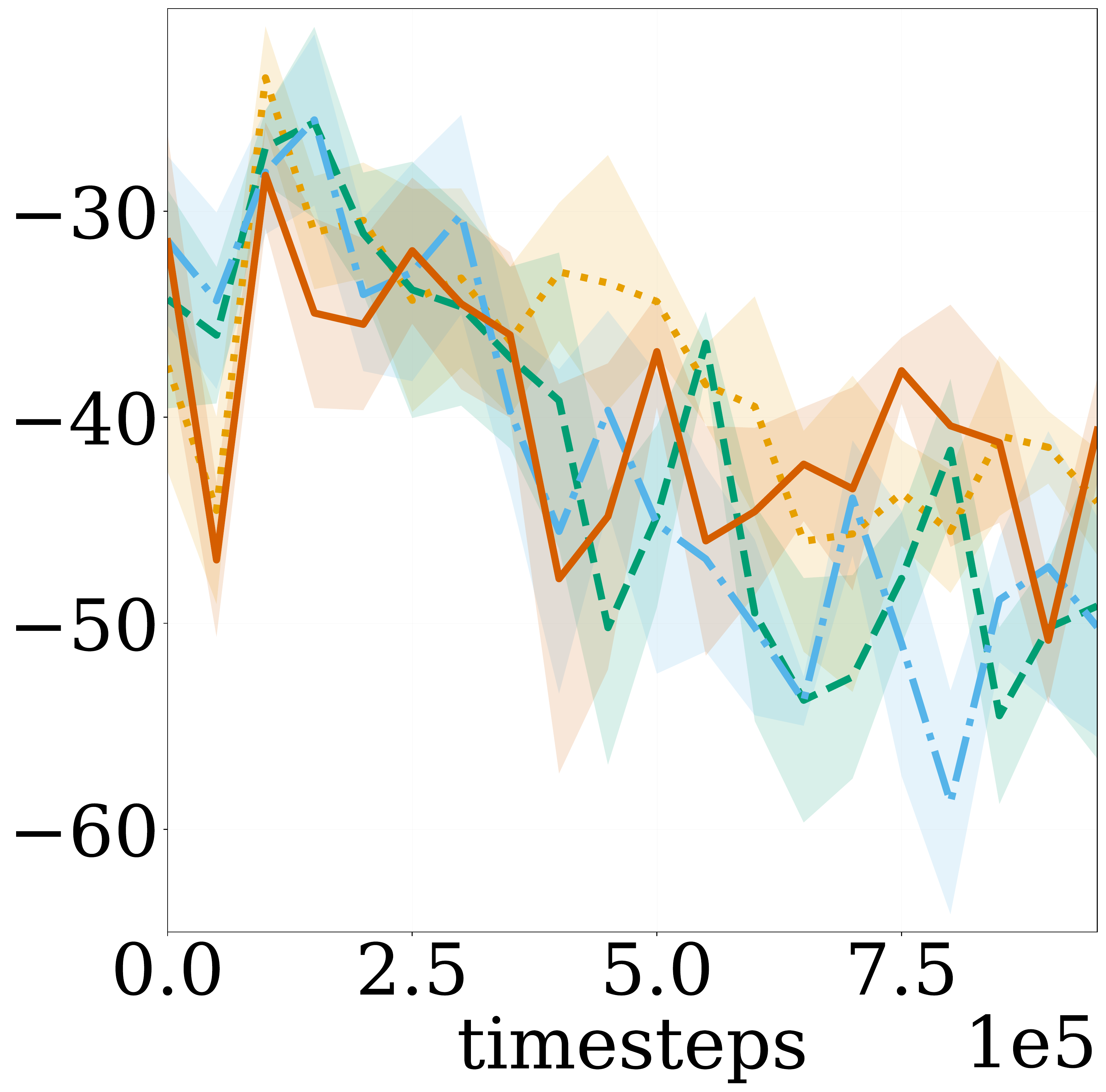}}\\
    \setcounter{subfigure}{0}
    \hspace{0.1cm}
    \subfigure[0.0]{\includegraphics[width=0.155\textwidth]{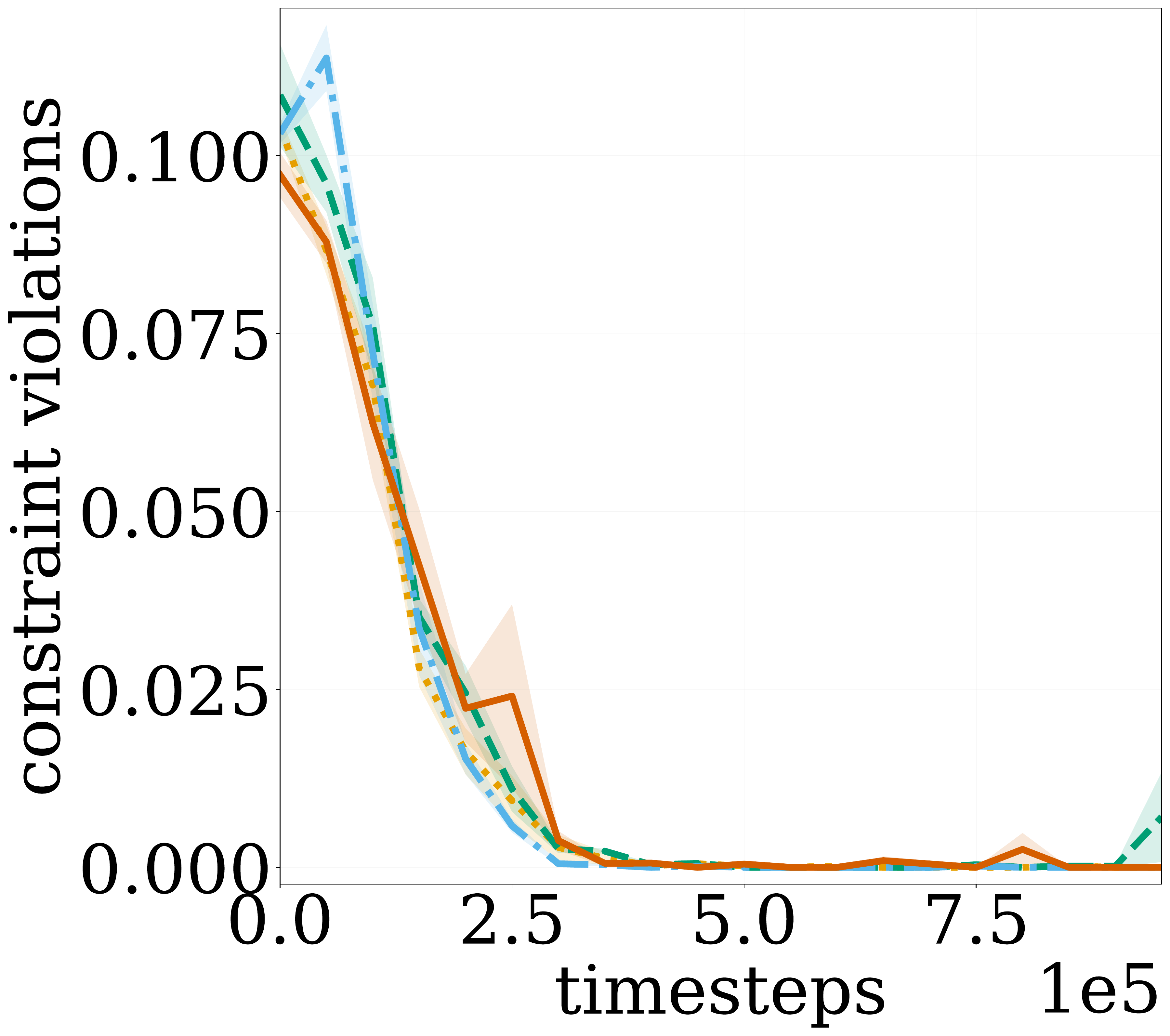}} 
    \hspace{0cm}
    \subfigure[0.1]{\includegraphics[width=0.155\textwidth]{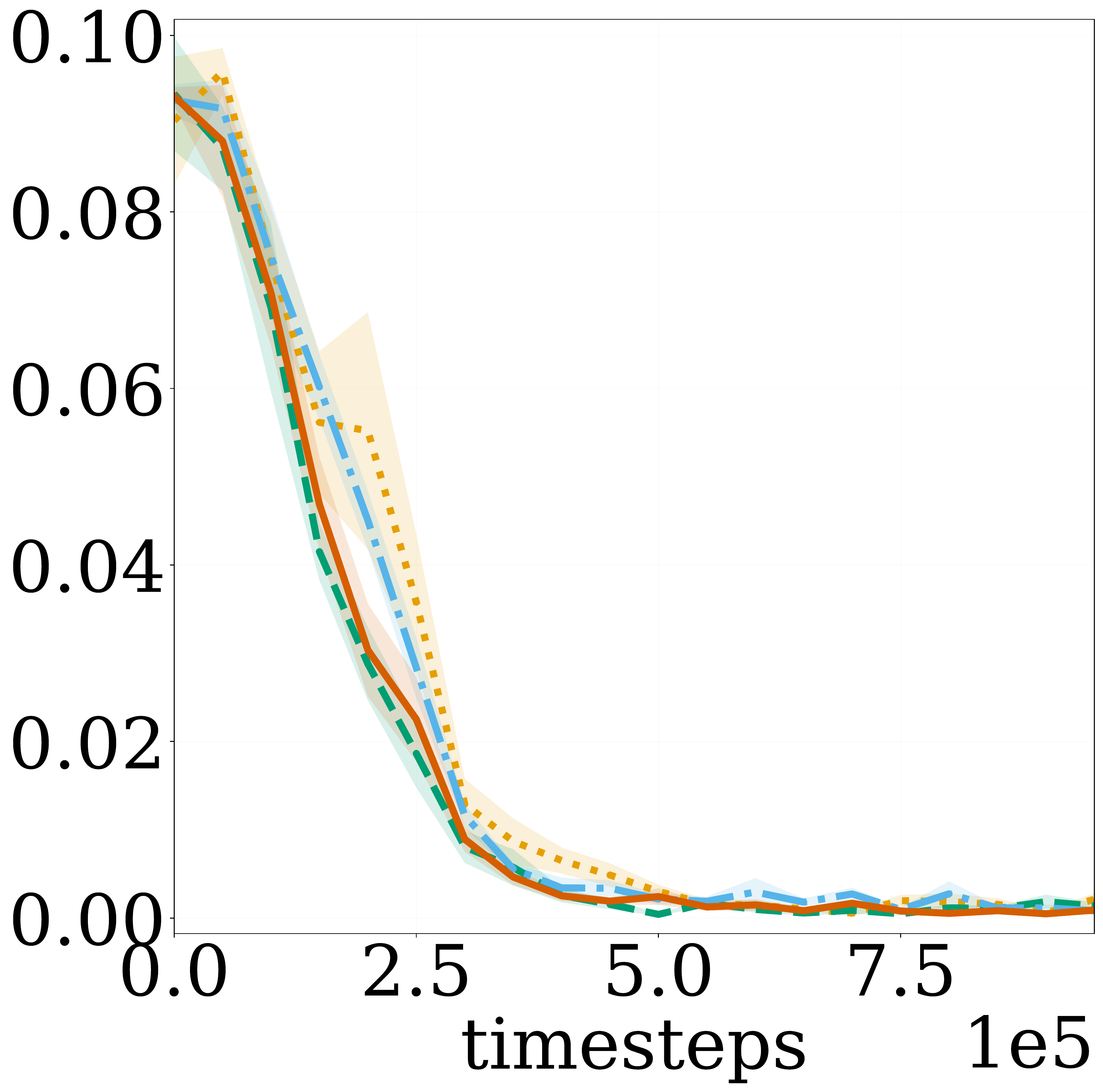}}
    \hspace{0cm}
    \subfigure[0.2]{\includegraphics[width=0.155\textwidth]{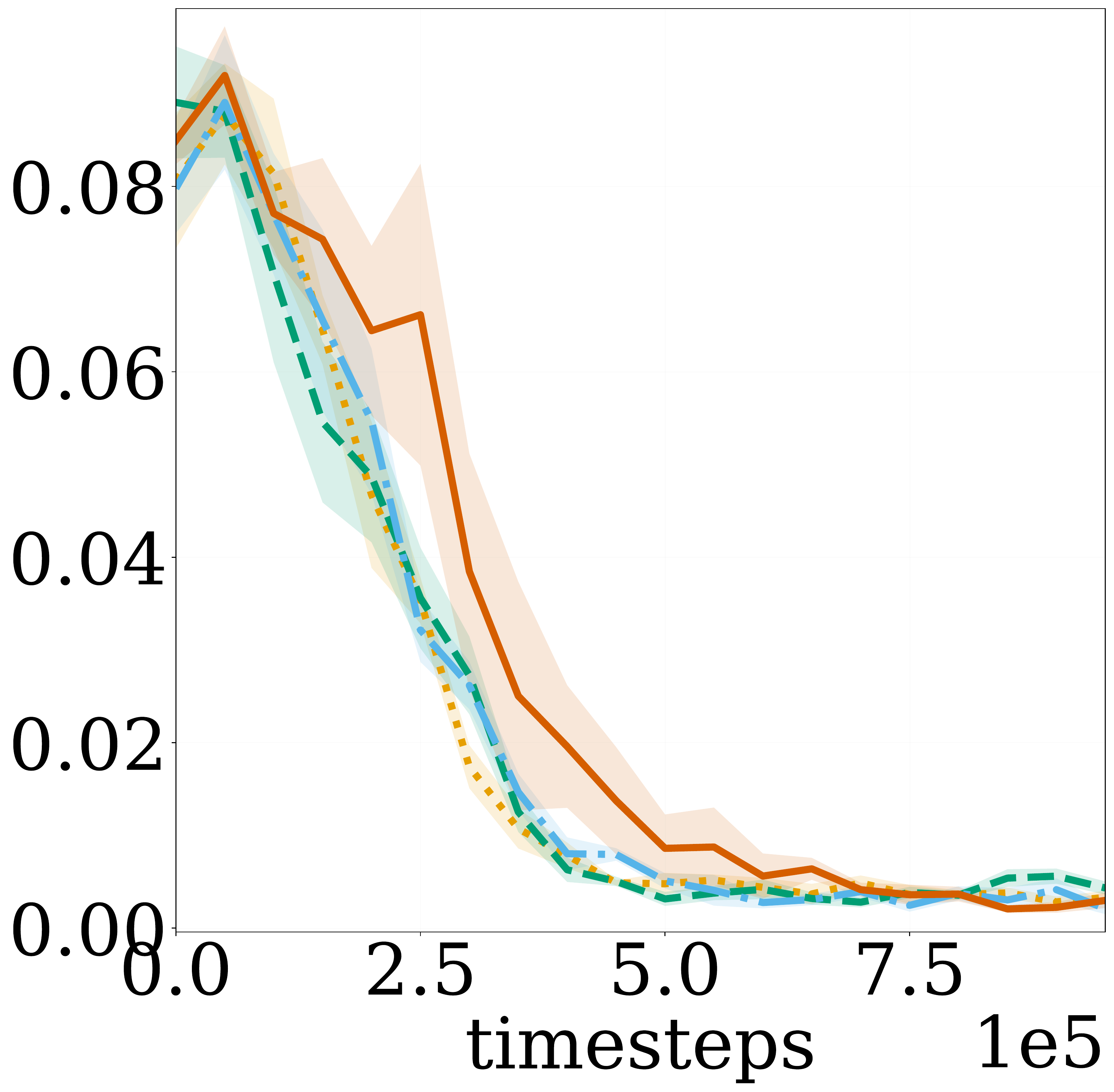}}
    \hspace{0cm}
    \subfigure[0.3]{\includegraphics[width=0.155\textwidth]{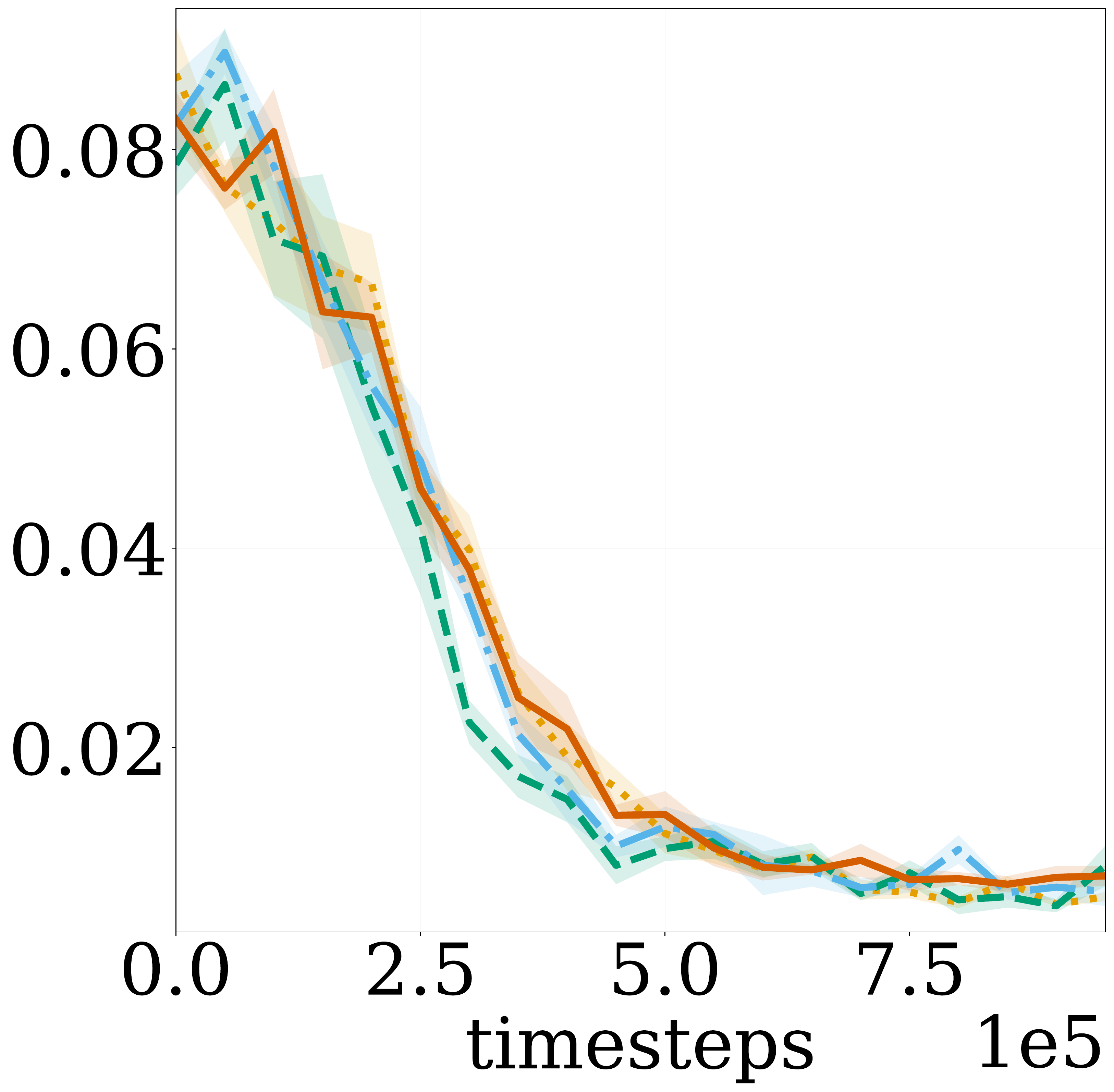}}
    \hspace{0cm}
    \subfigure[0.4]{\includegraphics[width=0.155\textwidth]{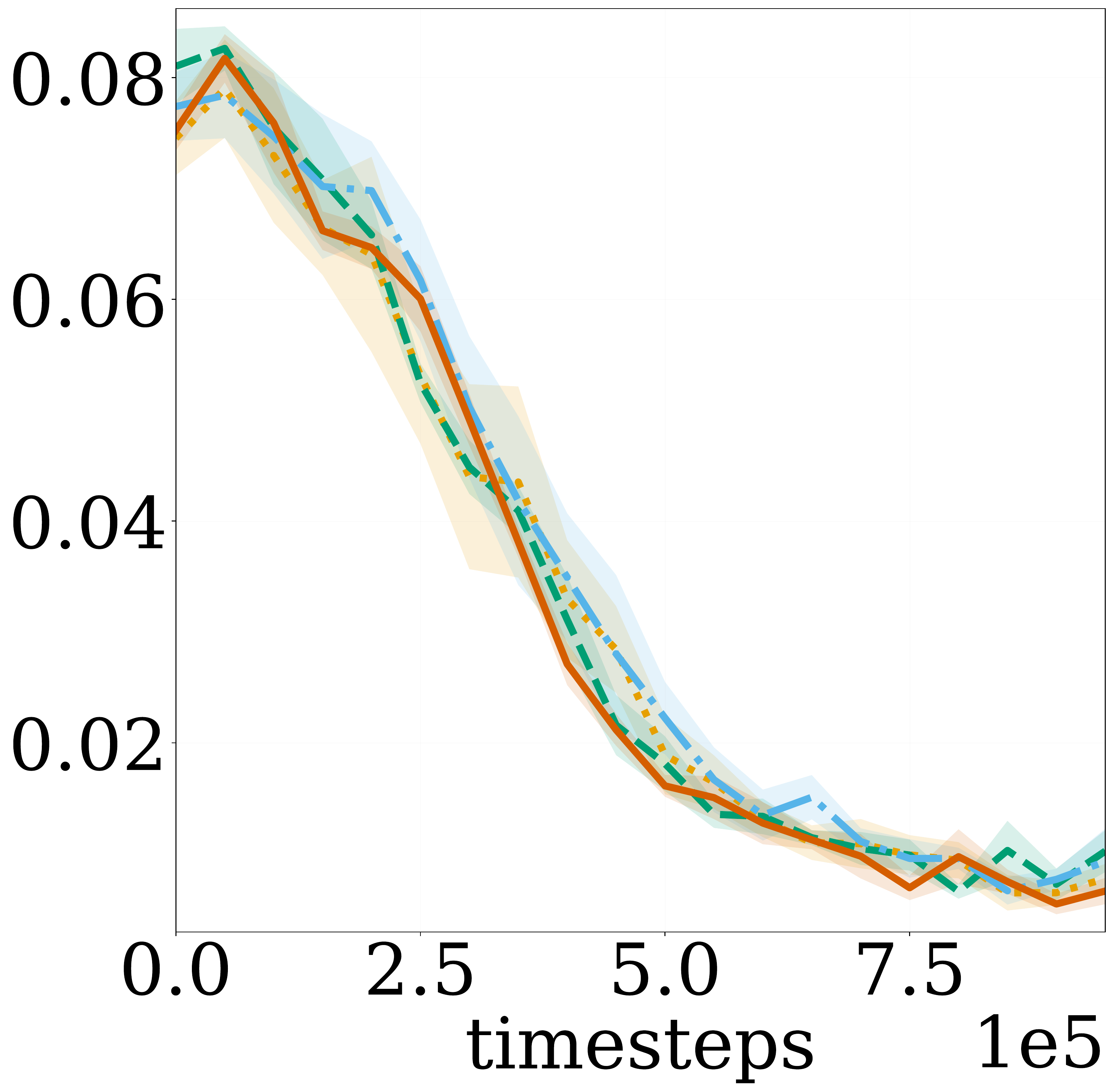}}
    \hspace{0cm}
    \subfigure[0.5]{\includegraphics[width=0.155\textwidth]{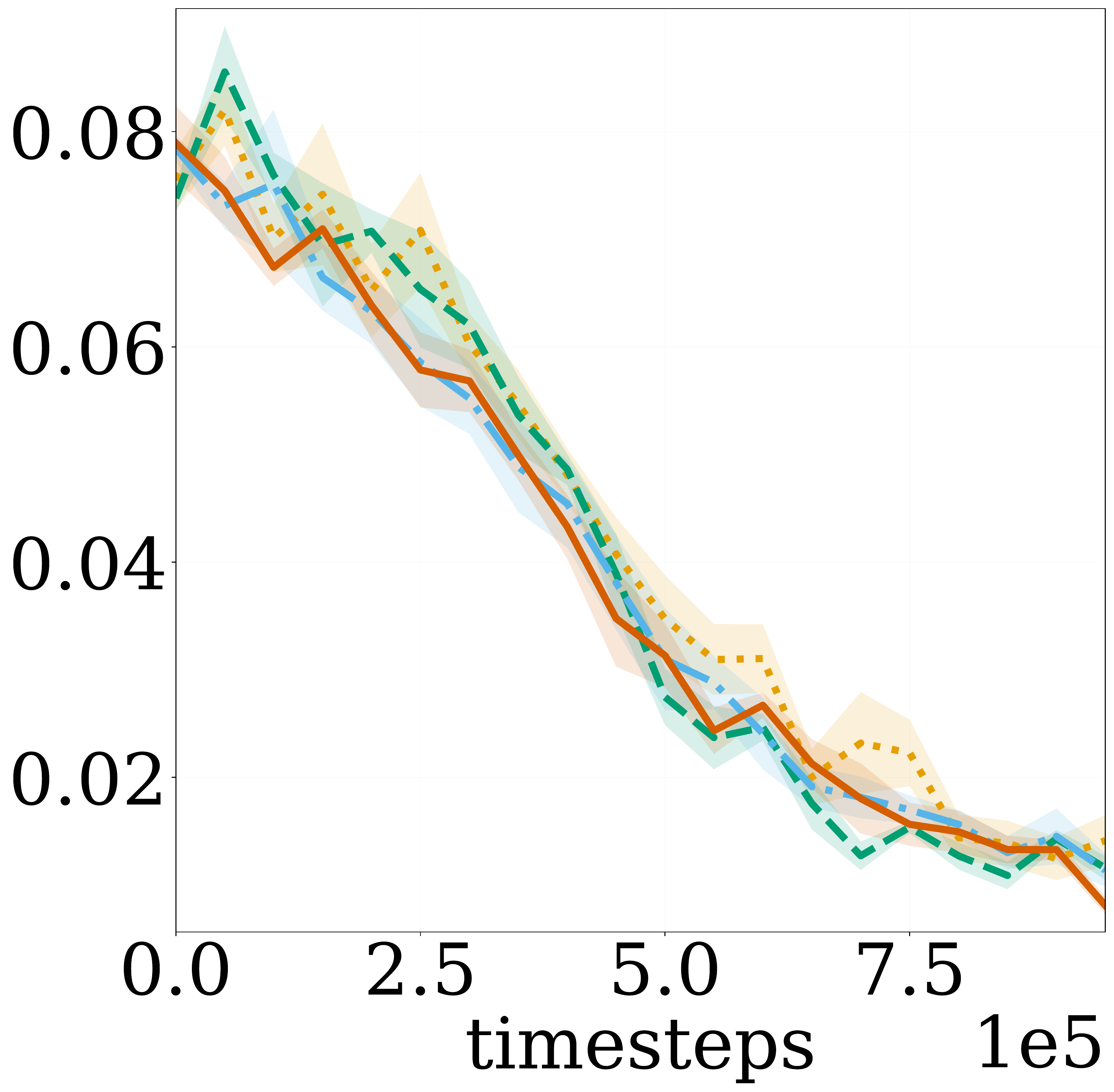}}
    \subfigure{\includegraphics[width=0.6\textwidth]{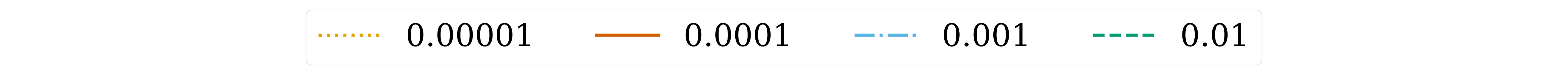}}\\
    \caption{Evaluation of our method in the gridworld environment for different values of $\beta$: reward and constraint violation rate of trajectories sampled from the nominal policy during training. Results are averaged over 5 random seeds. The x-axis is the number of timesteps taken in the environment during training. The shaded regions correspond with the standard error.}
    \label{fig:appendix-gridworld-beta-ablation}
\end{center}
%\vskip -0.2in
\end{figure*}

Figure~\ref{fig:appendix-gridworld-beta-ablation} depicts the reward and constraint violation rate during training at different timesteps for our method with various value of $\beta$ and for different rates of stochasticity.

\begin{figure*}[h!]
%\vskip -0.2in
\begin{center}    
    \subfigure{\includegraphics[width=0.17\textwidth]{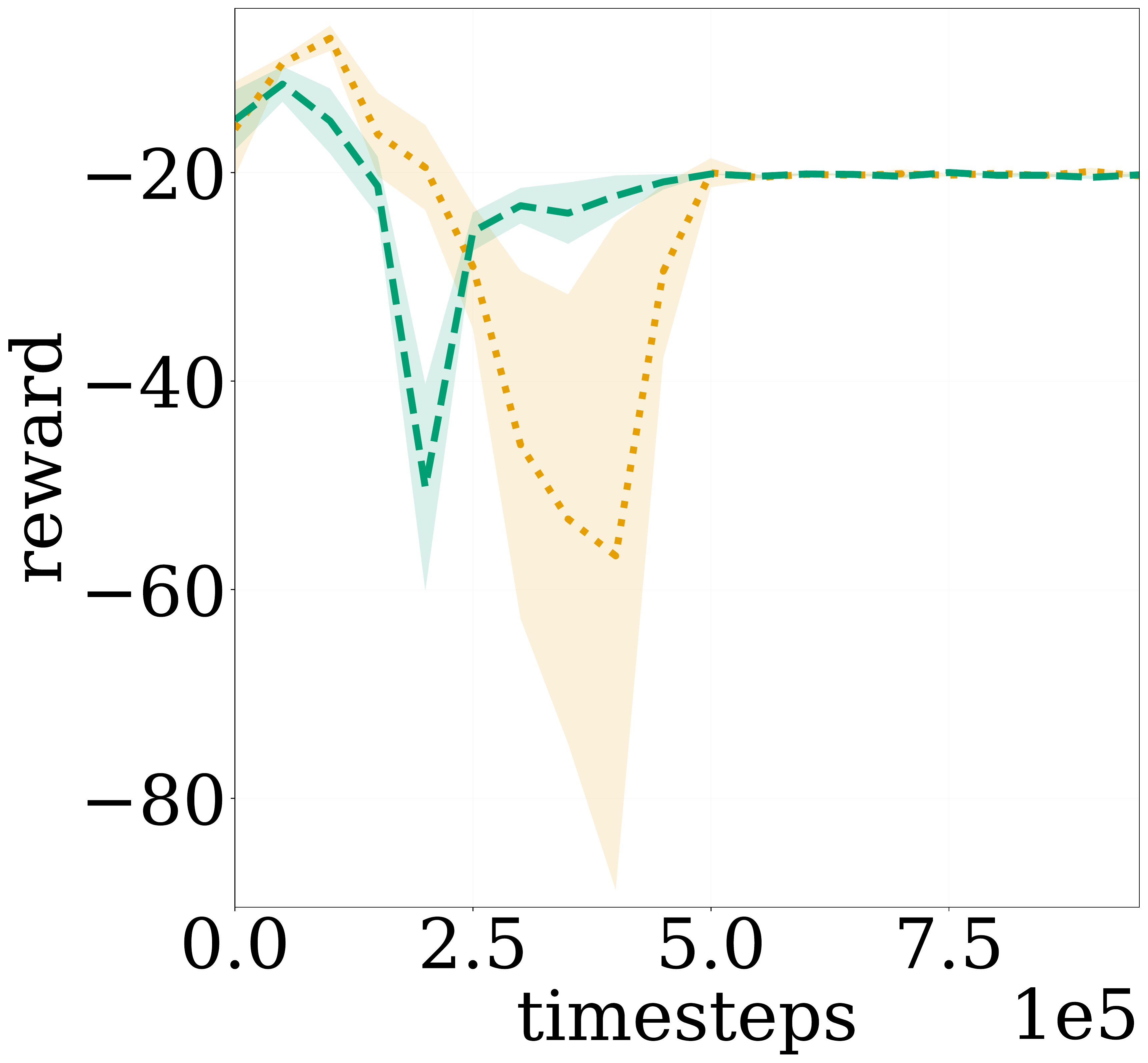}} %\hspace{0.2in}
    \subfigure{\includegraphics[width=0.16\textwidth]{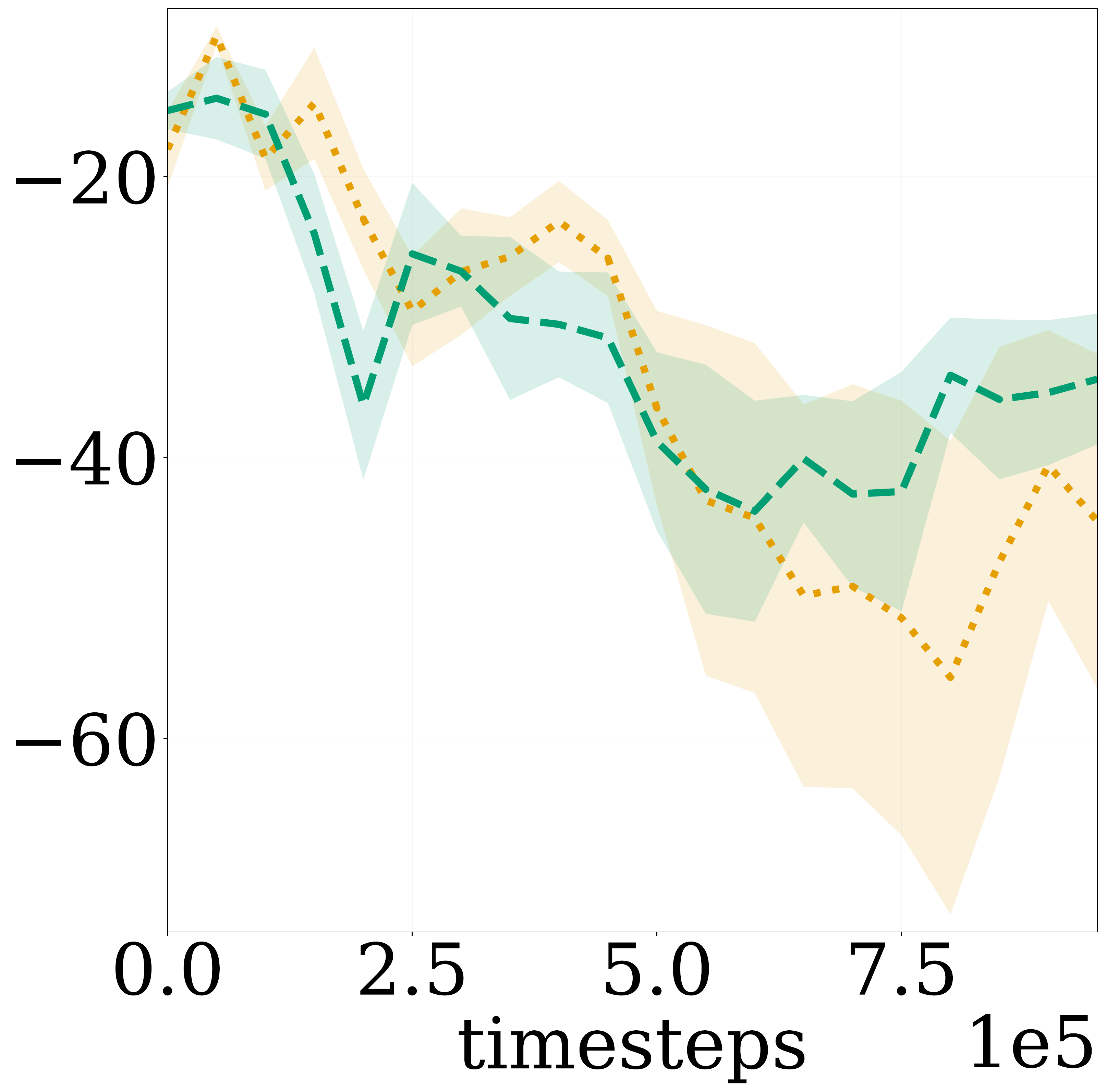}}
    \subfigure{\includegraphics[width=0.16\textwidth]{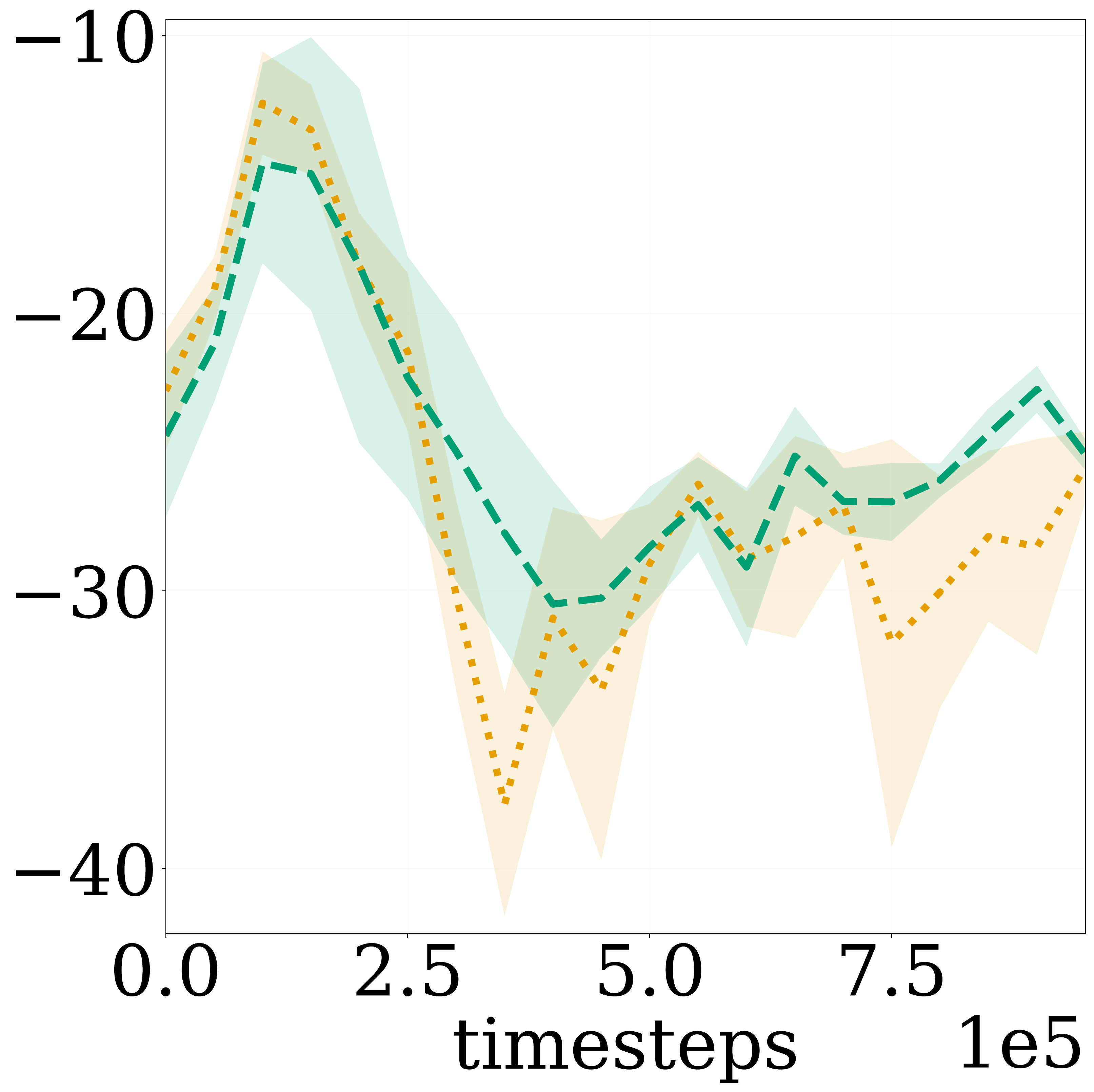}}
    \subfigure{\includegraphics[width=0.16\textwidth]{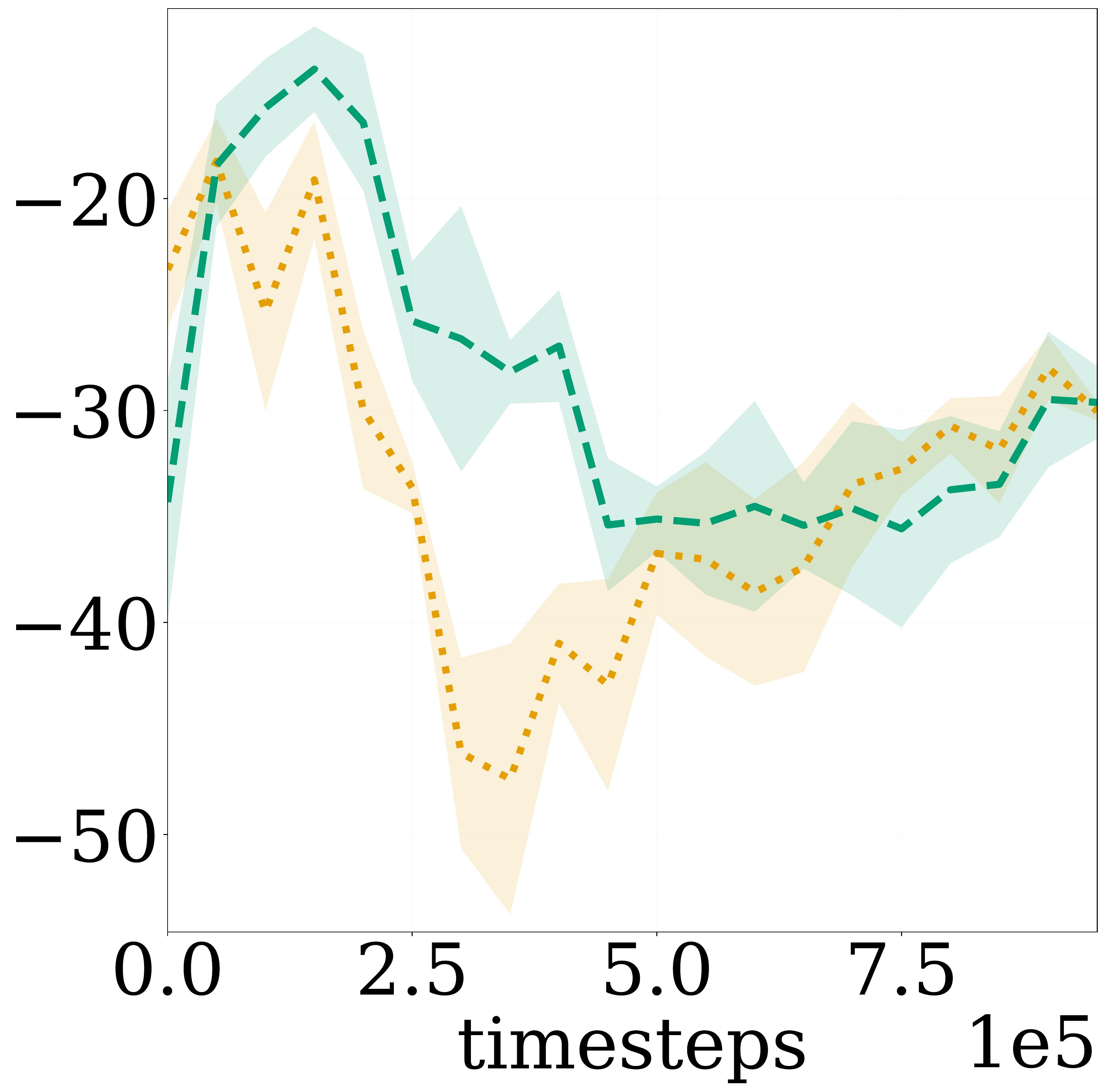}}
    \subfigure{\includegraphics[width=0.16\textwidth]{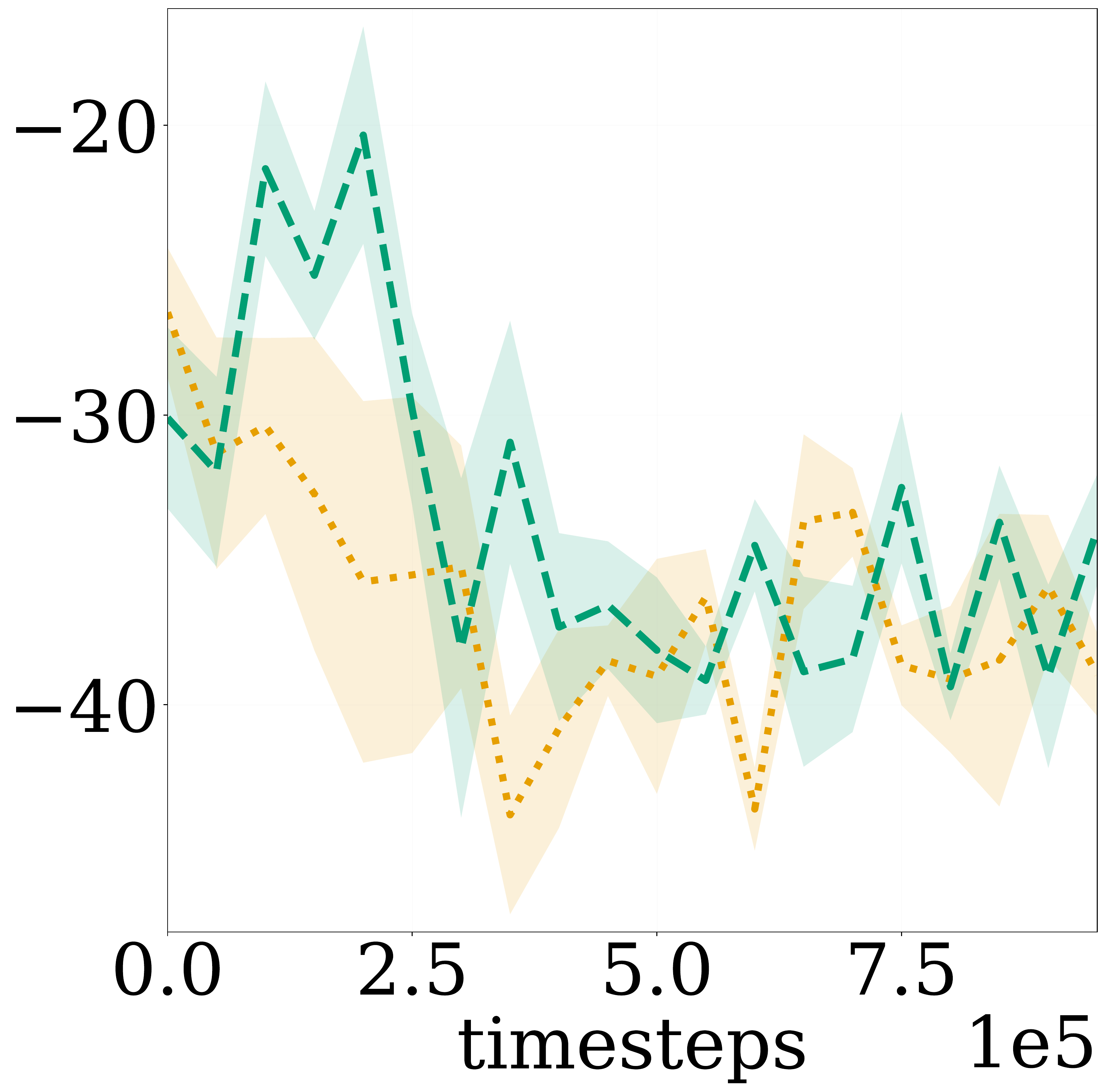}}
    \subfigure{\includegraphics[width=0.16\textwidth]{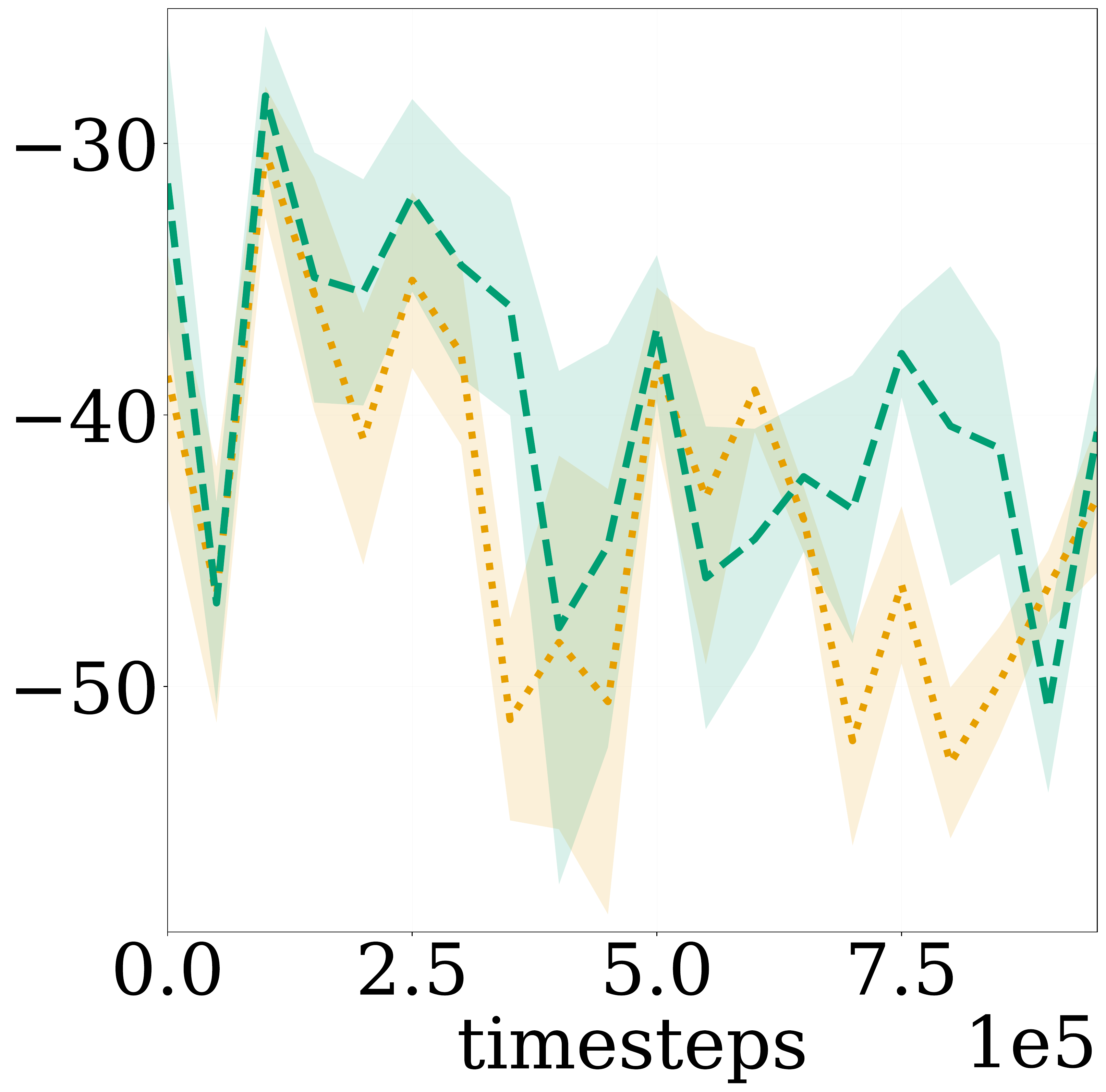}}\\
    \setcounter{subfigure}{0}
    \hspace{0.1cm}
    \subfigure[0.0]{\includegraphics[width=0.155\textwidth]{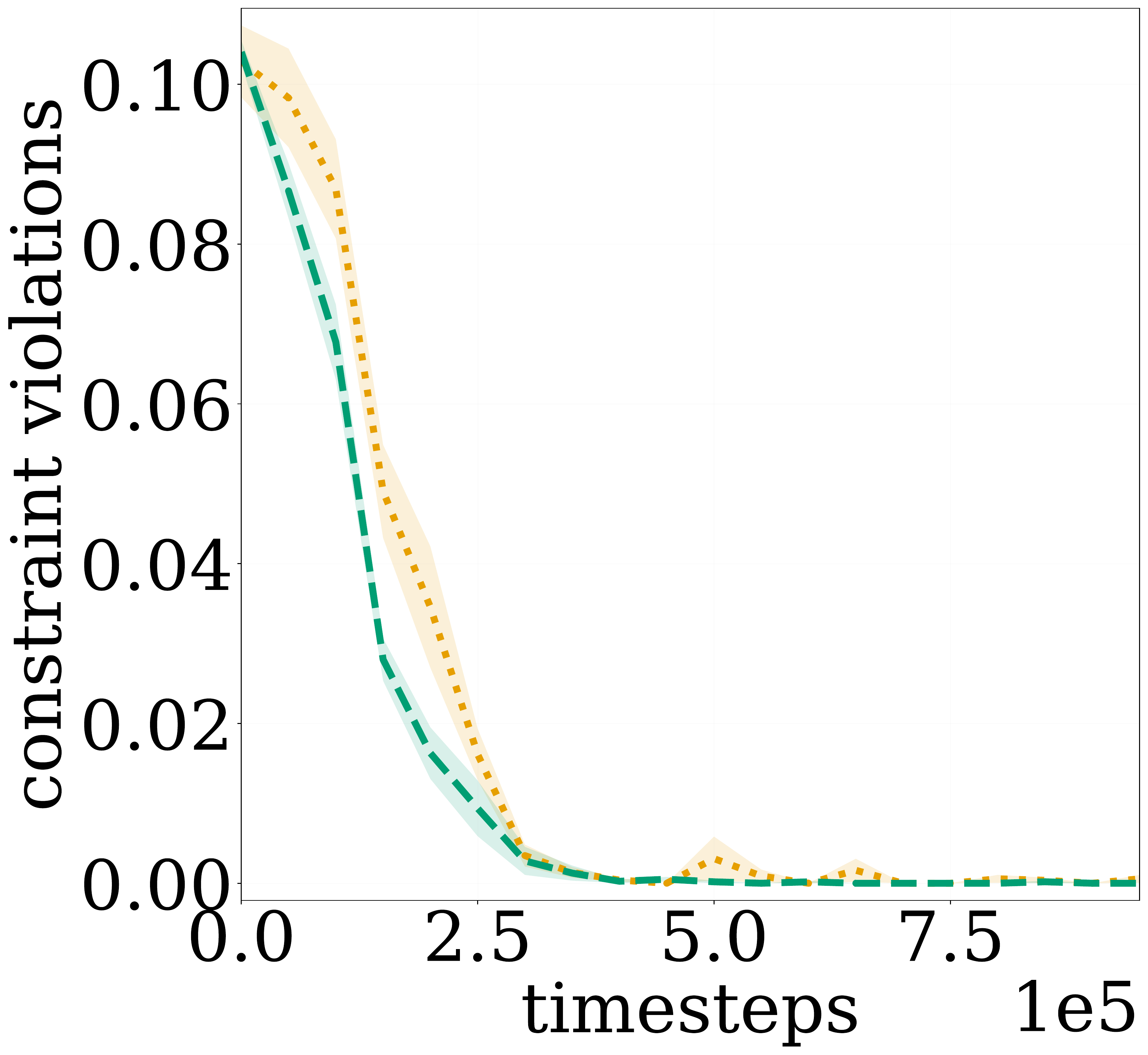}} 
    \hspace{0cm}
    \subfigure[0.1]{\includegraphics[width=0.155\textwidth]{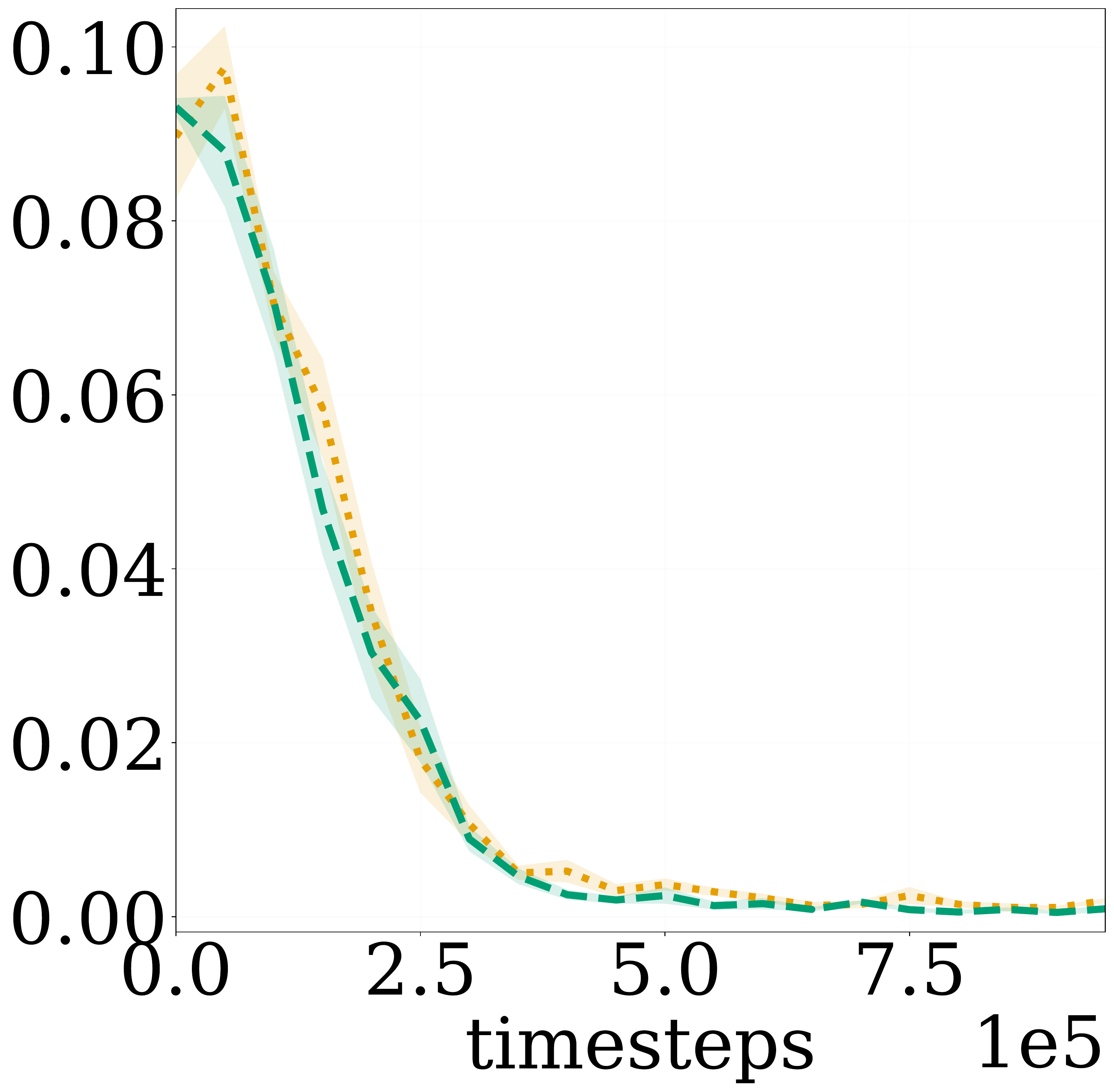}}
    \hspace{0cm}
    \subfigure[0.2]{\includegraphics[width=0.155\textwidth]{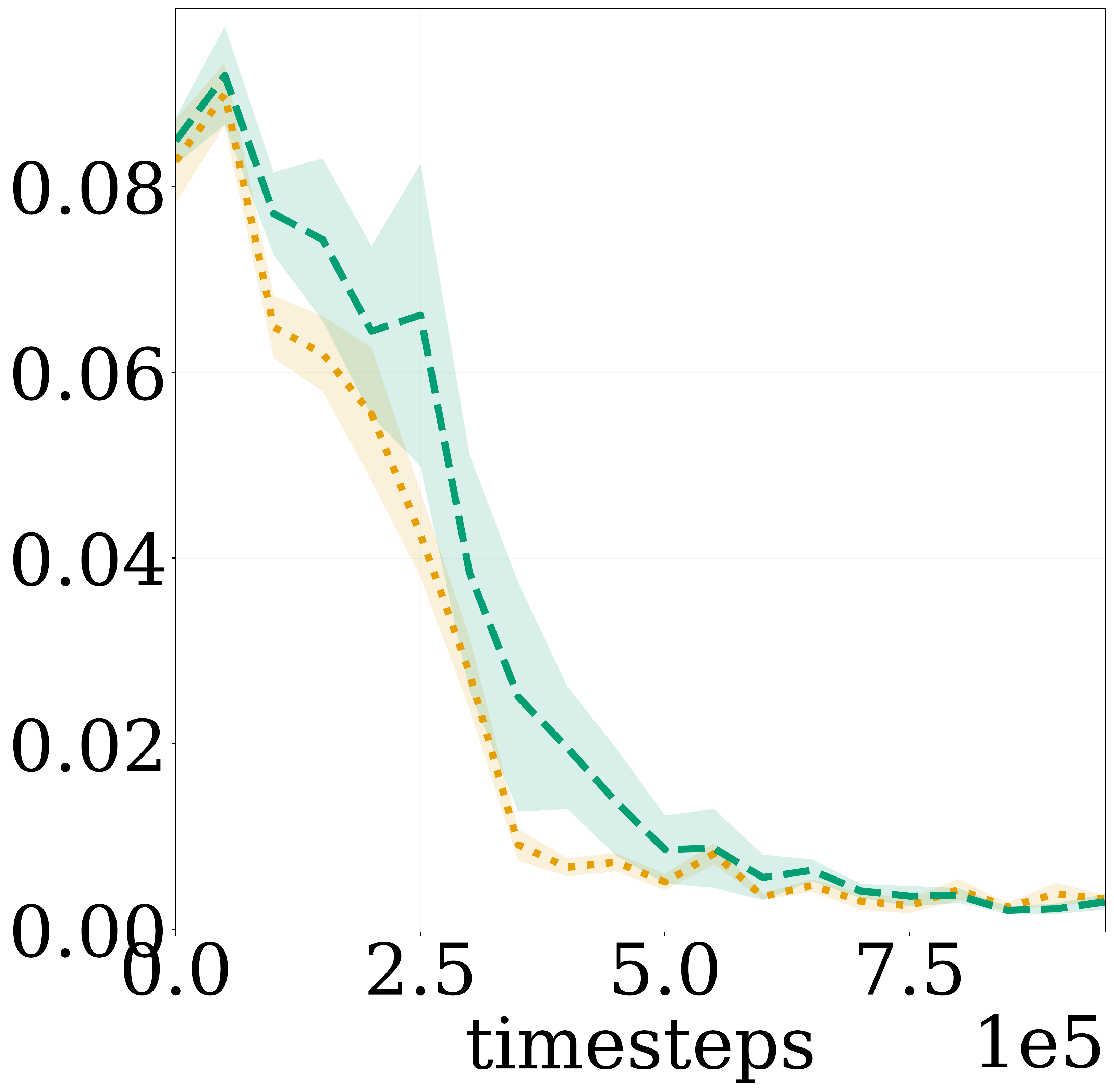}}
    \hspace{0cm}
    \subfigure[0.3]{\includegraphics[width=0.155\textwidth]{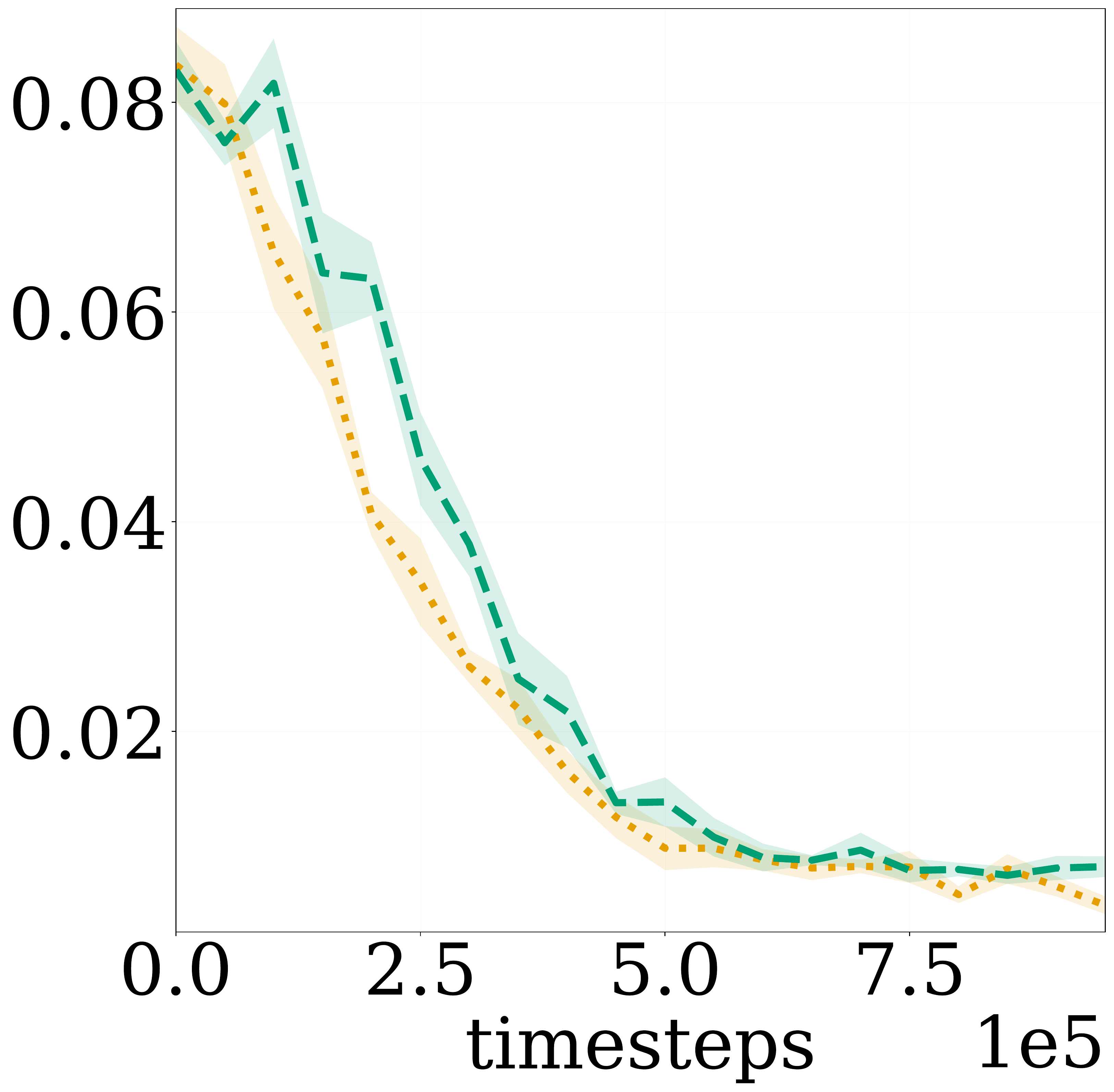}}
    \hspace{0cm}
    \subfigure[0.4]{\includegraphics[width=0.155\textwidth]{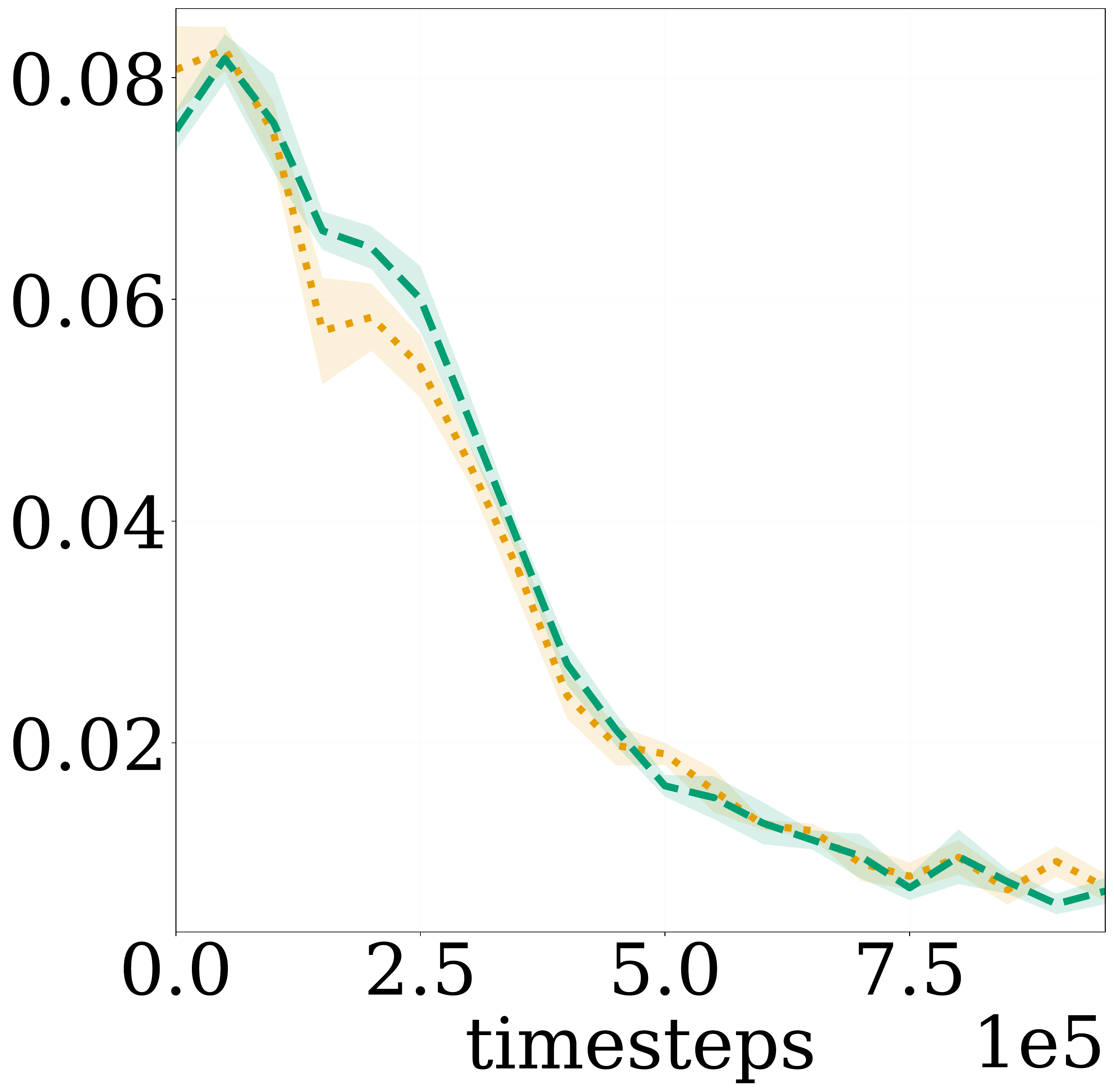}}
    \hspace{0cm}
    \subfigure[0.5]{\includegraphics[width=0.155\textwidth]{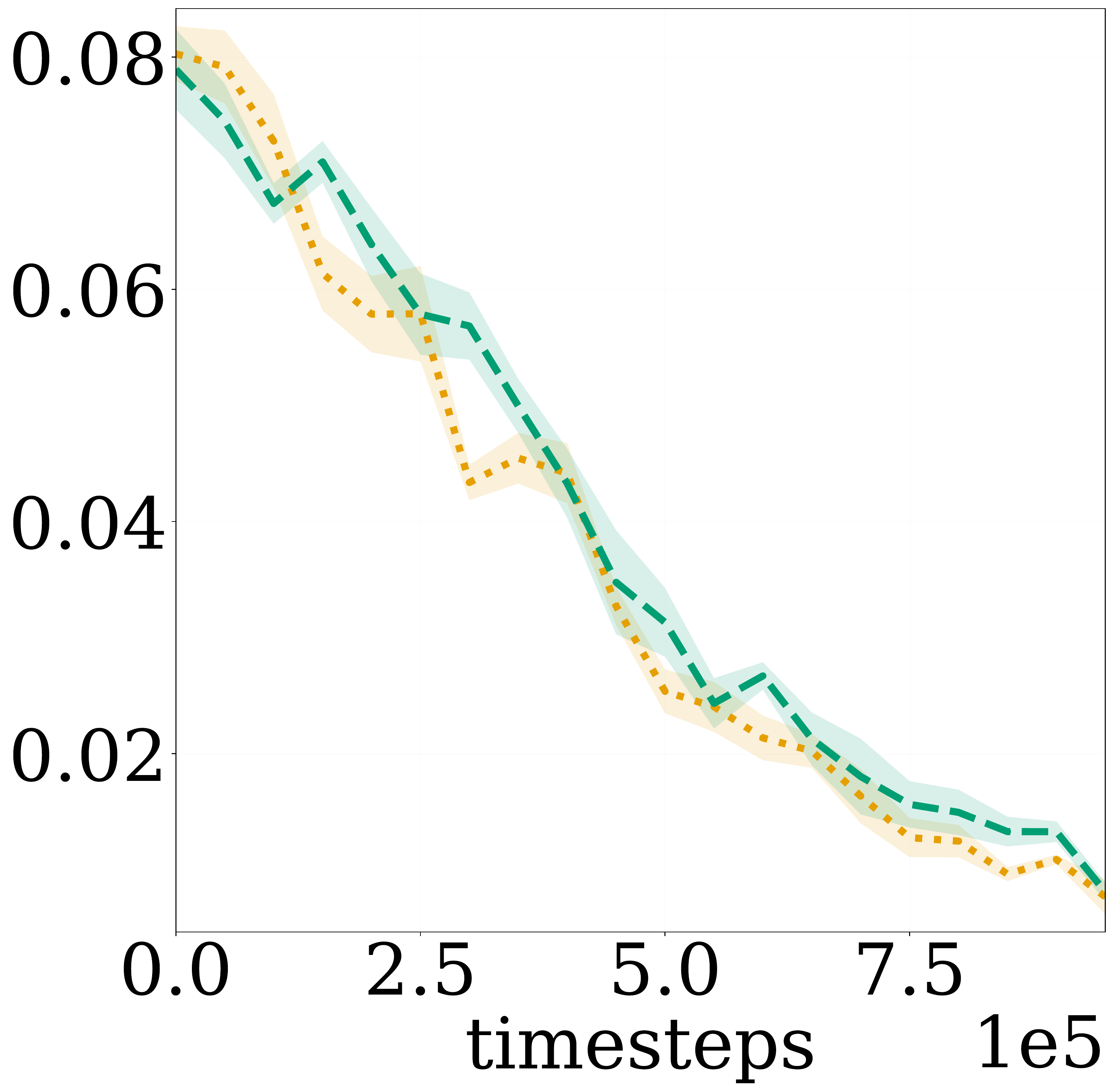}}
    \subfigure{\includegraphics[width=0.6\textwidth]{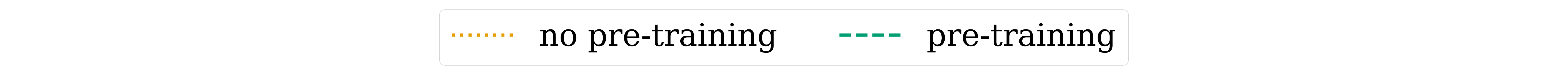}}\\
    \caption{Evaluation of our method in the gridworld environment for for feature encoder pre-training disabled: reward and constraint violation rate of trajectories sampled from the nominal policy during training. Results are averaged over 5 random seeds. The x-axis is the number of timesteps taken in the environment during training. The shaded regions correspond with the standard error.}
    \label{fig:appendix-gridworld-bootstrap-ablation}
\end{center}
%\vskip -0.2in
\end{figure*}

Figure~\ref{fig:appendix-gridworld-bootstrap-ablation} depicts the reward received and the constraint violation rate during training for the ablation study on the pre-training of the feature encoder. 
\clearpage

\subsection{Virtual Robotics Environments}
\begin{figure*}[h!]
%\vskip 0.2in
\begin{center}
    \hspace{0.01cm}
    \subfigure{\includegraphics[width=0.18\textwidth]{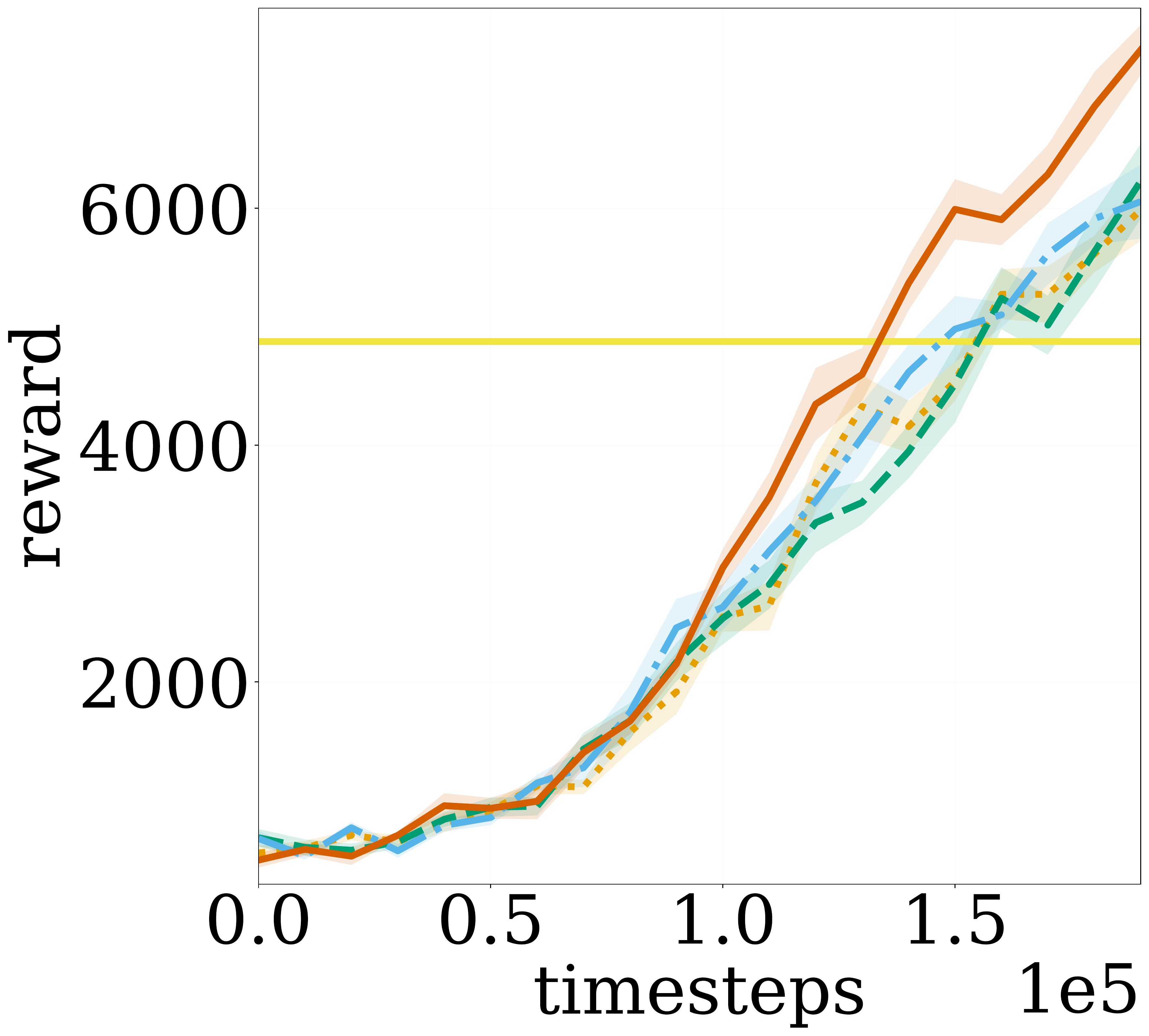}} %\hspace{0.2in}
    \subfigure{\includegraphics[width=0.17\textwidth]{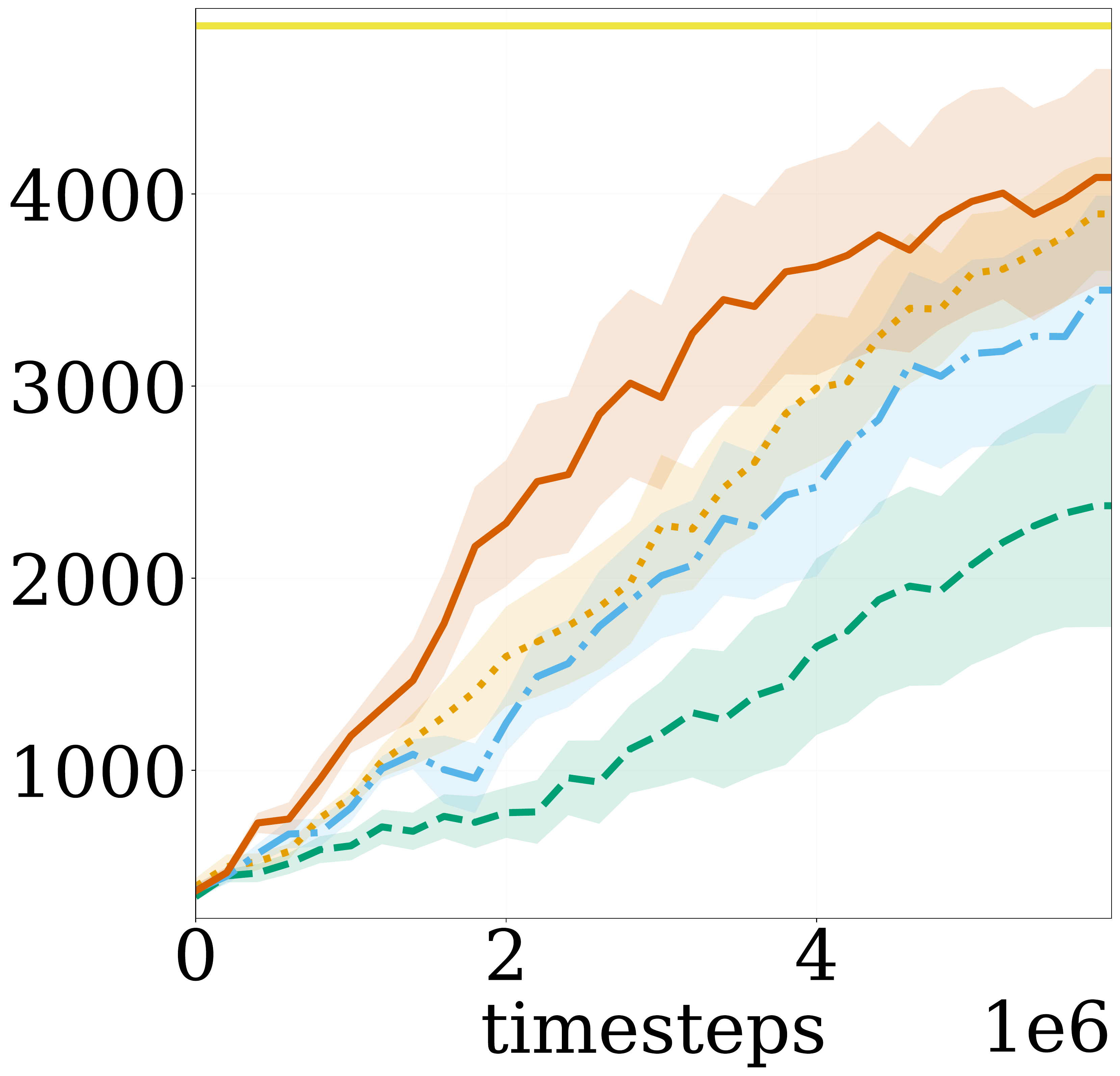}}
    \subfigure{\includegraphics[width=0.165\textwidth]{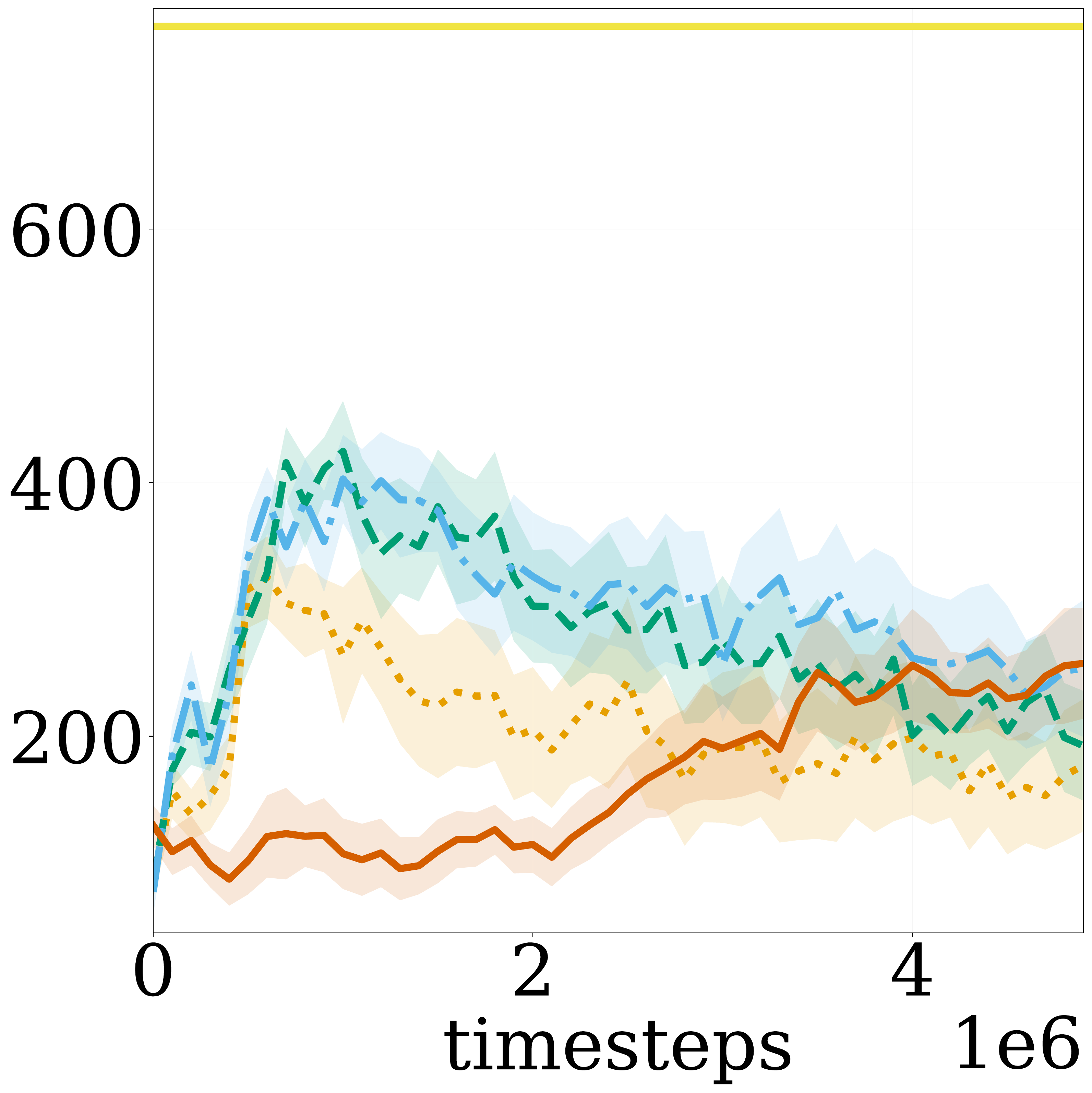}}
    \subfigure{\includegraphics[width=0.172\textwidth]{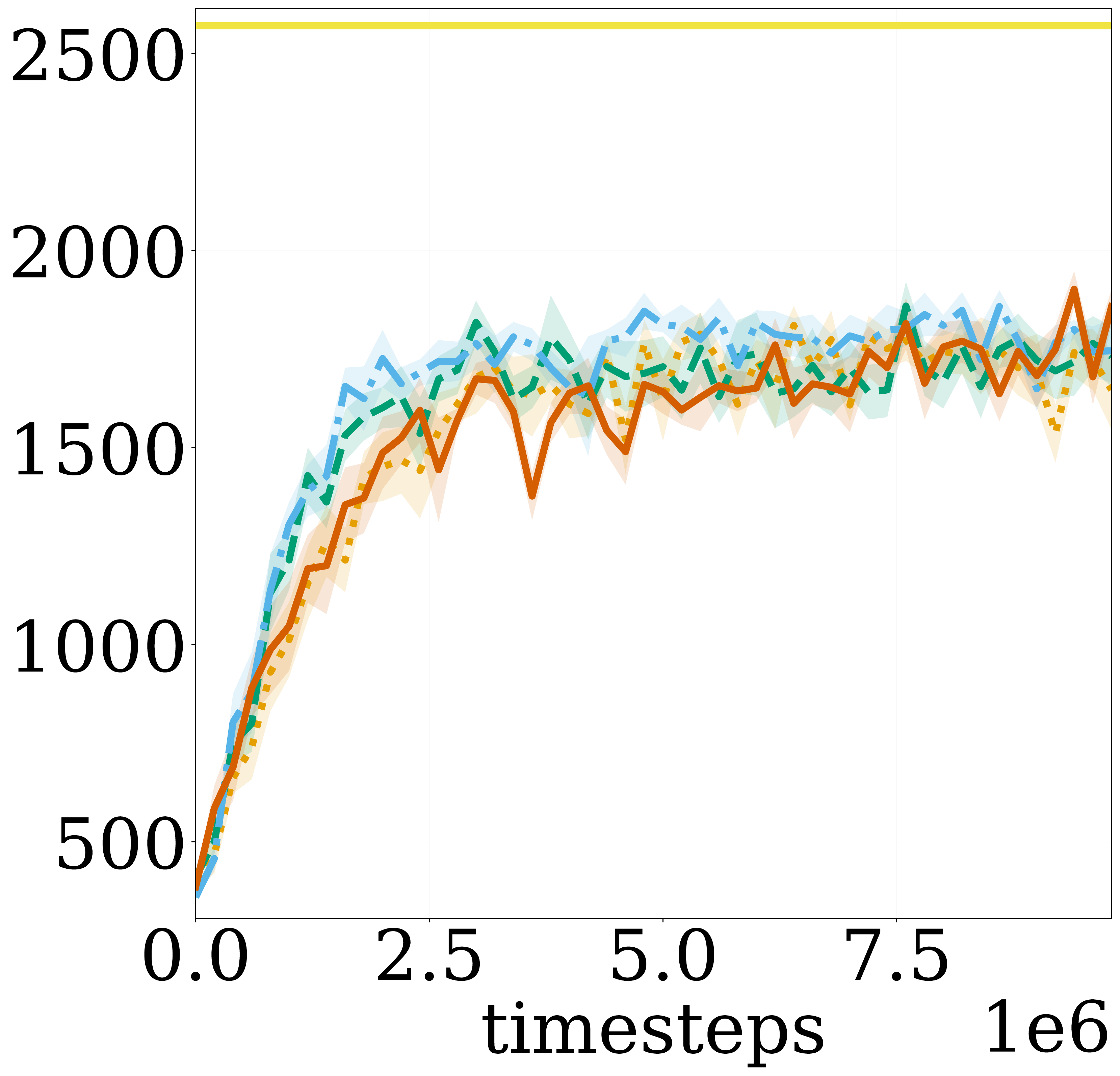}}
    \subfigure{\includegraphics[width=0.16\textwidth]{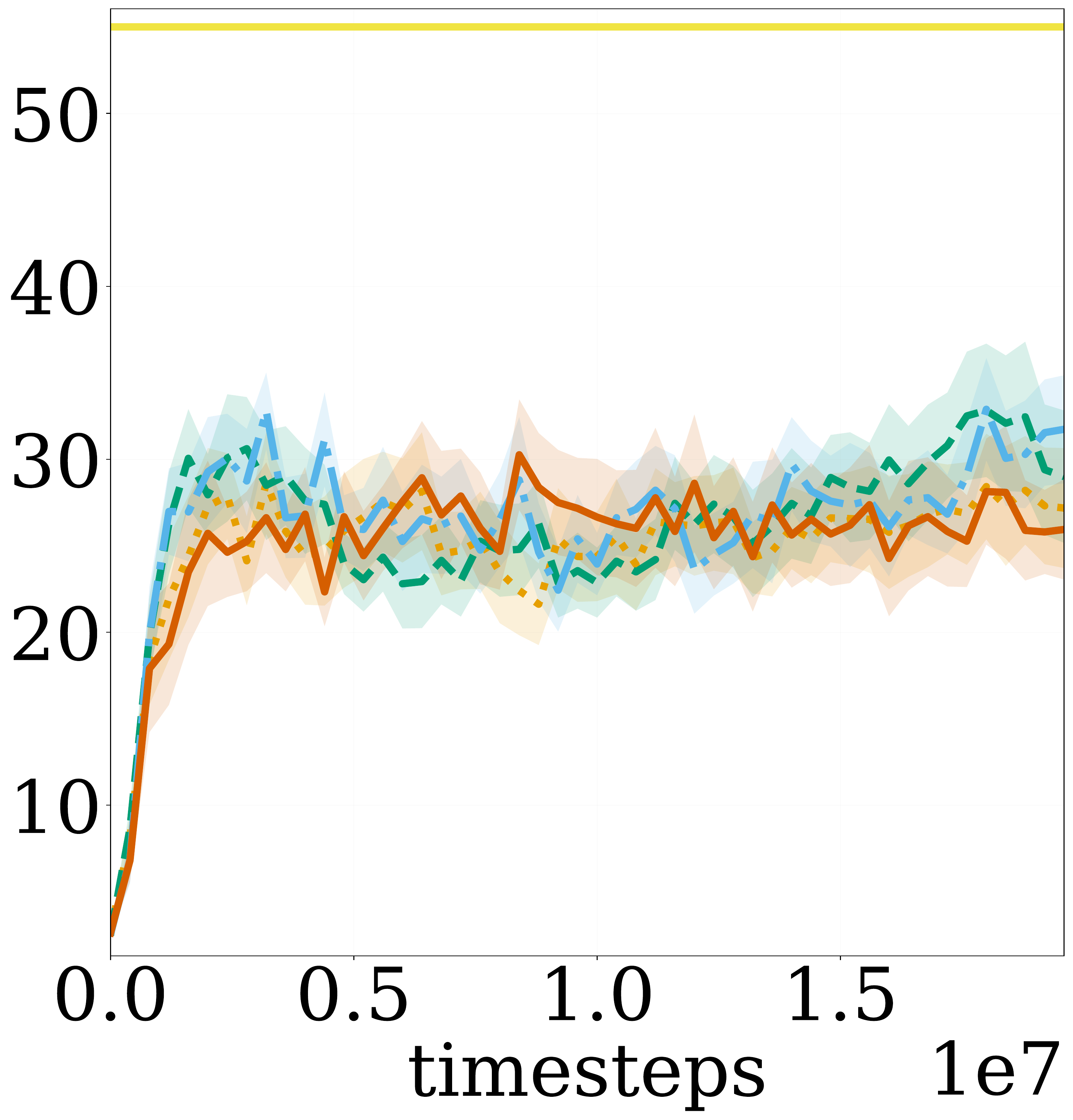}}\\
    \setcounter{subfigure}{0}
    \subfigure[Ant]{\includegraphics[width=0.185\textwidth]{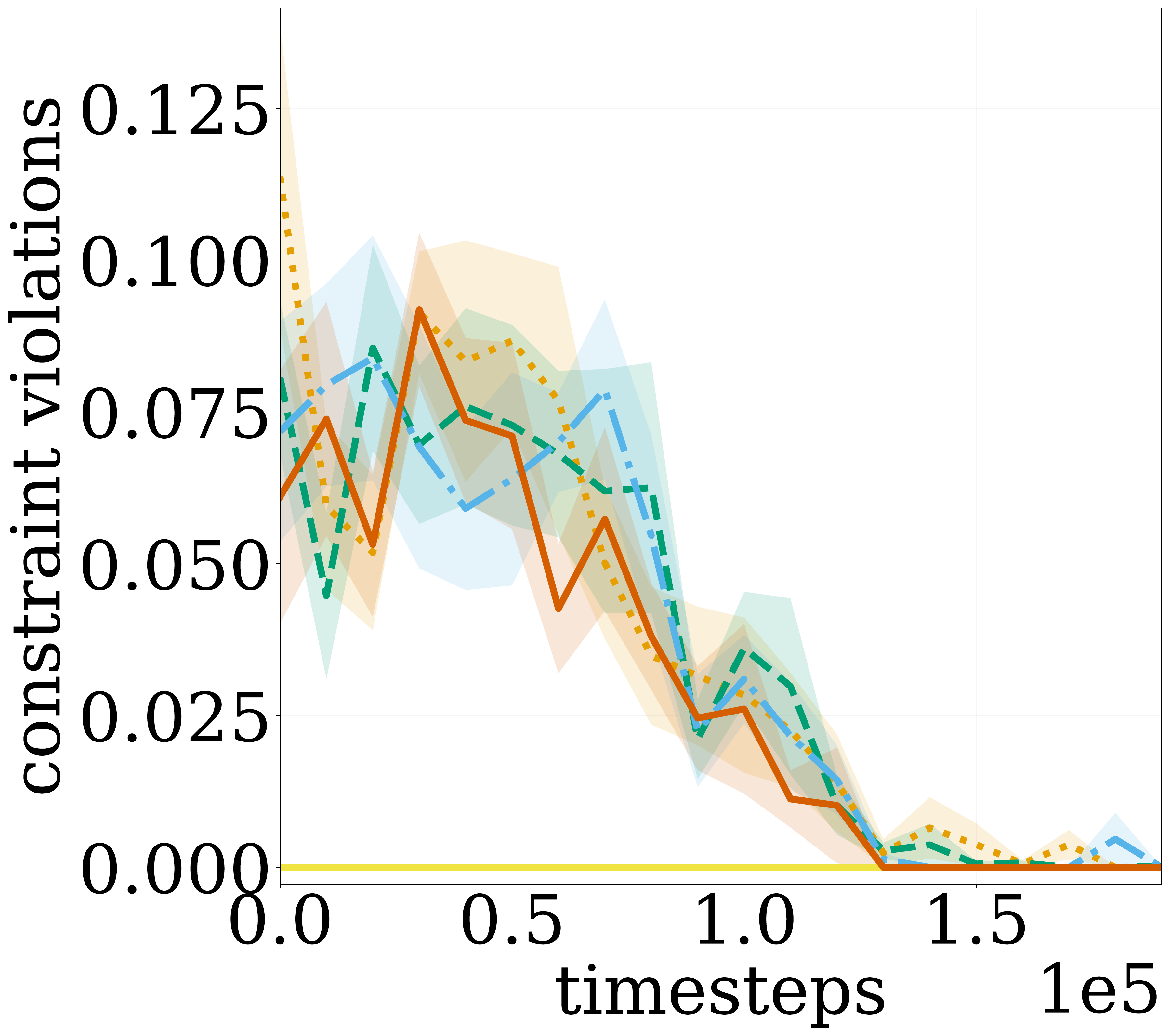}} 
    \hspace{0.12cm}
    \subfigure[Half-cheetah]{\includegraphics[width=0.16\textwidth]{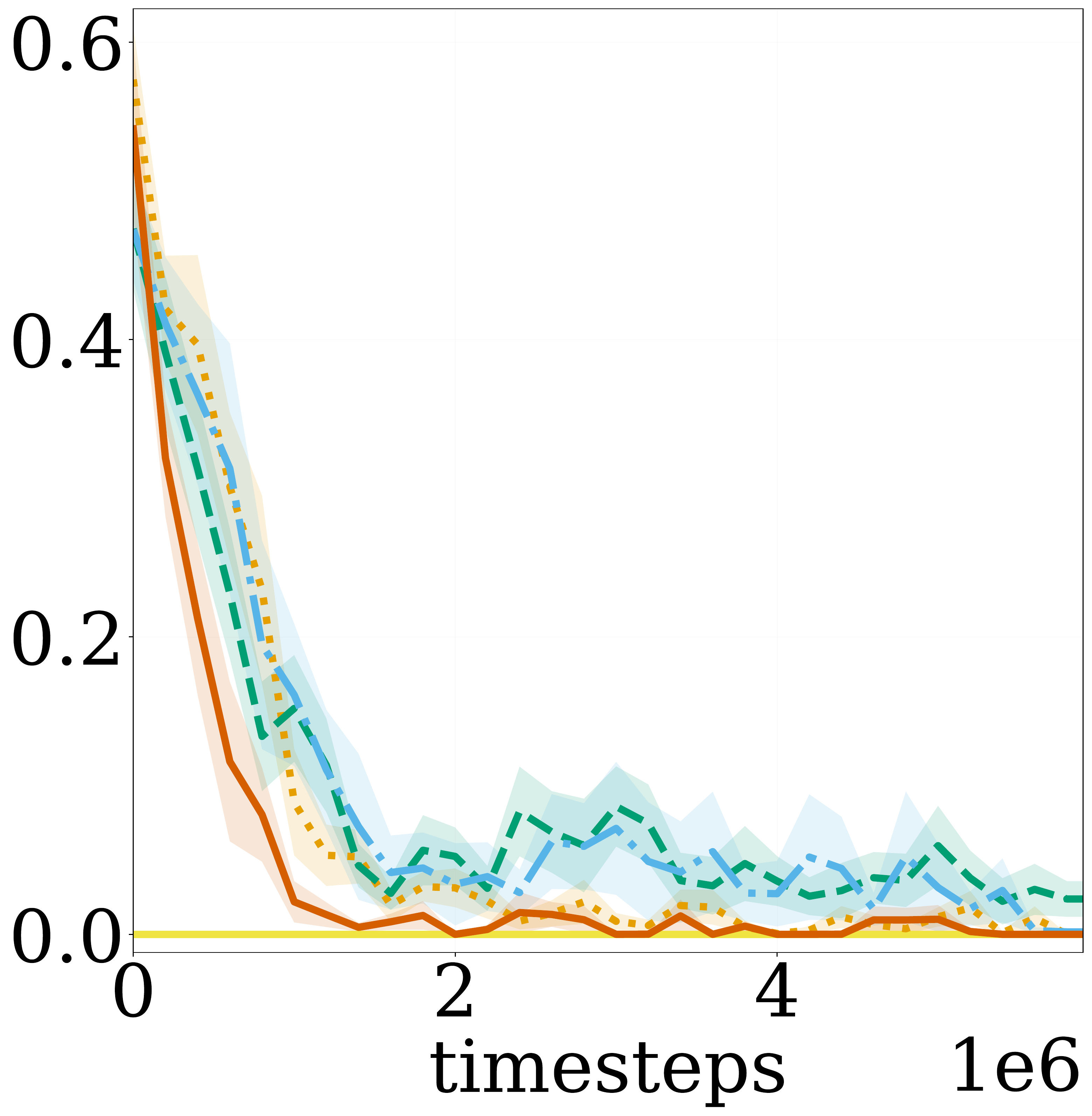}}
    \hspace{0cm}
    \subfigure[Swimmer]{\includegraphics[width=0.16\textwidth]{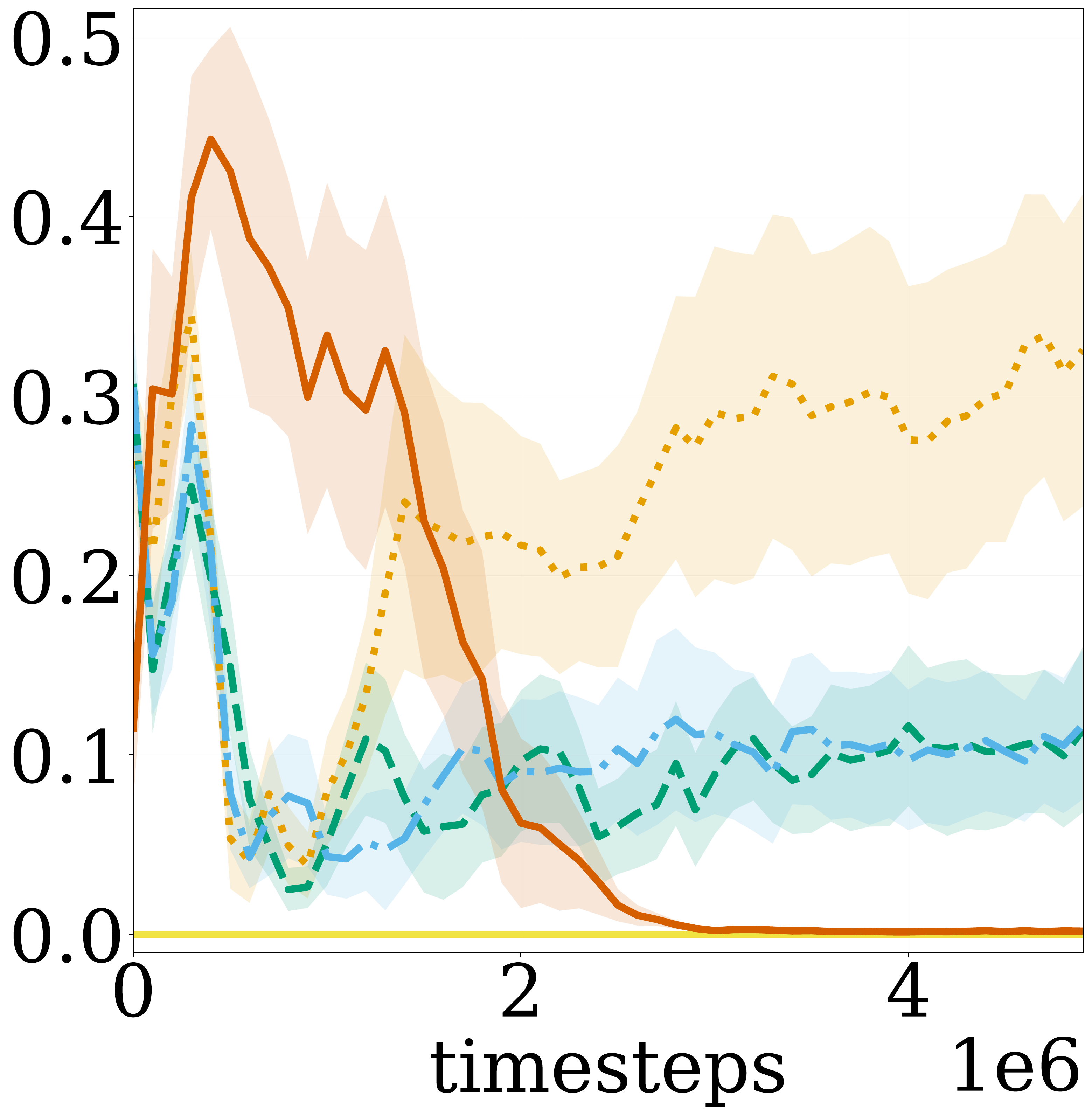}}
    \hspace{0cm}
    \subfigure[Walker]{\includegraphics[width=0.171\textwidth]{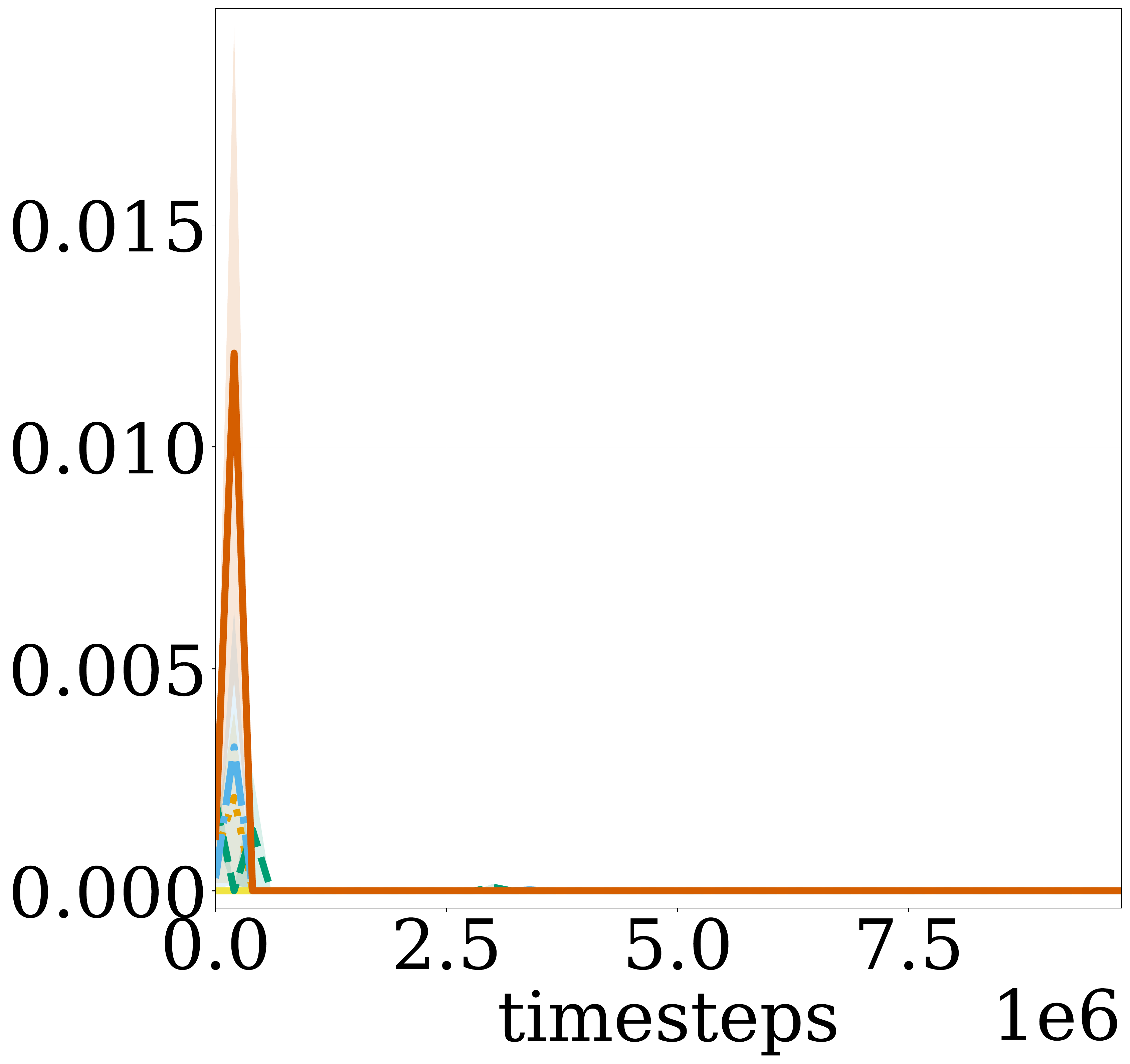}}
    \subfigure[Inverted pendulum]{\includegraphics[width=0.16\textwidth]{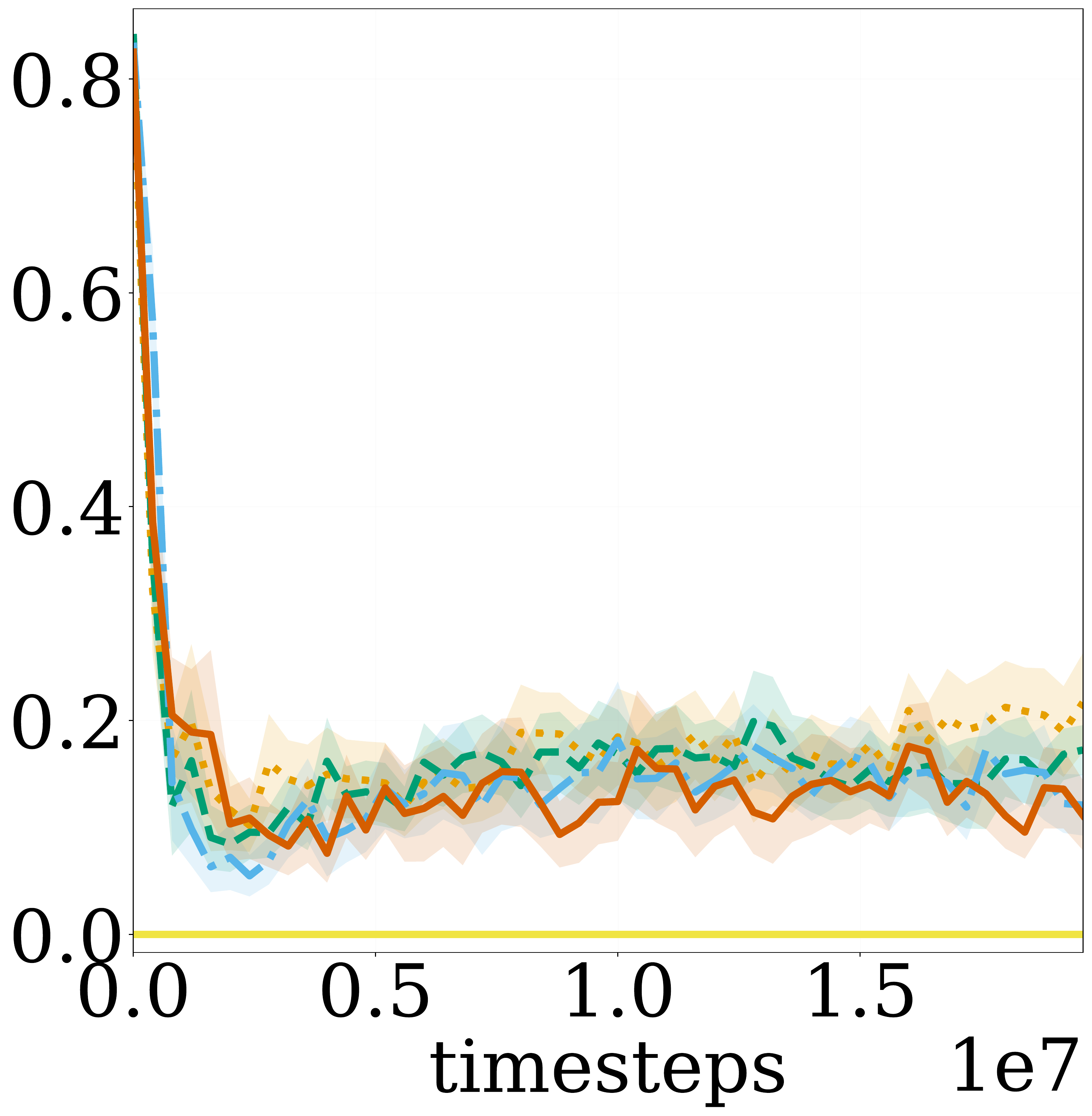}}
    \subfigure{\includegraphics[width=0.6\textwidth]{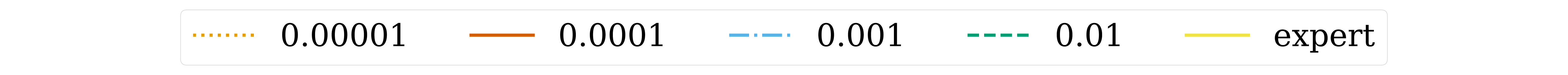}}\\
    \caption{Evaluation of our method in the virtual robotics environments for different values of $\beta$: reward and constraint violation rate of trajectories sampled from the nominal policy during training. Results are averaged over 10 random seeds. The x-axis is the number of timesteps taken in the environment during training. The shaded regions correspond with the standard error.}
    \label{fig:resultsmujoco-ablations}
\end{center}
\end{figure*}

\begin{figure*}[h!]
\vskip -0.8cm
\begin{center}
    \hspace{0.01cm}
    \subfigure{\includegraphics[width=0.18\textwidth]{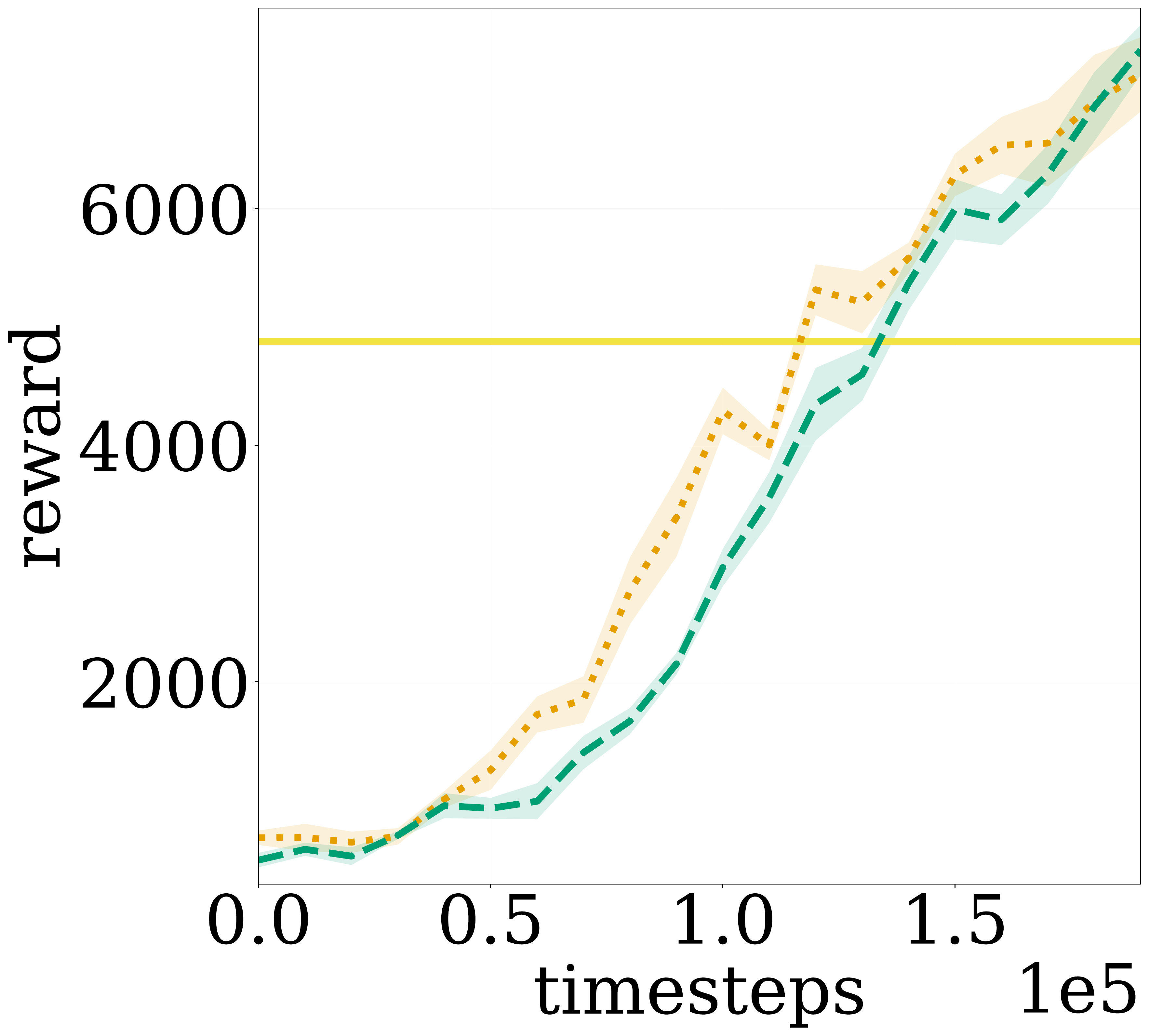}} %\hspace{0.2in}
    \subfigure{\includegraphics[width=0.17\textwidth]{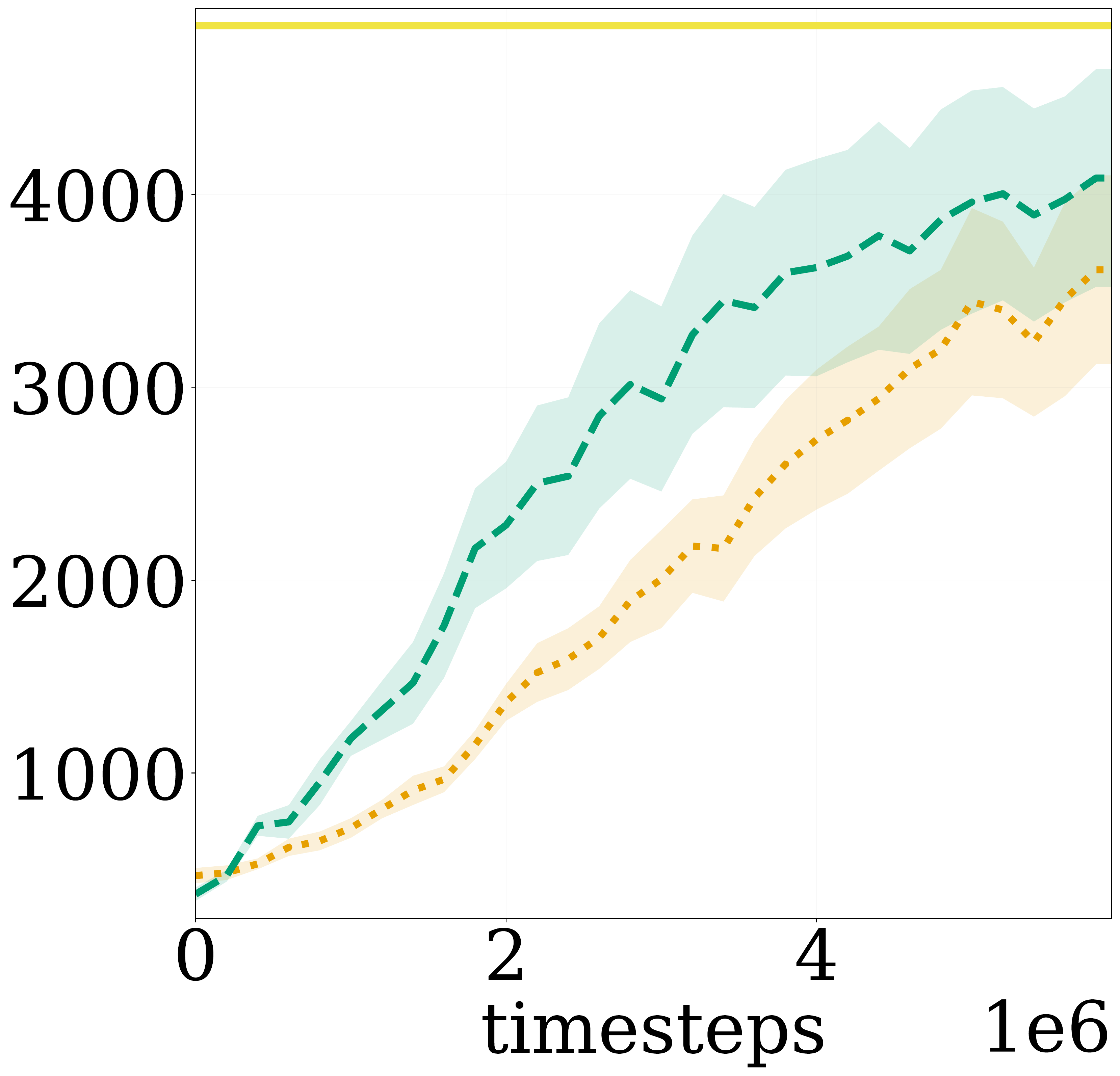}}
    \subfigure{\includegraphics[width=0.165\textwidth]{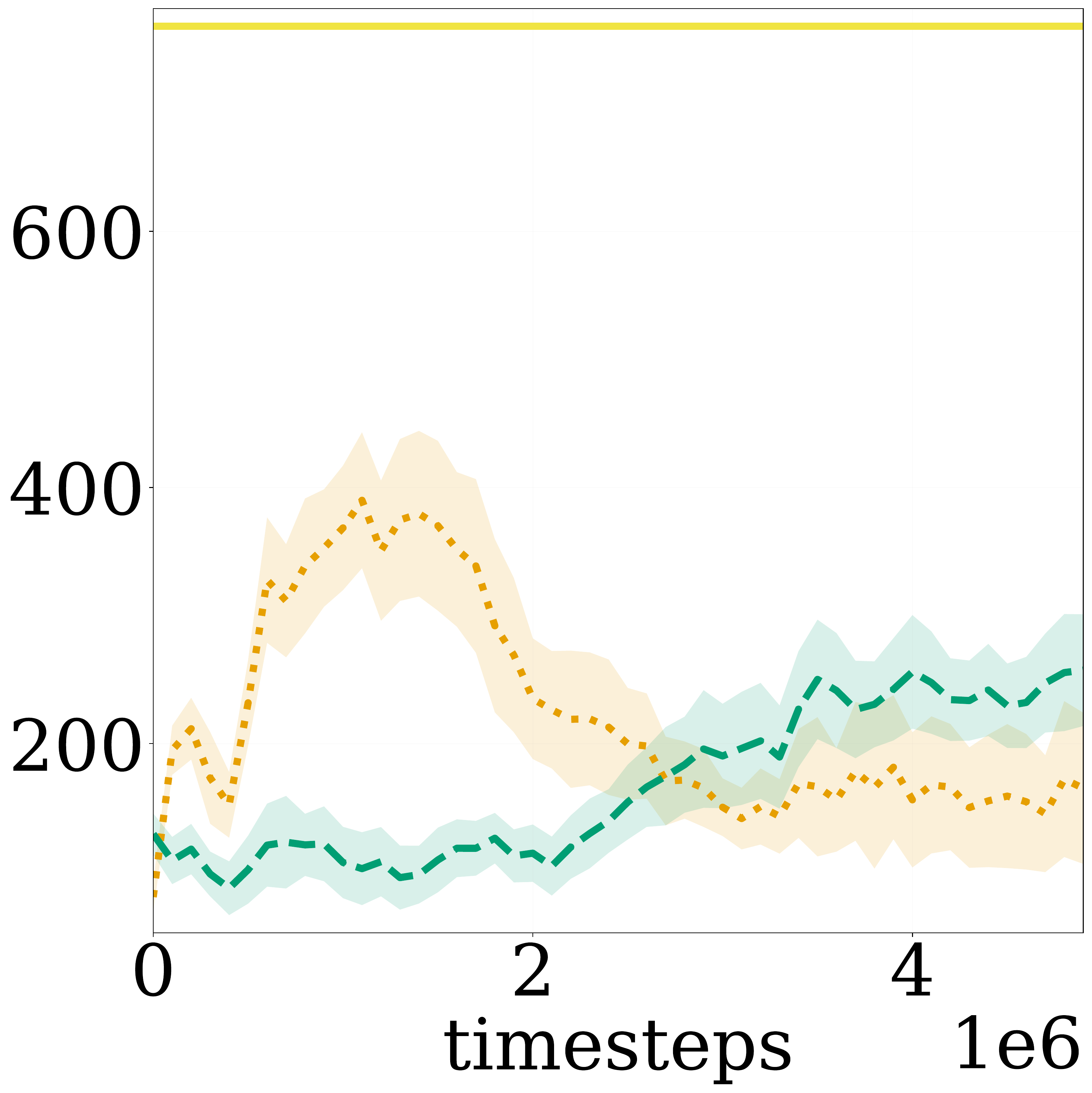}}
    \subfigure{\includegraphics[width=0.172\textwidth]{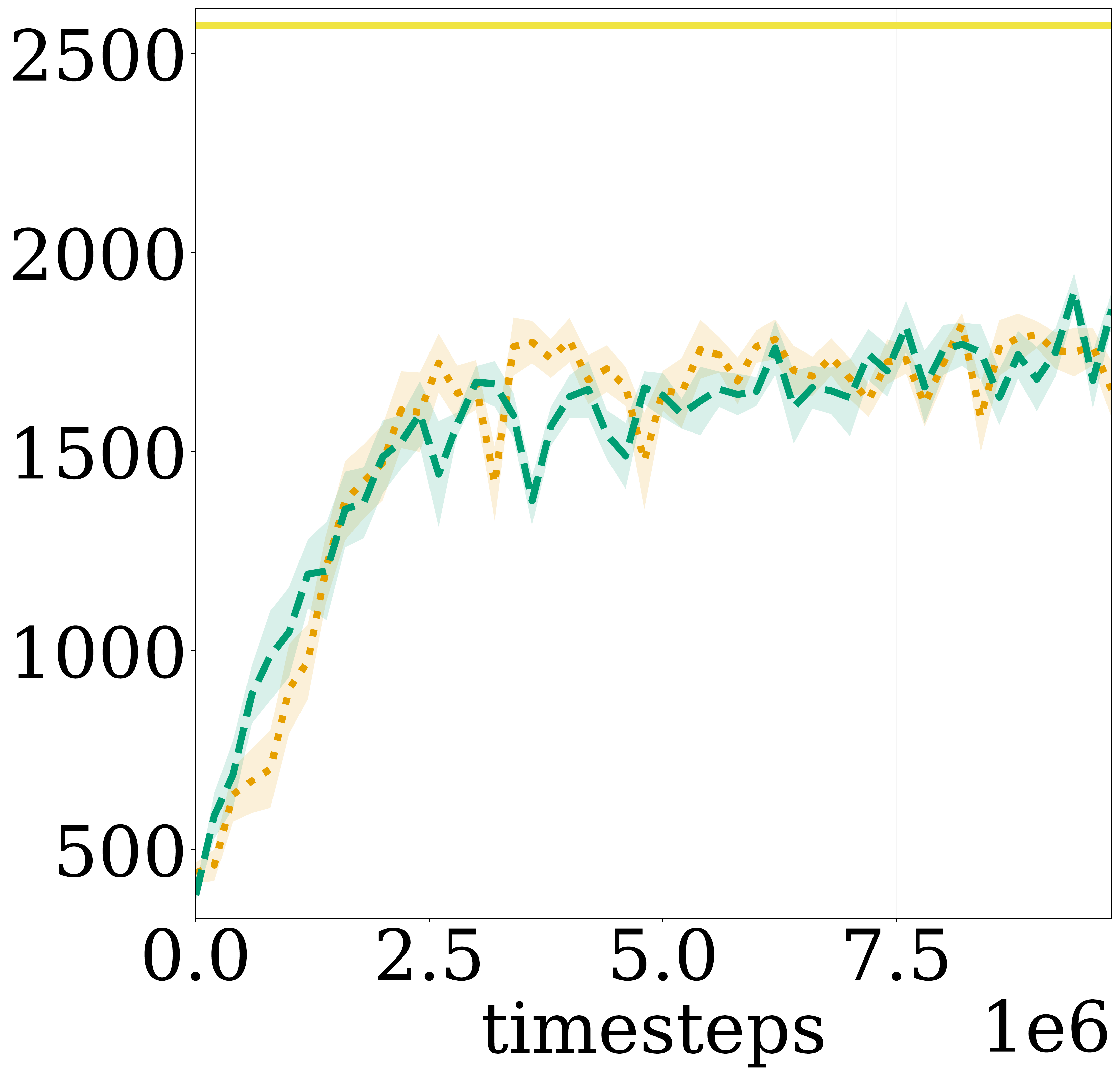}}
    \subfigure{\includegraphics[width=0.16\textwidth]{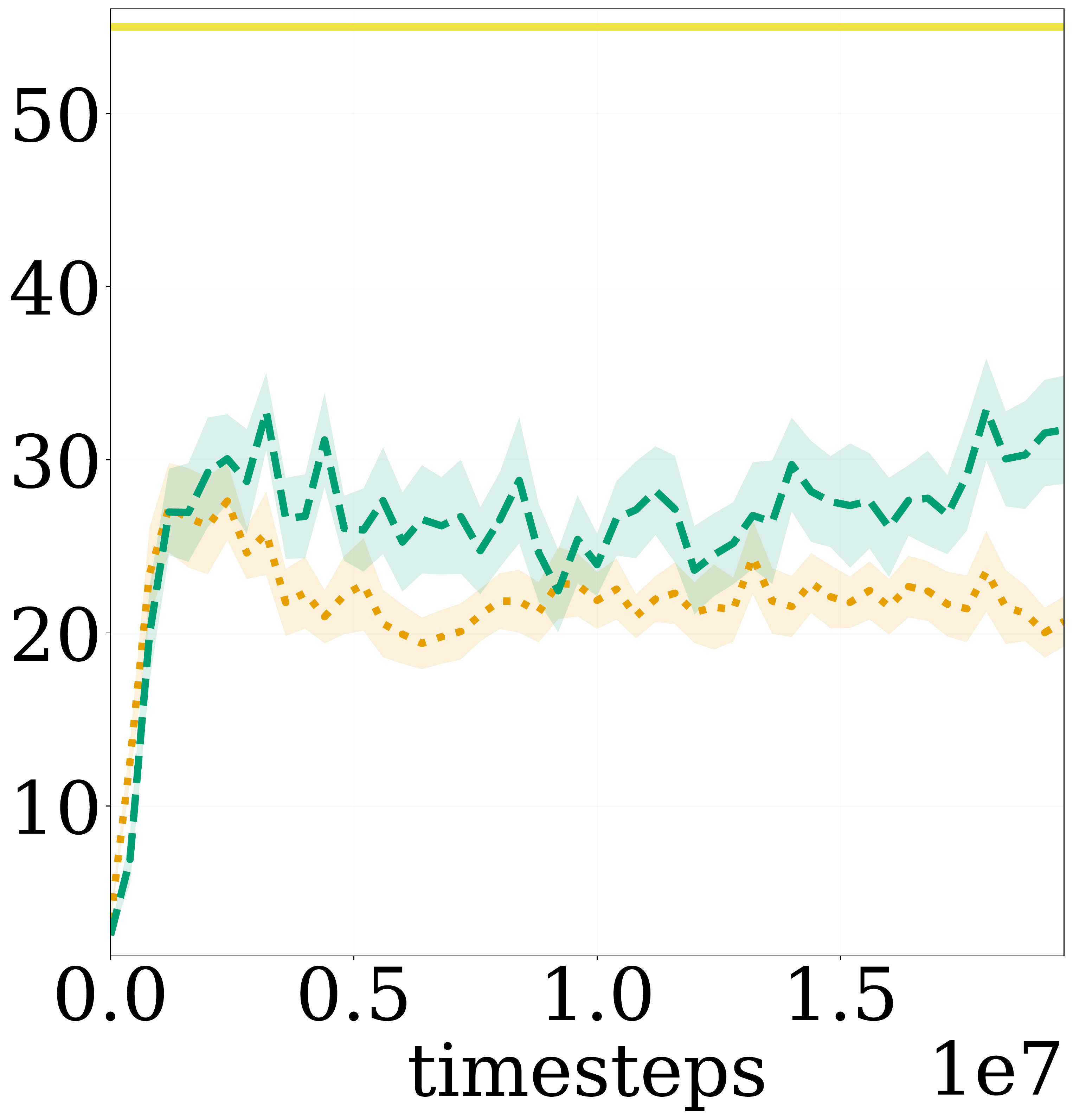}}\\
    \setcounter{subfigure}{0}
    \subfigure[Ant]{\includegraphics[width=0.185\textwidth]{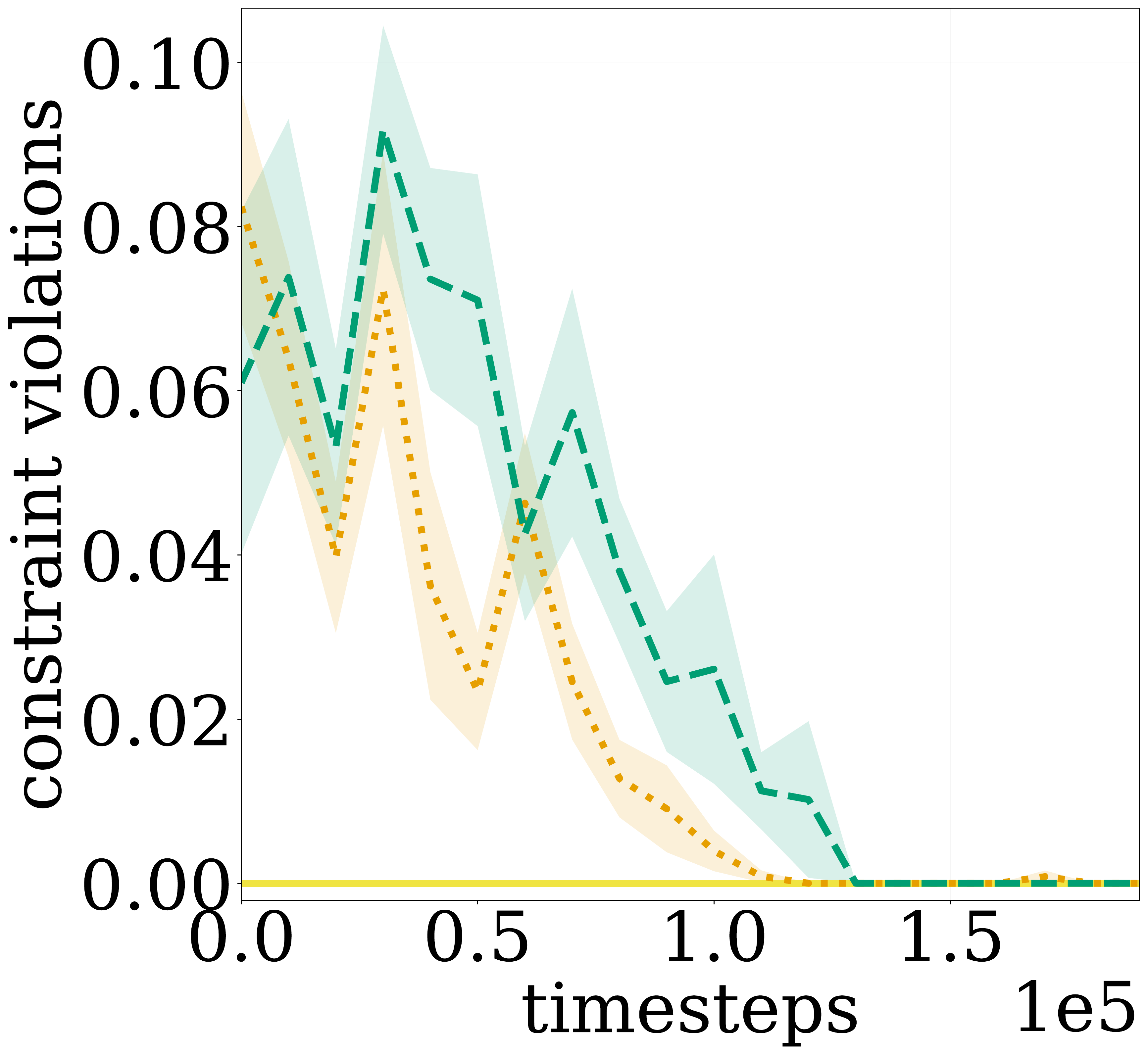}} 
    \hspace{0.12cm}
    \subfigure[Half-cheetah]{\includegraphics[width=0.16\textwidth]{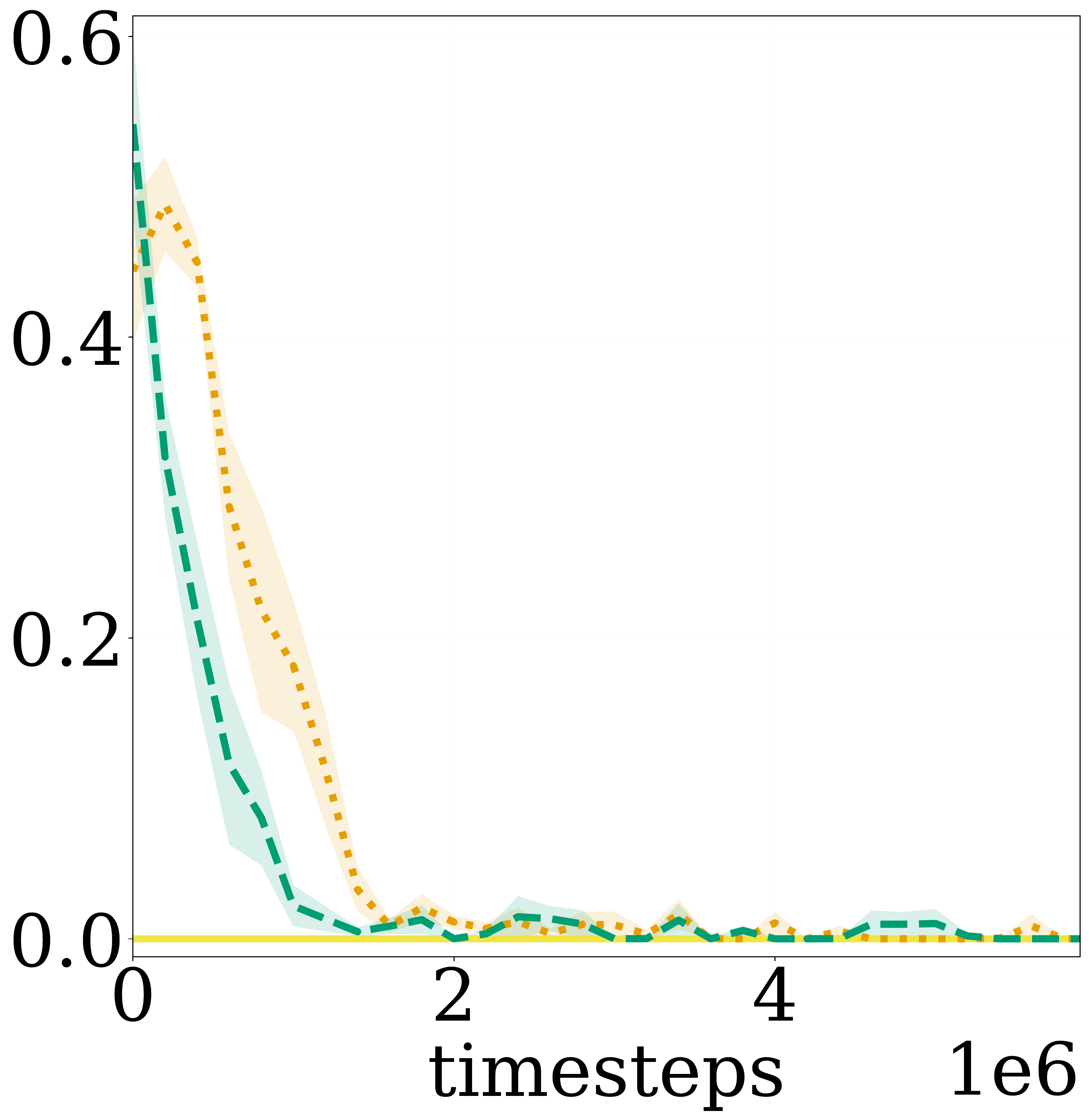}}
    \hspace{0cm}
    \subfigure[Swimmer]{\includegraphics[width=0.16\textwidth]{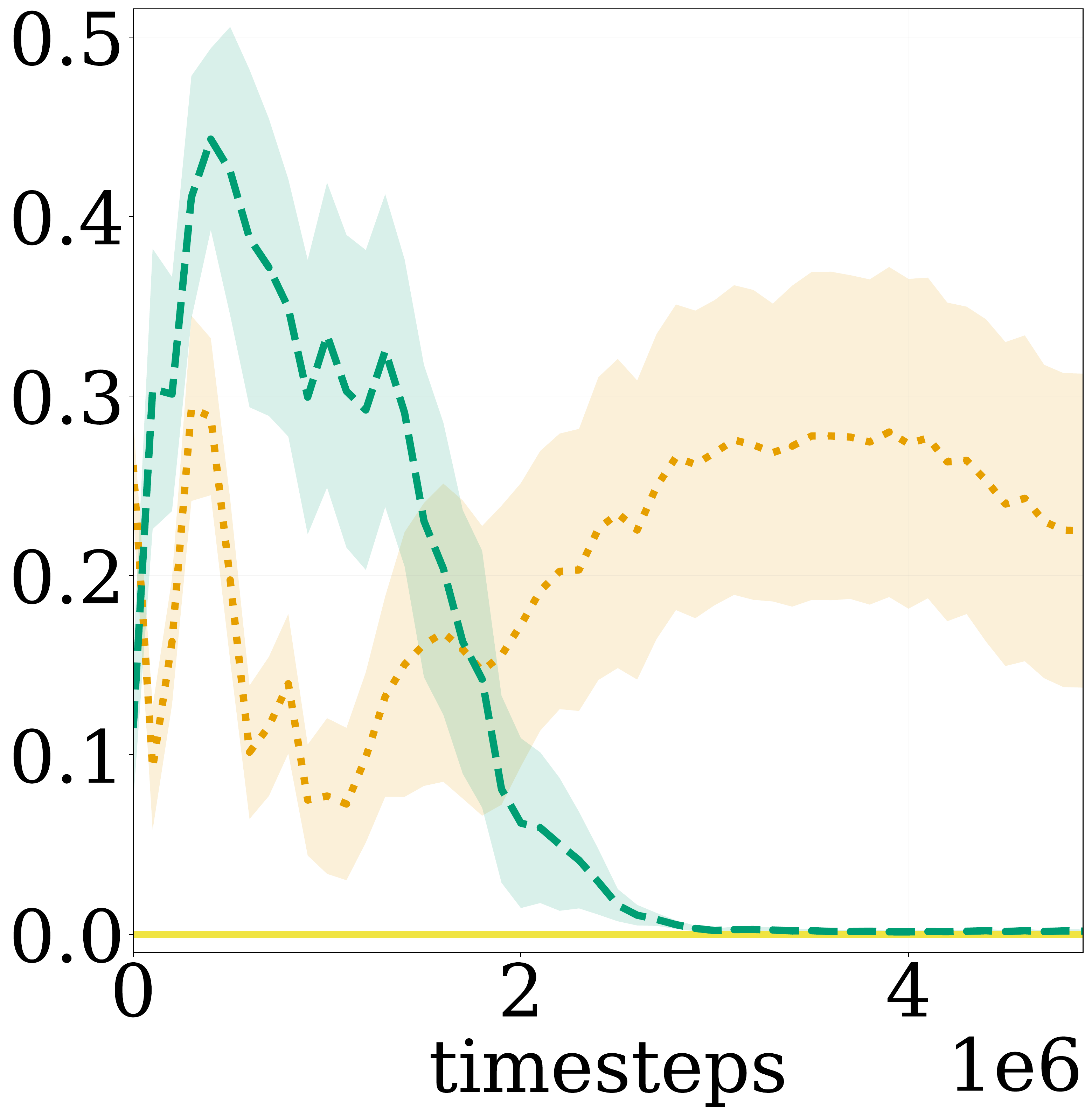}}
    \hspace{0cm}
    \subfigure[Walker]{\includegraphics[width=0.171\textwidth]{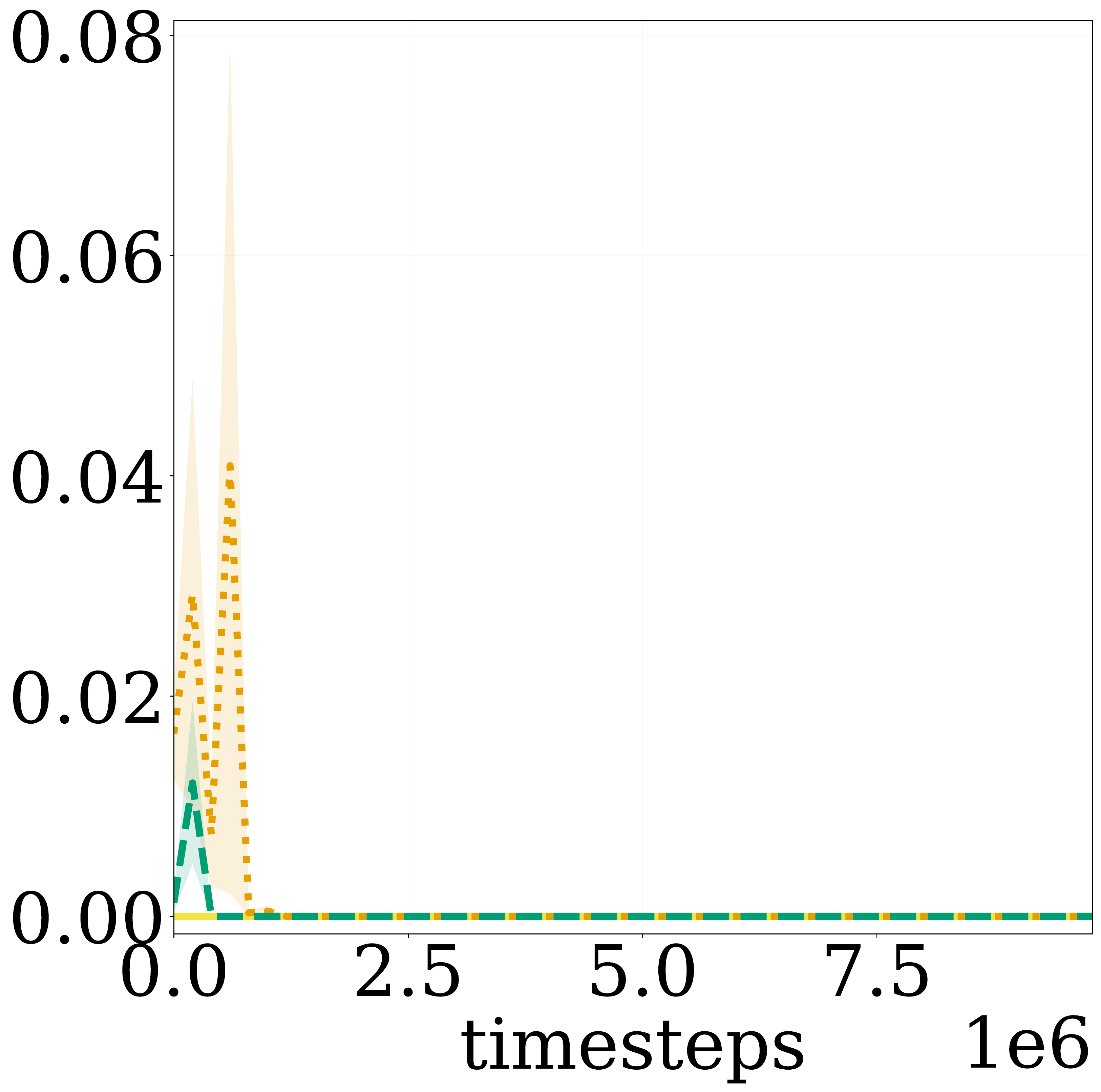}}
    \subfigure[Inverted pendulum]{\includegraphics[width=0.16\textwidth]{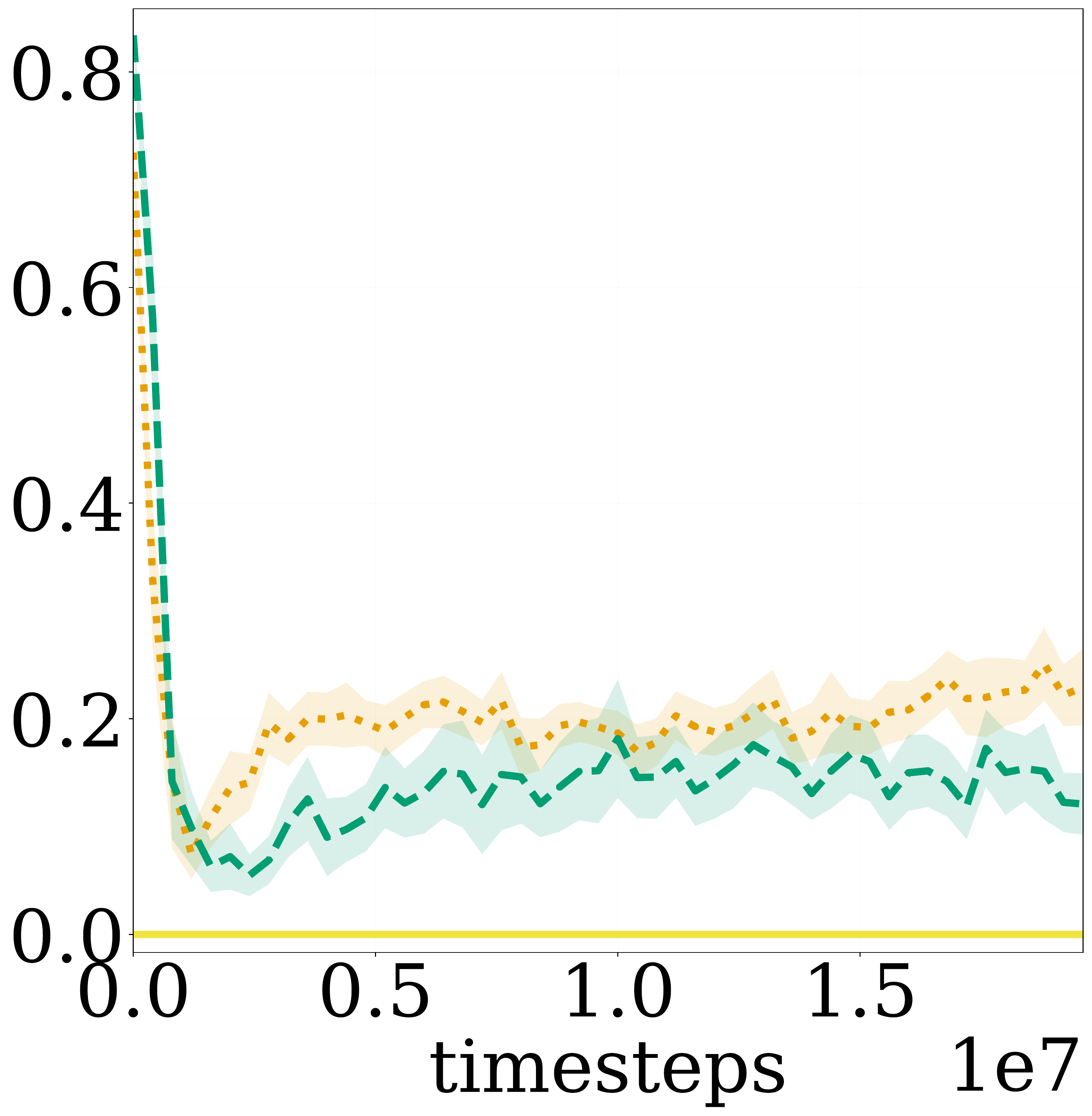}}
    \subfigure{\includegraphics[width=0.6\textwidth]{figures/legend_bootstrap_ablation.pdf}}\\
    \caption{Evaluation of our method in the virtual robotics environments for feature encoder pre-training disabled: reward and constraint violation rate of trajectories sampled from the nominal policy during training. Results are averaged over 10 random seeds. The x-axis is the number of timesteps taken in the environment during training. The shaded regions correspond with the standard error.}
    \label{fig:appendix-mujoco-bootstrap-ablation}
\end{center}
\end{figure*}

Figure~\ref{fig:resultsmujoco-ablations} depicts the reward received and the constraint violation rate during training for different values of $\beta$. \\
\\
Figure~\ref{fig:appendix-mujoco-bootstrap-ablation} depicts the reward received and the constraint violation rate during training for the ablation study on the pre-training of the feature encoder. 
\clearpage
\subsection{Realistic Traffic Environment}
\begin{figure}[h!]
%\vskip 0.2in
\begin{center}
\hspace{0.01cm}\subfigure{\includegraphics[width=0.18\textwidth]{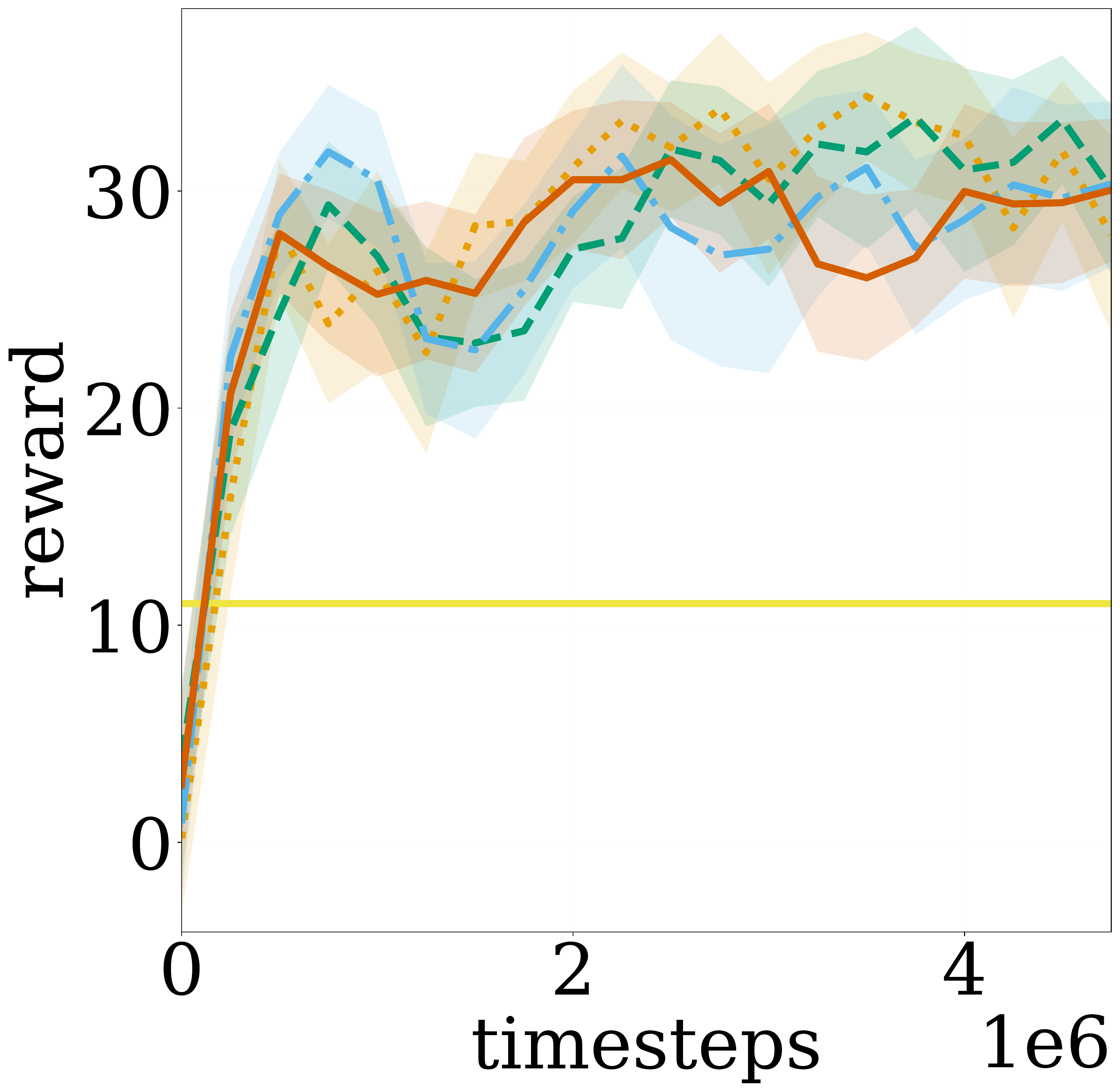}} \hspace{0.2cm}
    \subfigure{\includegraphics[width=0.18\textwidth]{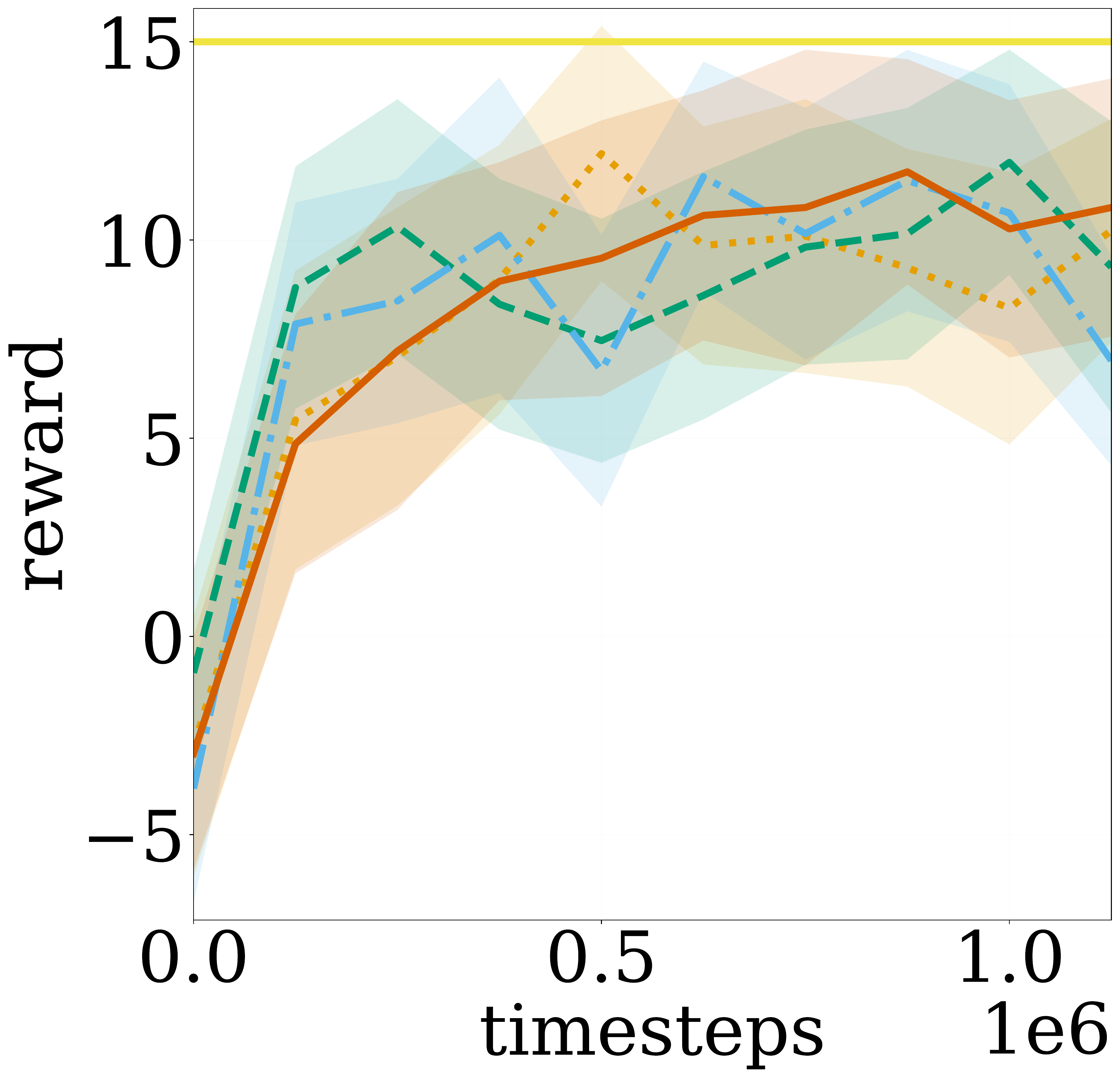}}\\
    \setcounter{subfigure}{0}
    \subfigure[Distance constraint]{\includegraphics[width=0.185\textwidth]{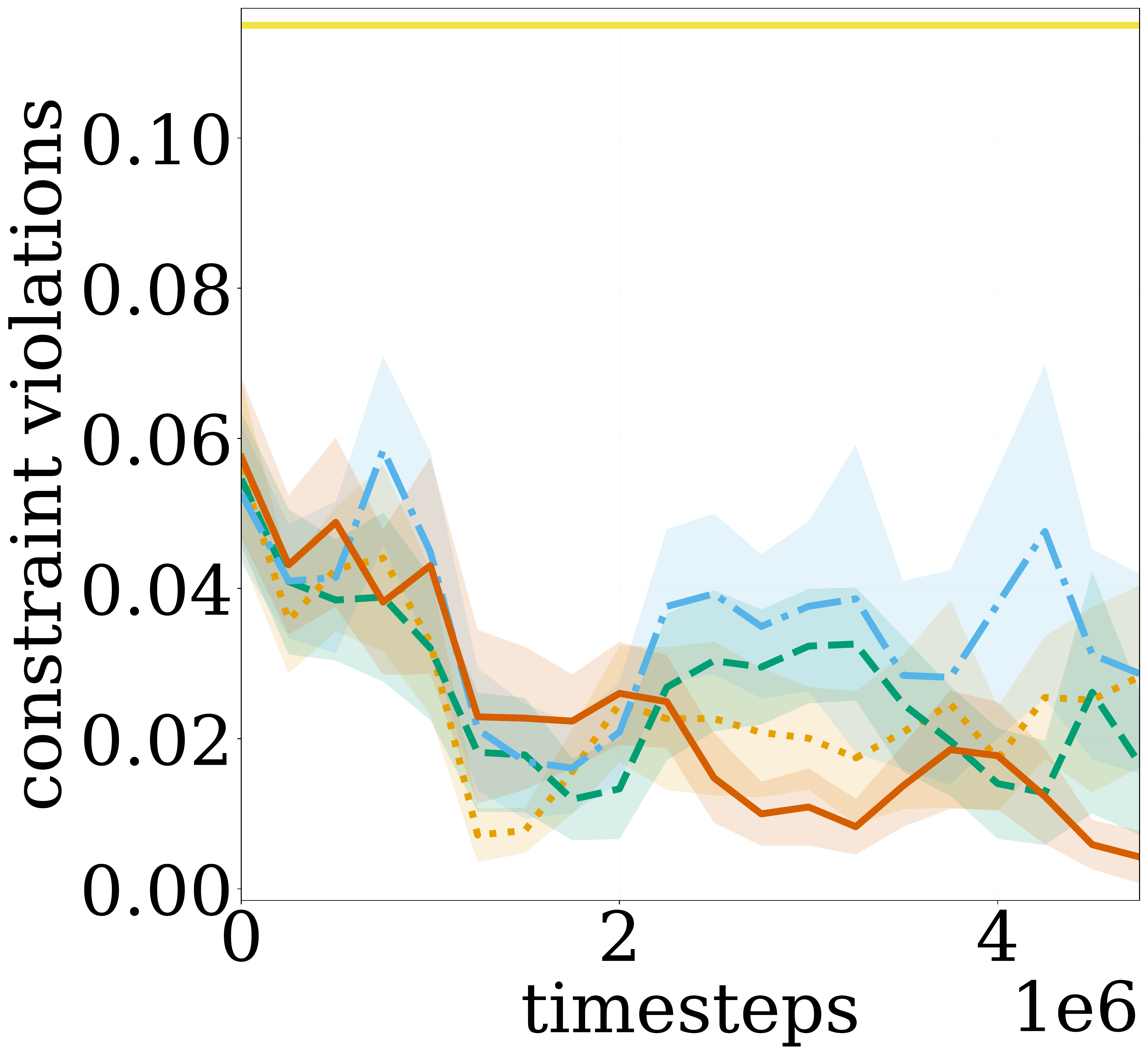}}\hspace{0.12cm}
    \subfigure[Velocity constraint]{\includegraphics[width=0.18\textwidth]{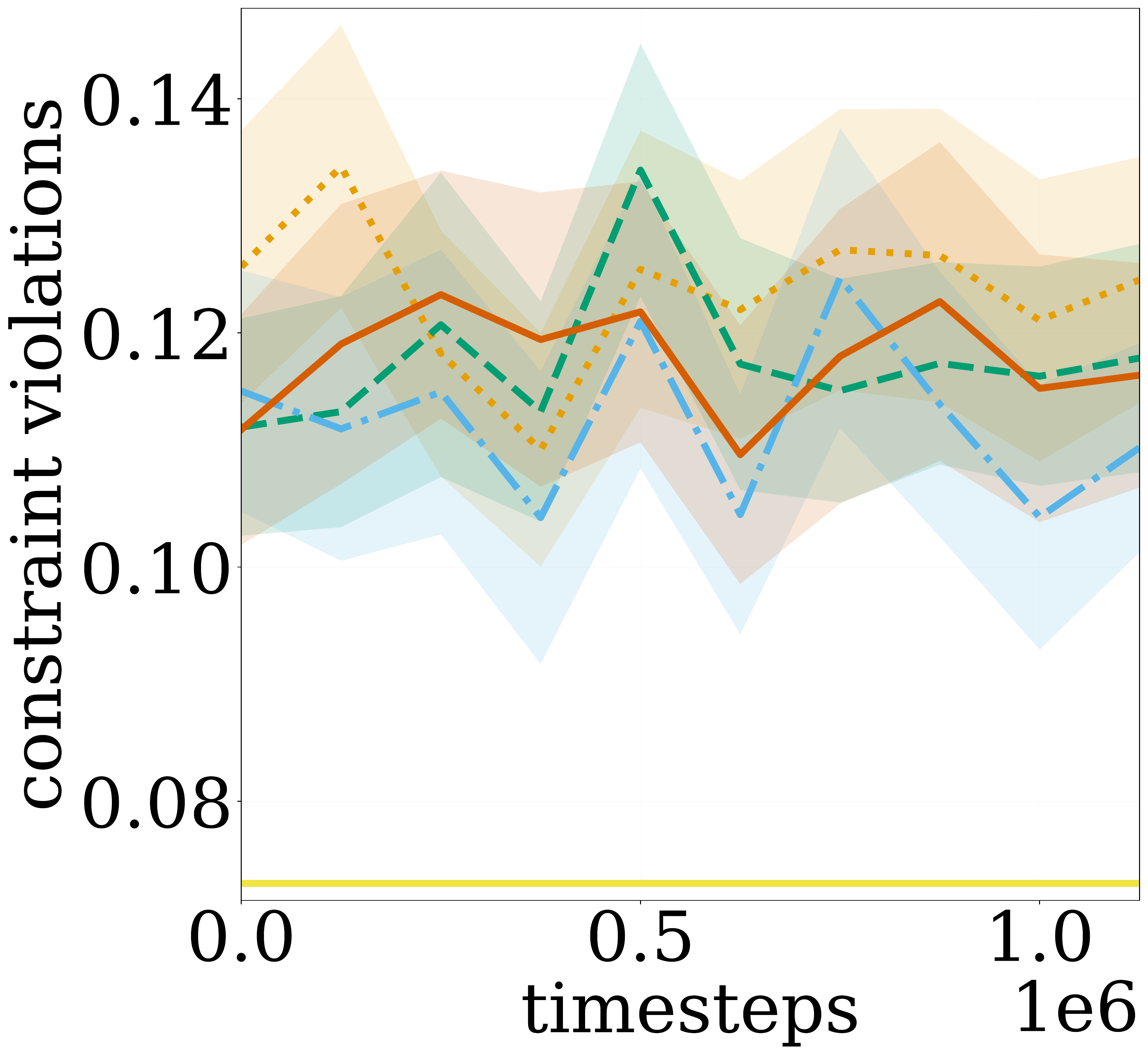}} \\
    \subfigure{\includegraphics[width=0.5\textwidth]{figures/legend_beta_ablation.pdf}}\\
    \caption{Evaluation of our method in the realistic traffic environment environment for different values of $\beta$: reward (top) and constraint violation rate (bottom) of trajectories sampled from the nominal policy during training. Results are averaged over 10 random seeds. The x-axis is the number of timesteps taken in the environment during training. The shaded regions correspond with the standard error.}
    \label{fig:appendix-highd-beta-ablation}
\end{center}
%\vskip -0.2in
\end{figure}

\begin{figure}[h!]
%\vskip 0.2in
\begin{center}
\hspace{0.01cm}\subfigure{\includegraphics[width=0.18\textwidth]{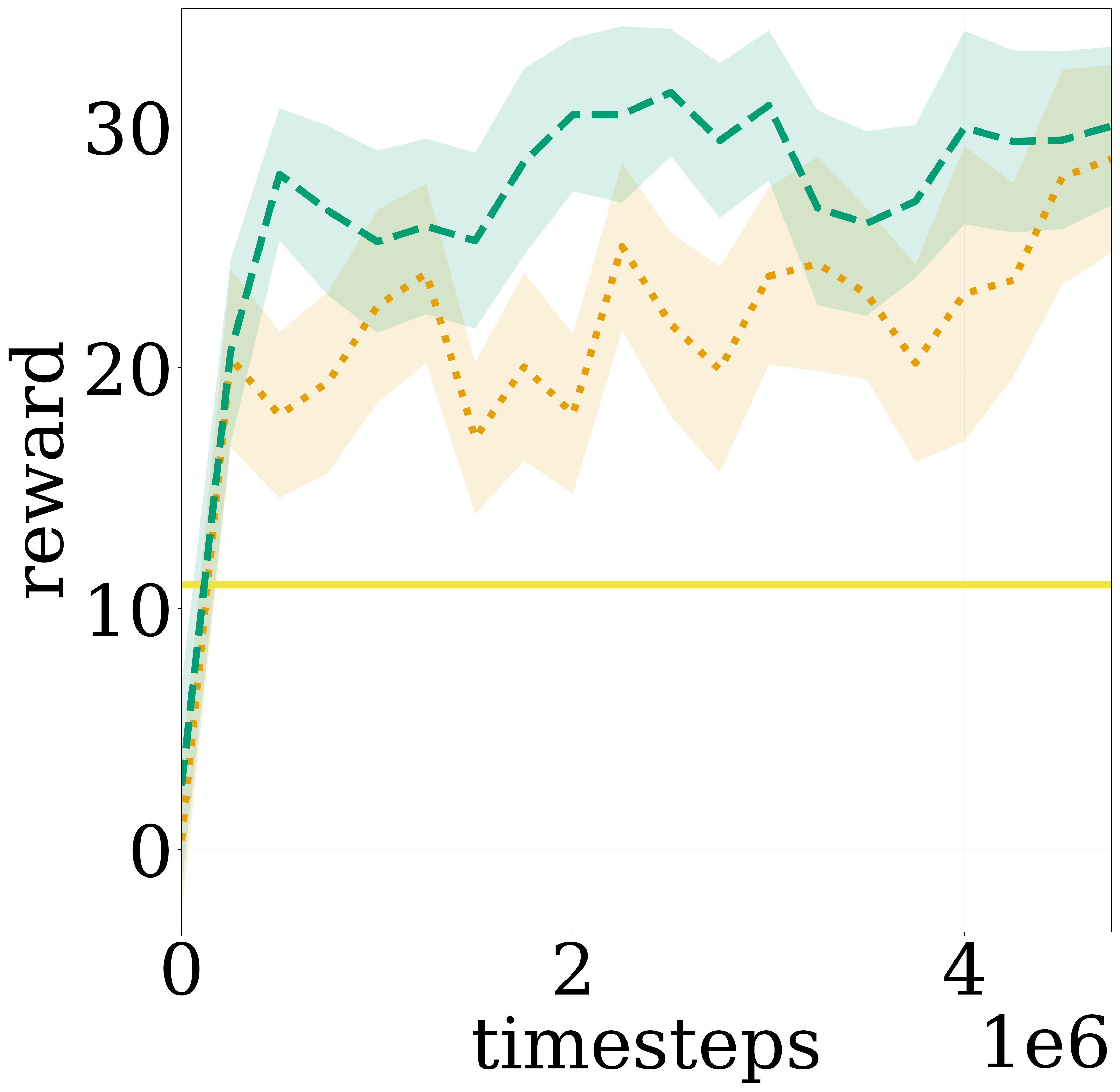}} \hspace{0.2cm}
    \subfigure{\includegraphics[width=0.18\textwidth]{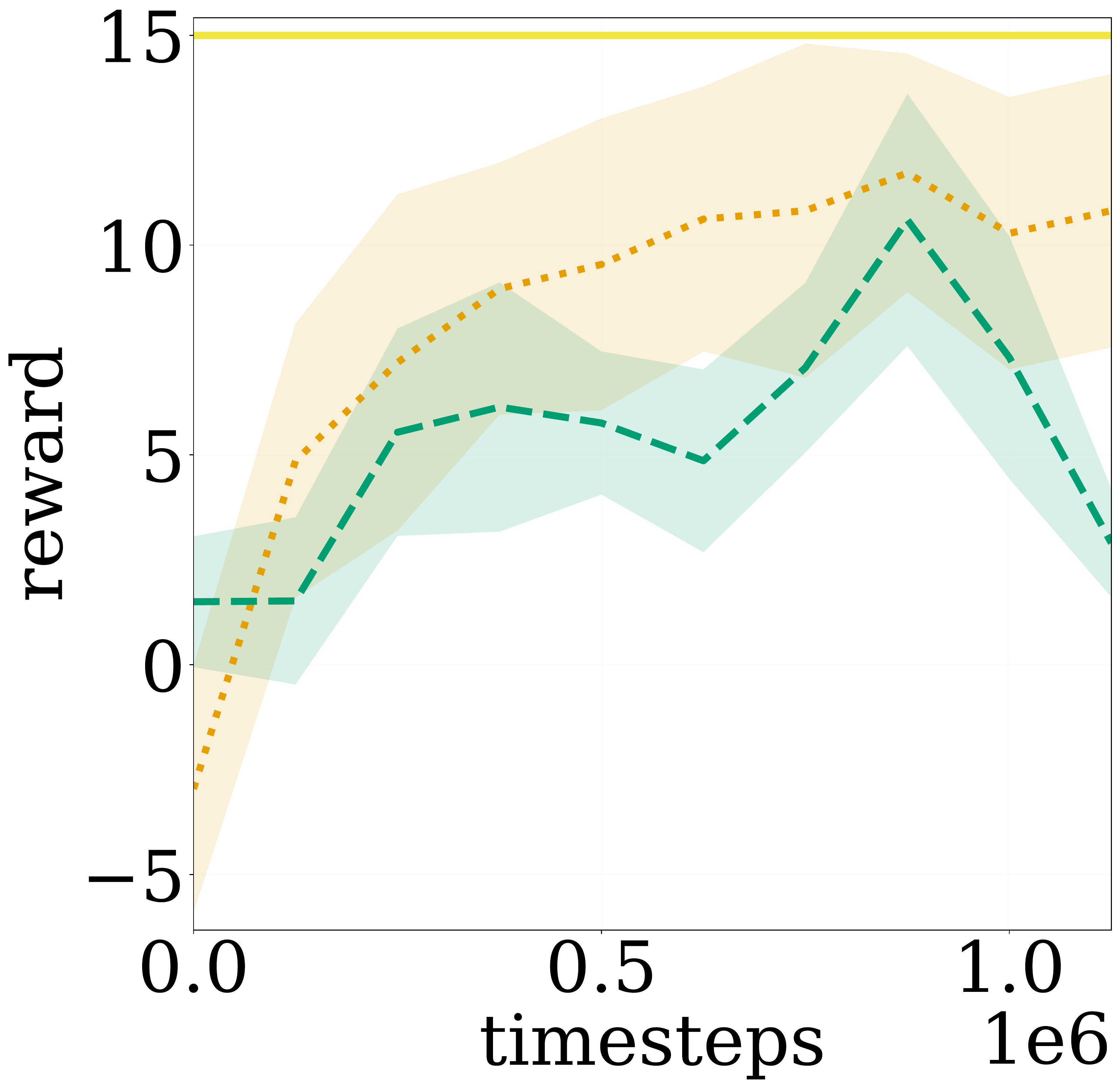}}\\
    \setcounter{subfigure}{0}
    \subfigure[Distance constraint]{\includegraphics[width=0.185\textwidth]{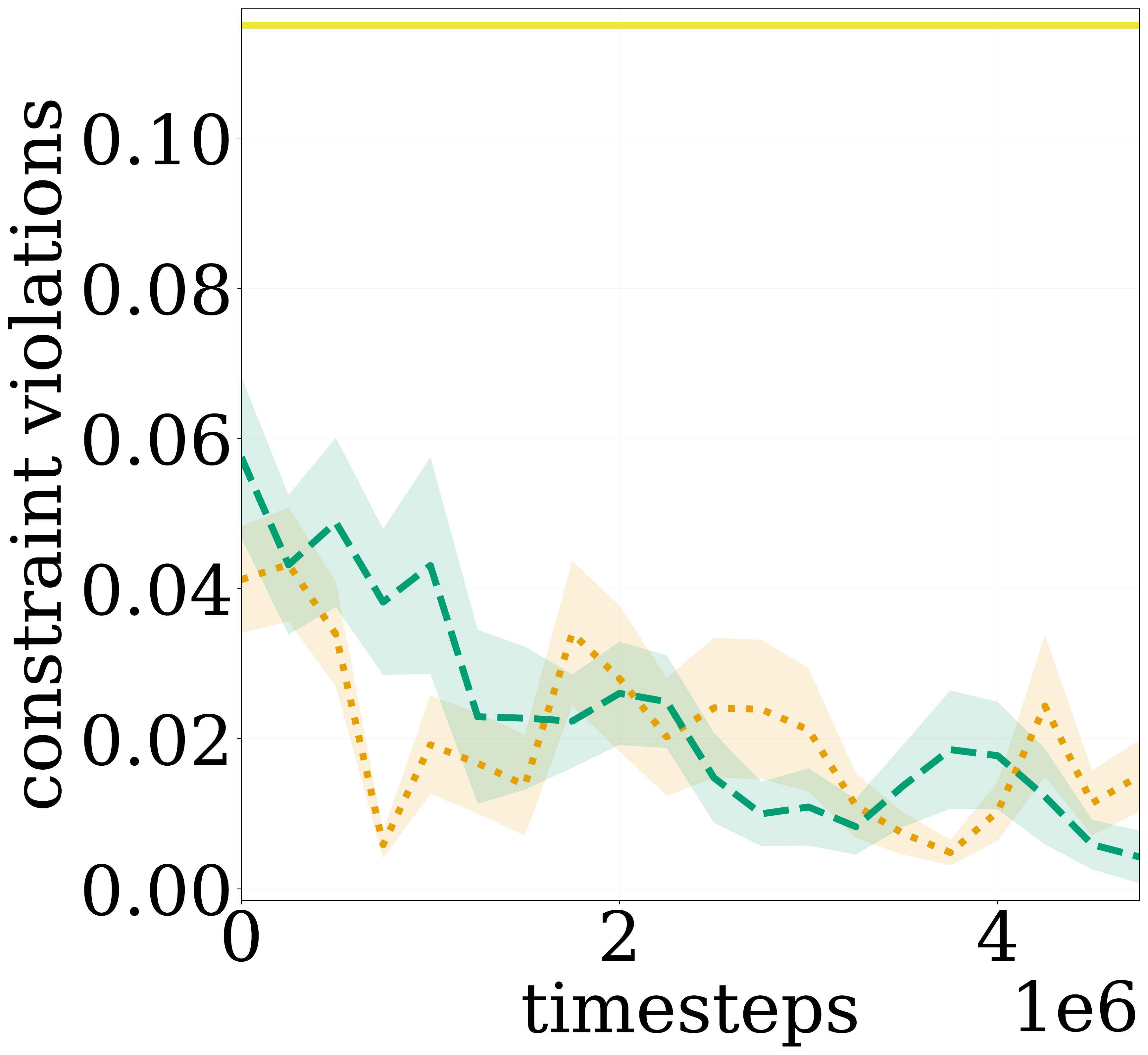}}\hspace{0.12cm}
    \subfigure[Velocity constraint]{\includegraphics[width=0.18\textwidth]{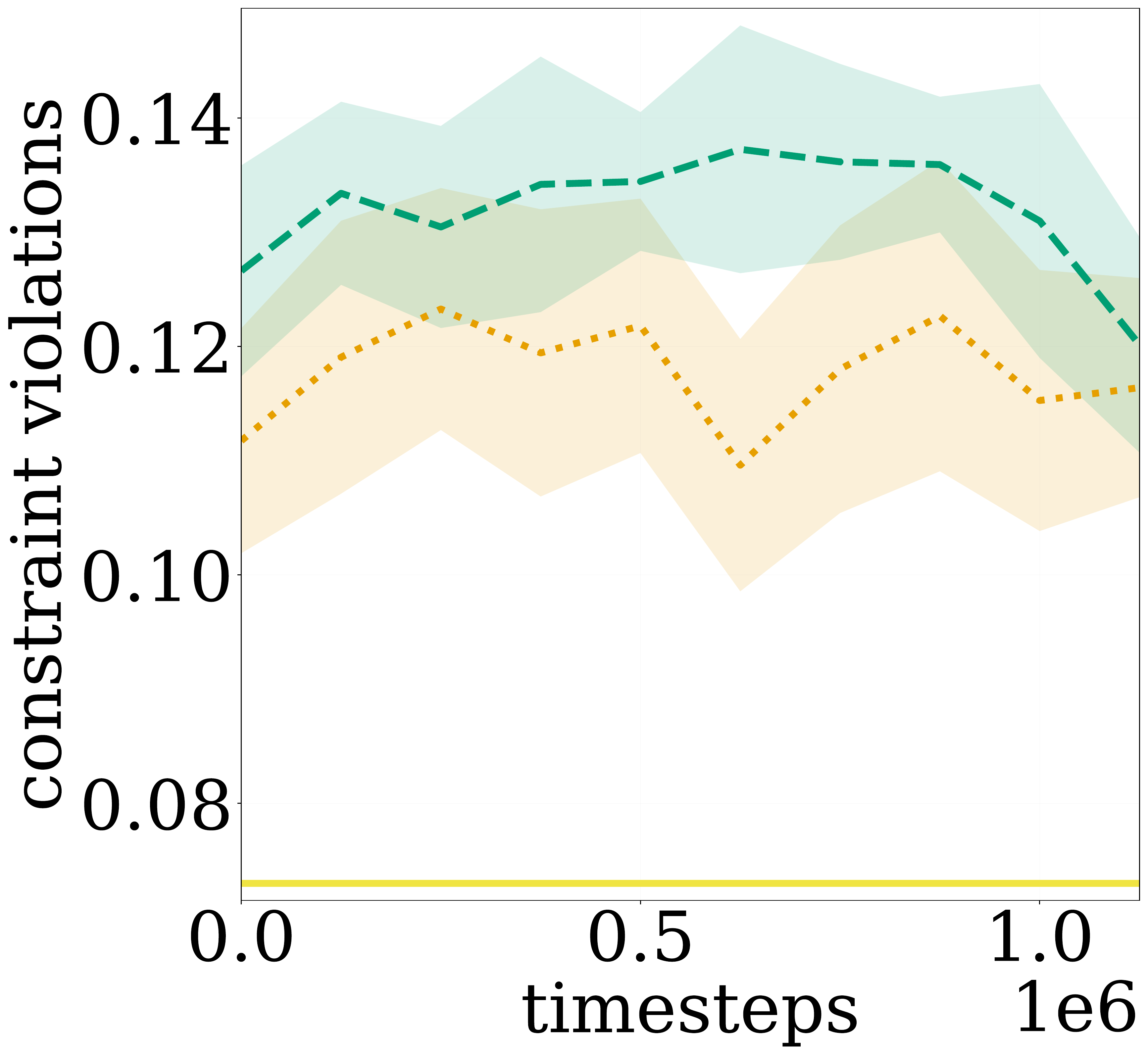}} \\
    \subfigure{\includegraphics[width=0.5\textwidth]{figures/legend_bootstrap_ablation.pdf}}\\
    \caption{Evaluation of our method in the realistic traffic environement for feature encoder pre-training disabled: reward and constraint violation rate of trajectories sampled from the nominal policy during training. Results are averaged over 10 random seeds. The x-axis is the number of timesteps taken in the environment during training. The shaded regions correspond with the standard error.}
    \label{fig:appendix-highd-bootstrap-ablation}
\end{center}
%\vskip -0.2in
\end{figure}

Figure~\ref{fig:appendix-highd-beta-ablation} depicts the reward and constraint violation rate during training at different timesteps for our method with various value of $\beta$.
\\
Figure~\ref{fig:appendix-highd-bootstrap-ablation} depicts the reward received and the constraint violation rate during training for the ablation study on the pre-training of the feature encoder. 

\end{document}